\documentclass[12pt]{article}

\usepackage[numbers,round]{natbib}
\usepackage{scicite}
\usepackage{url}

\usepackage{times}

\usepackage{amsmath}
\usepackage{amsfonts}
\usepackage{amssymb}
\usepackage{graphicx}
\usepackage{placeins}

\usepackage{subcaption}
\usepackage{float}
\usepackage{fancyvrb,fvextra}
\fvset{listparameters=\setlength{\topsep}{0pt}\setlength{\partopsep}{0pt}}
\usepackage{listings}
\usepackage{tikz}
\usetikzlibrary{shapes,arrows}
\usetikzlibrary{patterns}
\usetikzlibrary{decorations.pathreplacing, decorations.pathmorphing, calligraphy}
\tikzstyle{line} = [draw, very thick, -latex']
\tikzstyle{arrow} = [draw, ->, thick]

\definecolor{bpurp}{HTML}{984ea3}
\definecolor{bblue}{HTML}{377eb8}
\definecolor{bgreen}{HTML}{4daf4a}
\definecolor{borange}{HTML}{ff7f00}
\definecolor{bred}{HTML}{a50f15}
\definecolor{arbitrary}{HTML}{f1b6da}
\definecolor{consistent}{HTML}{80cdc1}
\definecolor{nonsense}{HTML}{737373} 
\definecolor{violate}{HTML}{8e1252}

\newenvironment{sciabstract}{%
\begin{quote} \bf}
{\end{quote}}

\title{Language models show human-like content effects on reasoning tasks}

\author
{
Andrew K. Lampinen,${}^{1\ast}$ Ishita Dasgupta,${}^{2*}$\\Stephanie C. Y. Chan,${}^{2}$ Hannah R. Sheahan,${}^{1}$ Antonia Creswell,${}^{3}$\\
 Dharshan Kumaran,${}^{1}$  James L. McClelland,${}^{2,4}$ Felix Hill,${}^{1}$\\
\normalsize{${}^{1}$Google DeepMind, London, UK}\\
\normalsize{${}^{2}$Google DeepMind, Mountain View, CA, USA}\\
\normalsize{${}^{3}$Work performed at Google DeepMind, London, UK}\\
\normalsize{${}^{4}$Department of Pscyhology, Stanford University, Stanford, CA, USA}\\
\normalsize{${}^{*}$Equal contribution; to whom correspondence should be addressed: \{lampinen,idg\}@google.com}\\
}

\date{}

\begin{document} 


\baselineskip24pt


\maketitle

\begin{sciabstract}
Abstract reasoning is a key ability for an intelligent system. Large language models (LMs) achieve above-chance performance on abstract reasoning tasks, but exhibit many imperfections. However, human abstract reasoning is also imperfect.
For example, human reasoning is affected by our real-world knowledge and beliefs, and shows notable ``content effects'';
humans reason more reliably when the semantic content of a problem supports the correct logical inferences.
These content-entangled reasoning patterns play a central role in debates about the fundamental nature of human intelligence. 
Here, we investigate whether language models --- whose prior expectations capture some aspects of human knowledge --- similarly mix content into their answers to logical problems.
We explored this question across three logical reasoning tasks: natural language inference, judging the logical validity of syllogisms, and the Wason selection task \citep{Wason1968ReasoningAA}. We evaluate state of the art large language models, as well as humans, and find that the language models reflect many of the same patterns observed in humans across these tasks --- like humans, models answer more accurately when the semantic content of a task supports the logical inferences.
These parallels are reflected both in answer patterns, and in lower-level features like the relationship between model answer distributions and human response times.
Our findings have implications for understanding both these cognitive effects in humans, and the factors that contribute to language model performance.
\end{sciabstract}

%\textbf{Teaser:} Language models and humans exhibit similar patterns of mixing semantic content into their performance on logical reasoning problems, leading to greater success in familiar situations, but more errors in unusual ones.

\section*{Introduction}

A hallmark of abstract reasoning is the ability to systematically perform algebraic operations over variables that can be bound to any entity \citep{newell1980physical, fodor1988connectionism}: the statement: ‘X is bigger than Y’ logically implies that ‘Y is smaller than X’, no matter the values of X and Y. That is, abstract reasoning is ideally content-independent \citep{newell1980physical}. The capacity for reliable and consistent abstract reasoning is frequently highlighted as a crucial missing component of current AI \citep{marcus2020next,mitchell2021abstraction,russin2020deep}. For example, while large Language Models (LMs) exhibit some impressive \emph{emergent} behaviors, including some performance on abstract reasoning tasks (\citealp{brown2020language,ganguli2022predictability, nye2021show, kojima2022large, wei2022emergent}; though cf. \citealp{schaeffer2023emergent}), they have been criticized for failing to achieve systematic consistency in their abstract reasoning \citep[e.g.][]{rae2021scaling,razeghi2022impact,patel2021nlp,valmeekam2022large}.

However, humans --- arguably the best known instances of general intelligence ---  are far from perfectly rational abstract reasoners \citep{gigerenzer2011heuristic, kahneman1982judgment, marcus2009kluge}. 
%However, these commentaries often overlook the fact that humans --- our standard for intelligent behavior --- are far from perfectly rational abstract reasoners \citep{gigerenzer2011heuristic, kahneman1982judgment, marcus2009kluge}. 
%
Patterns of biases in human reasoning have been studied across a wide range of tasks and domains \citep{kahneman1982judgment}. Here, we focus in particular on `content effects' --- the finding that humans are affected by the semantic content of a logical reasoning problem. In particular, humans reason more readily and more accurately about familiar, believable, or grounded situations, compared to unfamiliar, unbelievable, or abstract ones. For example, when presented with a syllogism like the following:
\begin{verbatim}
All students read.
Some people who read also write essays.
Therefore some students write essays.
\end{verbatim}
humans will often classify it as a valid argument. However, when presented with:
\begin{verbatim}
All students read.
Some people who read are professors.
Therefore some students are professors.
\end{verbatim}
humans are much less likely to say it is valid \citep{evans1983conflict,evans1995belief,klauer2000belief} --- despite the fact that the arguments above are logically equivalent (both are invalid). Similarly, humans struggle to reason about how to falsify conditional rules involving abstract attributes \citep{Wason1968ReasoningAA,johnson1999deductive}, but reason more readily about logically-equivalent rules grounded in realistic situations \citep{cheng1985pragmatic,cosmides1989logic,cosmides1992cognitive}. This human tendency also extends to other forms of reasoning e.g. probabilistic reasoning, where humans are notably worse when problems do not reflect intuitive expectations \citep{cohen2017beliefs}. 

The literature on human cognitive biases is extensive, but many of these biases can be idiosyncratic and context-dependent. For example, even some of the seminal findings in the influential work of Kahneman et al. \citep{kahneman1982judgment}, like `base rate neglect', are sensitive to context and experimental design \citep{dasgupta2020theory, benjamin18}, with several studies demonstrating exactly the opposite effect in a different context \citep{peterson1964uncertainty}.
However, the content effects on which we focus have been a notably consistent finding and have been documented in humans across different reasoning tasks and domains: deductive and inductive, or logical and probabilistic \citep{johnson1972reasoning,Wason1968ReasoningAA,wason1972psychology,evans1989bias,evans1983conflict,cohen2017beliefs}. This ubiquity is notable and makes these effects harder to explain as idiosyncracies. This ubiquitous sensitivity to content is in direct contradiction with the definition of abstract reasoning: that it is independent of content, and speaks directly to longstanding debates over the fundamental nature of human intelligence: are we best described as algebraic symbol-processing systems \citep{newell1980physical, marcus2003algebraic}, or emergent connectionist ones \citep{mcclelland2010letting,santoro2021symbolic} whose inferences are grounded in learned semantics? Yet explanations or models of content effects in the psychological sciences often focus on a single (task and content-specific) phenomenon and invoke bespoke mechanisms that only apply to these specific settings \citep[e.g.][]{cosmides1989logic}. Could content effects be explained more generally? Could they emerge from simple learning processes over naturalistic data?

In this work, we address these questions, by examining whether language models show this human-like blending of logic with semantic content effects. Language models possess prior knowledge --- expectations over the likelihood of particular sequences of tokens --- that are shaped by their training. Indeed, the goal of the ``pre-train and adapt'' or the ``foundation models'' \citep{bommasani2021opportunities} paradigm is to endow a model with broadly accurate prior knowledge that enables learning a new task rapidly. Thus, language model representations often \emph{reflect} human semantic cognition; e.g., language models reproduce patterns like association and typicality effects
\citep{bhatia2019distributed,bhatia2021transformer}, and language model predictions can reproduce human knowledge and beliefs \citep{trinh2018simple,petroni2019language,liu2021probing,jiang2021can}. In this work, we explore whether this prior knowledge impacts a language model's performance in logical reasoning tasks. While a variety of recent works have explored biases and imperfections in language models' performance \citep[e.g.][]{rae2021scaling,razeghi2022impact,patel2021nlp,sogaard2021locke,valmeekam2022large}, we focus on the specific question of whether content interacts with logic in these systems as it does % in the wide-ranging findings of content effects
in humans. This question has significant implications not only for characterizing LMs, but potentially also for understanding human cognition, by contributing new ways of understanding the balance, interactions, and trade-offs between the abstract and grounded capabilities of a system.

We explore how the content of logical reasoning problems affects the performance of a range of large language models \citep{hoffmann2022training,gpt35,anil2023palm}. To avoid potential dataset contamination, we create entirely new datasets using designs analogous to those used in prior cognitive work, and we also collect directly-comparable human data with our new stimuli.
% --- is affected by the about the entities and relationships in the reasoning problems. Indeed, 
We find that language models reproduce human content effects %from the cognitive literature
across three different logical reasoning tasks (Fig. \ref{fig:manipulating_structure}). We first explore a simple Natural Language Inference (NLI) task, and show that models and humans answer fairly reliably, with relatively modest influences of content. We then examine the more challenging task of judging whether a syllogism is a valid argument, and show that models and humans are biased by the believability of the conclusion. We finally consider realistic and abstract/arbitrary versions of the Wason selection task \citep{Wason1968ReasoningAA} --- a task introduced over 50 years ago that demonstrates a failure of systematic human reasoning --- and show that models and humans perform better with a realistic framing. Our findings with human participants replicate and extend existing findings in the cognitive literature. We also report novel analyses of item-level effects, and the effect of content and items on continuous measures of model and human responses. We close with a discussion of the implications of these findings for understanding human cognition as well as language models.

\subsection*{Evaluating content effects on logical tasks}

In this work, we evaluate content effects on three logical reasoning tasks, which are depicted in Fig. \ref{fig:manipulating_structure}. These three tasks involve different types of logical inferences, and different kinds of semantic content. However, these distinct tasks admit a consistent definition of content effects: the extent to which reason is facilitated in situations in which the semantic content supports the correct logical inference, and correspondingly the extent to which reasoning is harmed when semantic content conflicts with the correct logical inference (or, in the Wason tasks, when the content is simply arbitrary). We also evaluate versions of each task where the semantic content is replaced with nonsense non-words, which lack semantic content and thus should neither support nor conflict with reasoning performance. (However, note that in some cases, particularly the Wason tasks, changing to nonsense content requires more substantially altering the kinds of inferences required in the task; see Methods.)

\begin{figure*}[t!]
    \centering
    \resizebox{0.99\textwidth}{!}{
    \begin{tikzpicture}
    \tikzstyle{every node}=[font=\footnotesize] 

    %% labels 
    \node [align=right, anchor=east] at (-3.33, 0) {\large \bf NLI};
    \node [align=right, anchor=east] at (-3.33, -1.75) {\large \bf Syllogisms};
    \node [align=right, anchor=east] at (-3.33, -5.3) {\large \bf Wason};
    
    %% NLI
    \node [draw, text width=5.1cm] at (0, 0) (nlicon) {If {\color{consistent}\bf seas} are bigger than {\color{consistent}\bf puddles}, then {\color{consistent}\bf puddles} are smaller than {\color{consistent}\bf seas}};
    
    \node [draw, text width=5.1cm] at (6.5, 0) (nlivio)  
    {If {\color{violate}\bf puddles} are bigger than {\color{violate}\bf seas}, then {\color{violate}\bf seas} are smaller than {\color{violate}\bf puddles}};
    
    \node [draw, text width=5.1cm] at (13, 0) (nlinon)  {If {\color{nonsense}\bf vuffs} are bigger than {\color{nonsense}\bf feps}, then {\color{nonsense}\bf feps} are smaller than {\color{nonsense}\bf vuffs}};
    
    %% Syllogisms 
    \node [draw, text width=5.1cm] at (0, -1.75) (sylcon) {All {\color{consistent}\bf guns} are {\color{consistent}\bf weapons}.\\ All {\color{consistent}\bf weapons} are {\color{consistent}\bf dangerous things}.\\ All {\color{consistent}\bf guns} are {\color{consistent}\bf dangerous things}.};
    
    \node [draw, text width=5.1cm] at (6.5, -1.75) (sylvio)  {All {\color{violate}\bf dangerous things} are {\color{violate}\bf weapons}.\\ All {\color{violate}\bf weapons} are  {\color{violate}\bf guns}.\\ All {\color{violate}\bf dangerous things} are {\color{violate}\bf guns}.};
    
    \node [draw, text width=5.1cm] at (13, -1.75) (sylnon)  {All {\color{nonsense}\bf zoct} are {\color{nonsense}\bf spuff}.\\ All {\color{nonsense}\bf spuff} are {\color{nonsense}\bf thrund}.\\ All {\color{nonsense}\bf zoct} are {\color{nonsense}\bf thrund}.};
    
    %% Wason
    \node [draw, text width=5.1cm] at (0, -5.3) (wascon) {If the clients are going {\color{consistent}\bf skydiving}, then they must have a {\color{consistent}\bf parachute}.\\
    card: {\color{consistent}\bf skydiving}\\
    card: {\color{consistent}\bf scuba diving}\\
    card: {\color{consistent}\bf parachute}\\
    card: {\color{consistent}\bf wetsuit}
    };
    
    \node [draw, text width=5.2cm] at (6.5, -5.3) (wasvio)  {If the cards have {\color{arbitrary}\bf plural word}, then they must have a {\color{arbitrary}\bf positive emotion}.\\[-0.06em]
    card: {\color{arbitrary}\bf shoes}\\[-0.06em]
    card: {\color{arbitrary}\bf dog}\\[-0.06em]
    card: {\color{arbitrary}\bf happiness}\\[-0.06em]
    card: {\color{arbitrary}\bf anxiety}
    };
    
    \node [draw, text width=5.1cm] at (13, -5.3) (wasnon)  {If the cards have {\color{nonsense}\bf more bem}, then they must have {\color{nonsense}\bf less stope}. \\
    card: {\color{nonsense}\bf more bem}\\
    card: {\color{nonsense}\bf less bem}\\
    card: {\color{nonsense}\bf less stope}\\
    card: {\color{nonsense}\bf more stope}
    };

    %% labels
    \node at (0,   1) {\large \bf Consistent};
    \node at (6.5, 1) {\large \bf Violate};
    \node at (13,  1) {\large \bf Nonsense};
    
    \node at (0,   -3.5) {\large \bf Realistic};
    \node at (6.5, -3.5) {\large \bf Arbitrary};
    \node at (13,  -3.5) {\large \bf Nonsense};
    
    \end{tikzpicture}
    }
    \caption{Manipulating content within fixed logical structures. In each of our three datasets (rows), we instantiate different versions of the logical problems (columns). Different versions of a problem offer the same logical structures and tasks, but instantiated with different entities or relationships between those entities. The relationships in a task may either be consistent with, or violate real-world semantic relationships, or may be nonsense, without semantic content. In general, humans and models reason more accurately about belief-consistent or realistic situations or rules than belief-violating or arbitrary ones.}
    \label{fig:manipulating_structure}
\end{figure*}

\paragraph{Natural Language Inference}
The first task we consider has been studied extensively in the natural language processing literature \citep{maccartney2007natural}. In the classic Natural Language Inference (NLI) problem, a model receives two sentences, a `premise' and a `hypothesis', and has to classify them based on whether the hypothesis `entails',  `contradicts', or `is neutral to' the premise. Traditional datasets for this task were crowd-sourced \citep{bowman2015large} leading to sentence pairs that don't strictly follow logical definitions of entailment and contradiction.
To make this a more strictly logical task, we follow Dasgupta et al. \citep{dasgupta2018evaluating} to generate comparisons 
(e.g. \texttt{X is smaller than Y}).
%(e.g. \texttt{Premise: X is smaller than Y, Hypothesis: Y is bigger than X} is an entailment).
We then give participants an incomplete inference such as ``If puddles are bigger than seas, then...'' and give them a forced choice between two possible hypotheses to complete it: ``seas are bigger than puddles'' and ``seas are smaller than puddles.'' Note that one of these completions is consistent with real-world semantic beliefs i.e. `believable' while the other is logically consistent with the premise but contradicts real world beliefs.
We can then evaluate whether models and humans answer more accurately when the logically correct hypothesis is believable; that is, whether the content affects their logical reasoning.

However, content effects are generally more pronounced in difficult tasks that require extensive logical reasoning %, and are stronger in children or adults under cognitive load
\citep{evans1989bias,evans1995belief}. 
We therefore consider two more challenging tasks where human content effects have been observed in prior work.

\paragraph{Syllogisms}
Syllogisms~\citep{sep-aristotle-logic} are a simple argument form in which two true statements necessarily imply a third. For example, the statements ``All humans are mortal'' and ``Socrates is a human'' together imply that ``Socrates is mortal''. But human syllogistic reasoning is not purely abstract and logical; instead it is affected by our prior beliefs about the contents of the argument \citep{evans1983conflict,klauer2000belief,tessler2022logic}.
For example, Evans et al. \citep{evans1983conflict} showed that if participants were asked to judge whether a syllogism was logically valid or invalid, they were biased by whether the conclusion was consistent with their beliefs. Participants were very likely (90\% of the time) to mistakenly say an invalid syllogism was valid if the conclusion was believable, and thus mostly relied on belief rather than abstract reasoning. Participants would also sometimes say that a valid syllogism was invalid if the conclusion was not believable, but this effect was somewhat weaker \citep[but cf.][]{dube2010assessing}. These ``belief-bias'' effects have been replicated and extended in various subsequent studies \citep{klauer2000belief,dube2010assessing,trippas2014using,tessler2015understanding}. We similarly evaluate whether models and humans are more likely to endorse an argument as valid if its conclusion is believable, or to dismiss it as invalid if its conclusion is unbelievable.

\paragraph{The Wason Selection Task}

The Wason Selection Task \citep{Wason1968ReasoningAA} is a logic problem that can be challenging even for subjects with substantial education in mathematics or philosophy. Participants are shown four cards, and told a rule such as: ``if a card has a `D' on one side, then it has a `3' on the other side.'' The four cards respectively show `D', `F', `3', and `7'. The participants are then asked which cards they need to flip over to check if the rule is true or false.
The correct answer is to flip over the cards showing `D' and `7'. However, Wason \citep{Wason1968ReasoningAA} showed that while most participants correctly chose `D', they were much more likely to choose `3' than `7'. That is, the participants should check the \emph{contrapositive} of the rule (``not 3 implies not D'', which is logically implied), but instead they confuse it with the \emph{converse} (``3 implies D'', which is not logically implied). 
This is a classic task in which reasoning according to the rules of formal logic does not come naturally for humans, and thus there is potential for prior beliefs and knowledge to affect reasoning.

Indeed, the difficulty of the Wason task depends upon the content of the problem. Past work has found that if an identical logical structure is instantiated in a common situation, particularly a social rule, participants are much more accurate \citep{wason1971natural,cheng1985pragmatic,cosmides1989logic,cosmides1992cognitive}. For example, if participants are told the cards represent people, and the rule is ``if they are drinking alcohol, then they must be 21 or older'' and the cards show `beer', `soda', `25', `16', then many more participants correctly choose to check the cards showing `beer' and `16'. We therefore similarly evaluate whether language models and humans are facilitated in reasoning about realistic rules, compared to more-abstract arbitrary ones. (Note that in our implementations of the Wason task, we forced participants and language models to choose exactly two cards, in order to most closely match answer formats between the humans and language models.)

The extent of content effects on the Wason task are also affected by background knowledge; education in mathematics appears to be associated with improved reasoning in abstract Wason tasks \citep{inglis2004mathematicians,cresswell2020does}. 
However, even those experienced participants were far from perfect --- undergraduate mathematics majors and academic mathematicians achieved less than 50\% accuracy at the arbitrary Wason task \citep{inglis2004mathematicians}. This illustrates the challenge of abstract logical reasoning, even for experienced humans. As we will see in the next section, many human participants did struggle with the abstract versions of our tasks.

\section*{Results}

\subsection*{Content effects on accuracy}

\begin{figure}[t]
    \centering
    \begin{subfigure}[t]{0.33\textwidth}
    \includegraphics[width=\textwidth]{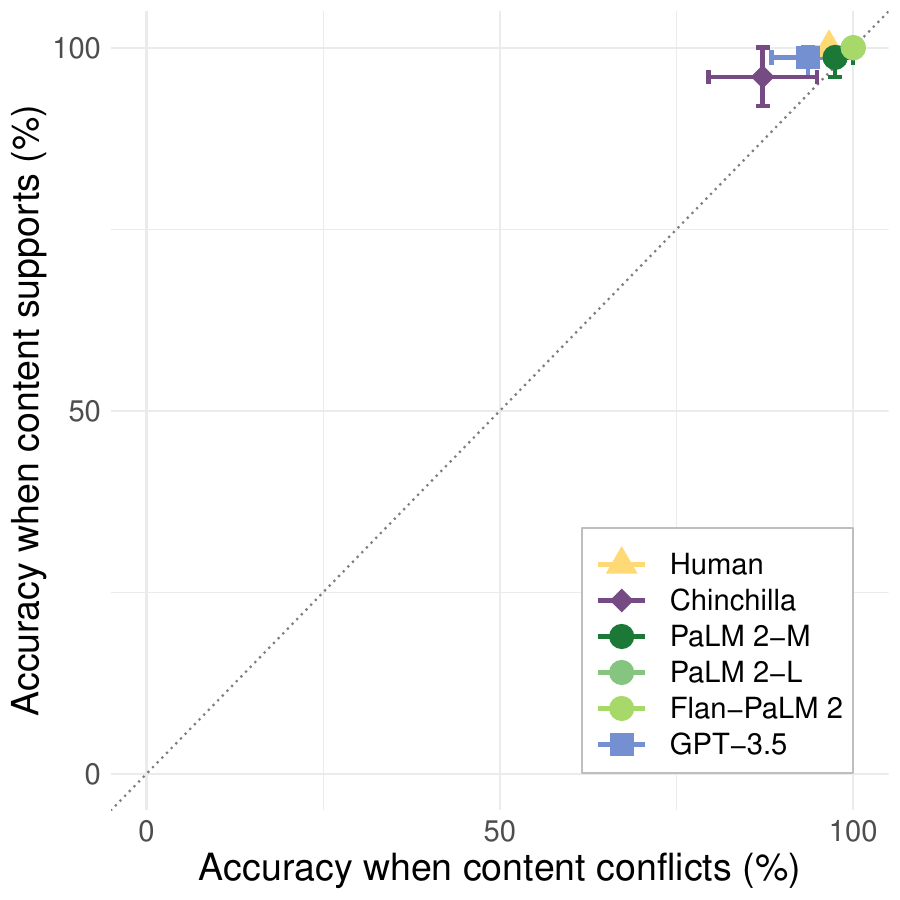}
    \caption{Natural language inference.} \label{fig:summary:nli}
    \end{subfigure}%%
    \begin{subfigure}[t]{0.33\textwidth}
    \includegraphics[width=\textwidth]{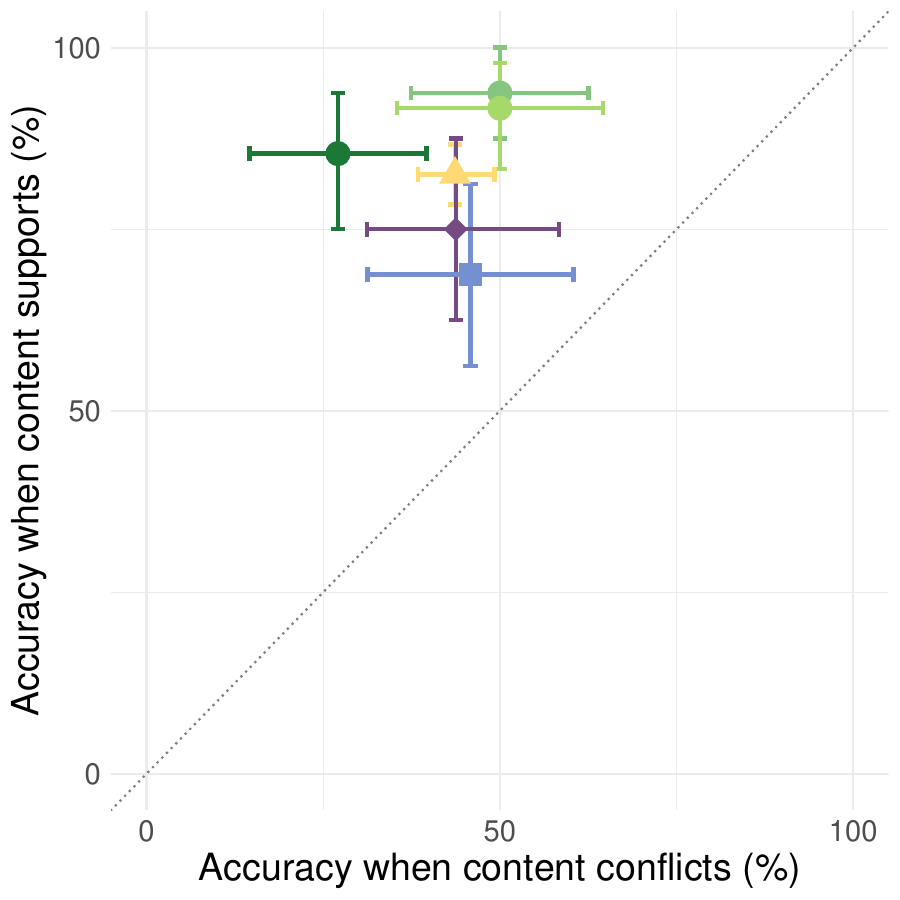}
    \caption{Syllogisms.} \label{fig:summary:syllogisms}
    \end{subfigure}%%
    \begin{subfigure}[t]{0.33\textwidth}
    \includegraphics[width=\textwidth]{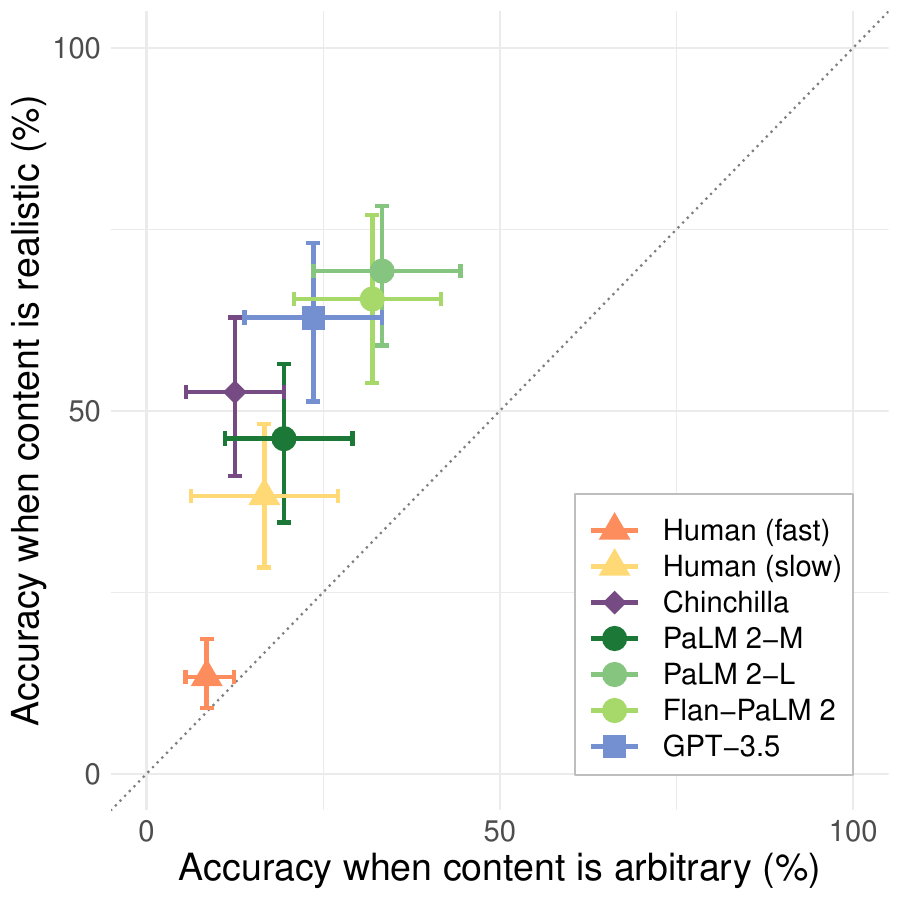}
    \caption{Wason selection task.} \label{fig:summary:wason}
    \end{subfigure}%%
    \caption{Across the three tasks we consider, various language models and humans show similar patterns of overall accuracy and content effects on reasoning. The vertical axis shows accuracy when the content of the problems supports the logical inference. The horizontal axis shows accuracy when the content conflicts (or, in the Wason task, when it is arbitrary). Thus, points above the diagonal indicate an advantage when the content supports the logical inference. (\subref{fig:summary:nli}) On basic natural language inferences, both humans and LMs demonstrate fairly high accuracy across all conditions, and thus relatively little effect of content. (\subref{fig:summary:nli}) When identifying whether syllogisms are logically valid or invalid, both humans and LMs exhibit moderate accuracy, and substantial content effects. (\subref{fig:summary:wason}) On the Wason selection task, the majority of humans show fairly poor performance overall. However, the subset of subjects who are take the longest to answer show somewhat higher accuracy, but primarily on the realistic tasks --- i.e. substantial content effects. On this difficult task, language models generally exceed humans in both accuracy and magnitude of content effects. (Throughout, errorbars are bootstrap 95\%-CIs, and dashed lines are chance performance.)}
    \label{fig:summary}
\end{figure}

We summarize our primary results in Fig. \ref{fig:summary}. In each of our three tasks, humans and models show similar levels of accuracy across conditions. Furthermore, humans and models show similar content effects on each task, which we measure as the degree of advantage when reasoning about logical situations that are consistent with real-world relationships or rules. In the simplest Natural Language Inference task, humans and all models show high accuracy and relatively minor effects of content. When judging the validity of syllogisms, both humans and models show more moderate accuracy, and significant advantages when content supports the logical inference. Finally, on the Wason selection task, humans and models show even lower accuracy, and again substantial content effects. We describe each task, and the corresponding results and analyses, in more detail below. 

\begin{figure}[t]
    \centering
    \includegraphics[width=0.5\textwidth]{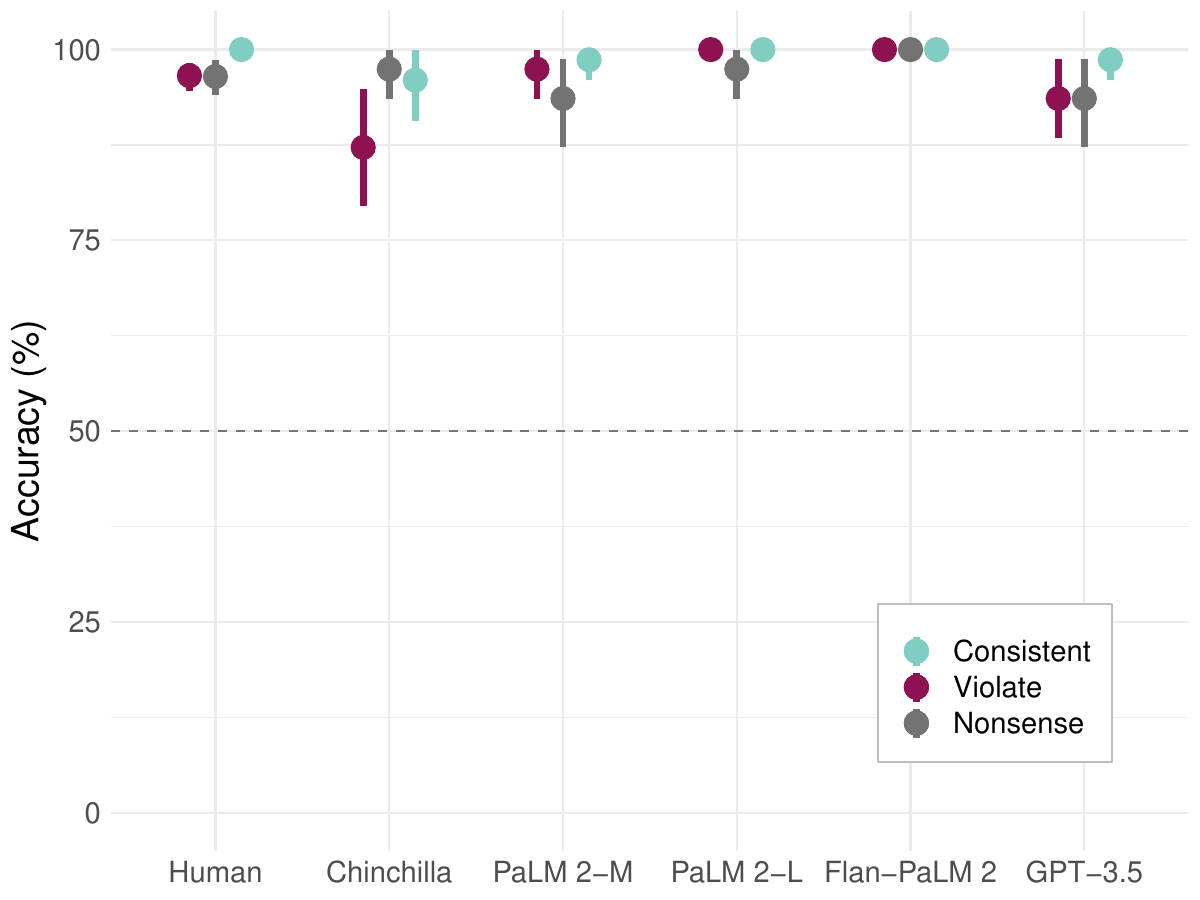}
    \caption{Detailed results on the Natural Language Inference tasks. Both humans (left) and all models show relatively high performance, and relatively little difference in accuracy between belief-consistent and belief-violating inferences, or even nonsense ones.}
    \label{fig:zero_shot:NLI}
\end{figure}

\paragraph{Natural Language Inference}
The relatively simple logical reasoning involved in this task means that both humans and models exhibit high performance, and correspondingly show relatively little effect of content on their reasoning (Fig. \ref{fig:zero_shot:NLI}).  Specifically, we do not detect a statistically-significant effect of content on accuracy in humans or any of the language models in mixed-effects logistic regressions controlling for the random effect of items (or \(\chi^2\) tests where regressions did not converge due to ceiling effects; all \(z < 1.21\) or \(\chi^2 < 0.1\), all \(p > 0.2\); see Appx. \ref{app:statistical_analyses:nli} for full results). However, we do find a statistically significant relationship between human and model accuracy at the item level (\(t(832) = 3.49\), \(p < 0.001\); Appx. \ref{app:fig:item_correlation:nli}) --- even when controlling for condition.
Furthermore, as we discuss below, further investigation into the model confidence shows evidence of content effects on this task as well.

\paragraph{Syllogisms}

\begin{figure}[t]
    \centering
    \includegraphics[width=0.75\textwidth]{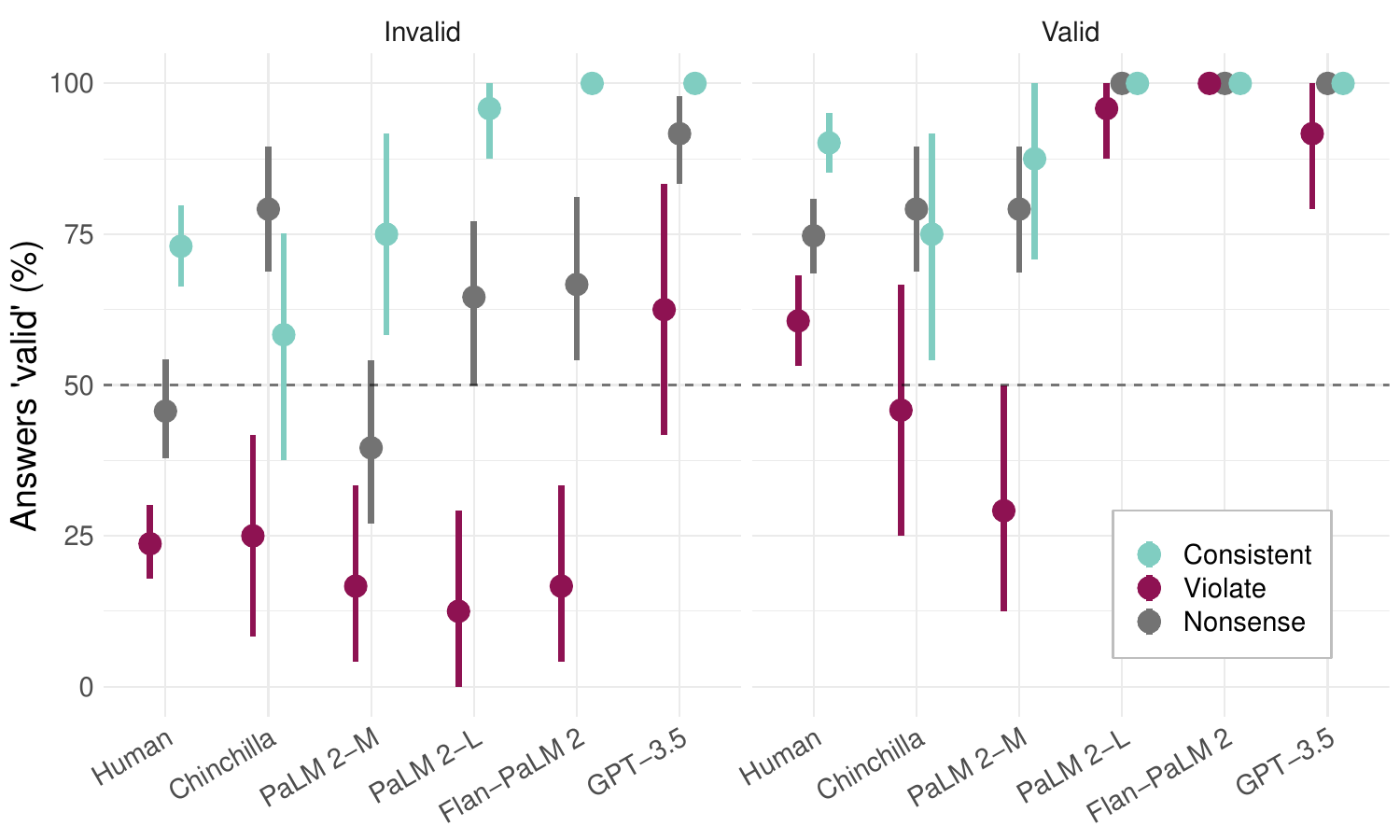}
    \caption{Detailed results on syllogism validity judgements. The vertical axis shows the proportion of the time that each system answers that an argument is valid. Both humans and models exhibit substantial content effects --- they are strongly biased towards saying an argument is valid if the conclusion is consistent with expectations (cyan), and somewhat biased towards saying the argument is invalid if the conclusion violates expectations (maroon). If the argument contains nonsense words (grey), both humans and models show a slight bias towards saying ``valid.''  (Note that this figure plots the proportion of the time the humans or models answer `valid' rather than raw accuracy, to more clearly illustrate the bias. To see accuracy, simply reverse the vertical axis for the invalid arguments.)}
    \label{fig:zero_shot:syllogisms}
\end{figure}

Syllogism validity judgements are significantly more challenging than the NLI task above; correspondingly, we find lower accuracy in both humans and language models. Nevertheless, humans and most language models are sensitive to the logical structure of the task.
However, we find that both humans and language models are strongly affected by the content of the syllogisms  (Fig. \ref{fig:zero_shot:syllogisms}), as in the past literature on syllogistic belief bias in humans \citep{evans1995belief}. 
Specifically, if the semantic content supports the logical inference --- that is, if the conclusion is believable and the argument is valid, or if the conclusion is unbelievable and the argument is invalid --- both humans and all language models tend to answer more accurately (all \(z \geq 2.25\) or \(\chi^2 > 6.39\), all \(p \leq 0.01\); see Appx. \ref{app:statistical_analyses:syllogisms} for full results). 

Two simple effects contribute to this overall content effect --- that belief-consistent conclusions are judged as logically valid and that belief-inconsistent conclusions are judged as logically invalid. As in the past literature, we find that the dominant effect is that humans and models tend to say an argument is valid if the conclusion is belief-consistent, regardless of the actual logical validity. If the conclusion is belief-violating, humans and models do tend to say it is invalid more frequently, but most humans and models are more sensitive to actual logical validity in this case. Specifically, we observe a significant interaction between the content effect and believability in humans, PaLM 2-L, Flan-PaLM 2, and GPT-3.5 (all \(z >5.9\) or \(\chi^2 > 14.3\), all \(p < 0.001\)); but do not observe a significant interaction in Chinchilla or PaLM 2-M (both \(\chi^2 < 0.001\), \(p > 0.99\)). Both humans and models appear to show a slight bias towards saying syllogisms with nonsense words are valid, but again with some sensitivity to the actual logical structure.

Furthermore, even when controlling for condition, we observe a significant correlation between item-level accuracy in humans and language models (\(t(345) = 4.98\), \(p < 0.001\)), suggesting shared patterns in the use of lower-level details of the logic or content.

\begin{figure}[t]
    \centering
    \includegraphics[width=0.66\textwidth]{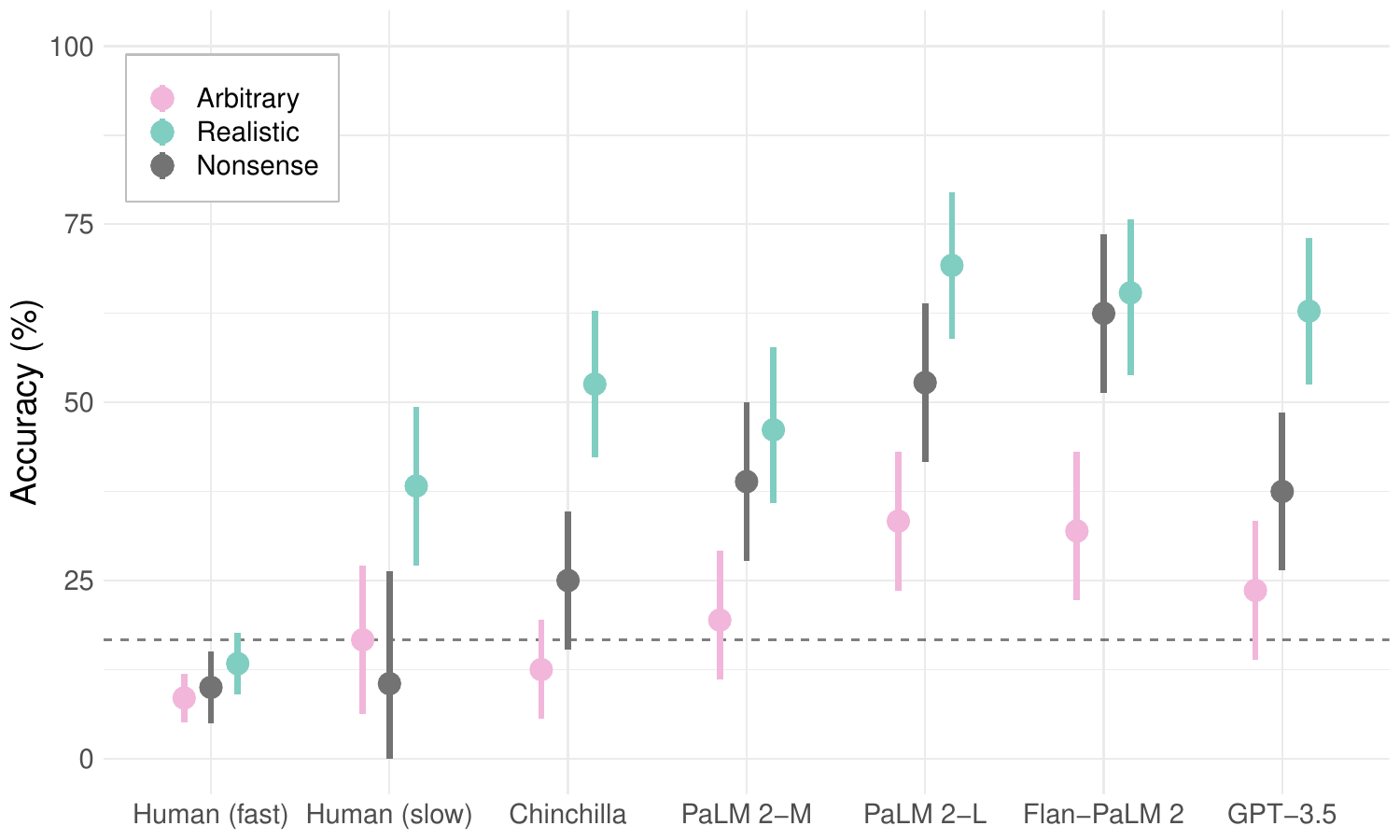}
    \caption{Detailed results on the Wason selection task. Human performance is low, even on the Realistic rules. In particular, the majority of the subjects show at-or-below-chance accuracy in all conditions (though this behavior is not random; see below). However, the subset of subjects who answer more slowly show above chance accuracy for the realistic rules (cyan), but not for the arbitrary ones (pink). This pattern matches the prior results in the cognitive literature. Furthermore, each of the language models reproduces this pattern of advantage for the realistic rules. In addition, two of the larger models perform above chance at the arbitrary rules. (The dashed line corresponds to chance --- a random choice of two cards among the four shown. Both models and humans were forced to choose exactly two cards.)}
    \label{fig:zero_shot:wason}
\end{figure}

\paragraph{The Wason Selection Task}
As in the prior human literature, we found that the Wason task was relatively challenging for humans, as well as for language models (Fig. \ref{fig:zero_shot:wason}). Nevertheless, we observed significant content advantages for the Realistic tasks in humans, and in Chinchilla, PaLM 2-L, and GPT-3.5 (all \(z > 2.2\), all \(p < 0.03\); Appx. \ref{app:statistical_analyses:wason}). We only observed marginally significant advantages of realistic rules in PaLM 2-M and Flan-PaLM 2 (both \(z \geq 1.78\), both \(p \leq 0.08\)), due to stronger item-level effects in these models (though the item-level variance does not seem particularly unusual; see Appx. \ref{app:analyses:wason_items} for further analysis). Intriguingly, some language models also show better performance at the versions of the tasks with Nonsense nouns compared to the Arbitrary ones, though generally Realistic rules are still easier.  We also consider several variations on these rules in Appx.\ \ref{app:fig:wason_shuffled_violate}.

\begin{figure}[t]
    \centering
    \includegraphics[width=0.66\textwidth]{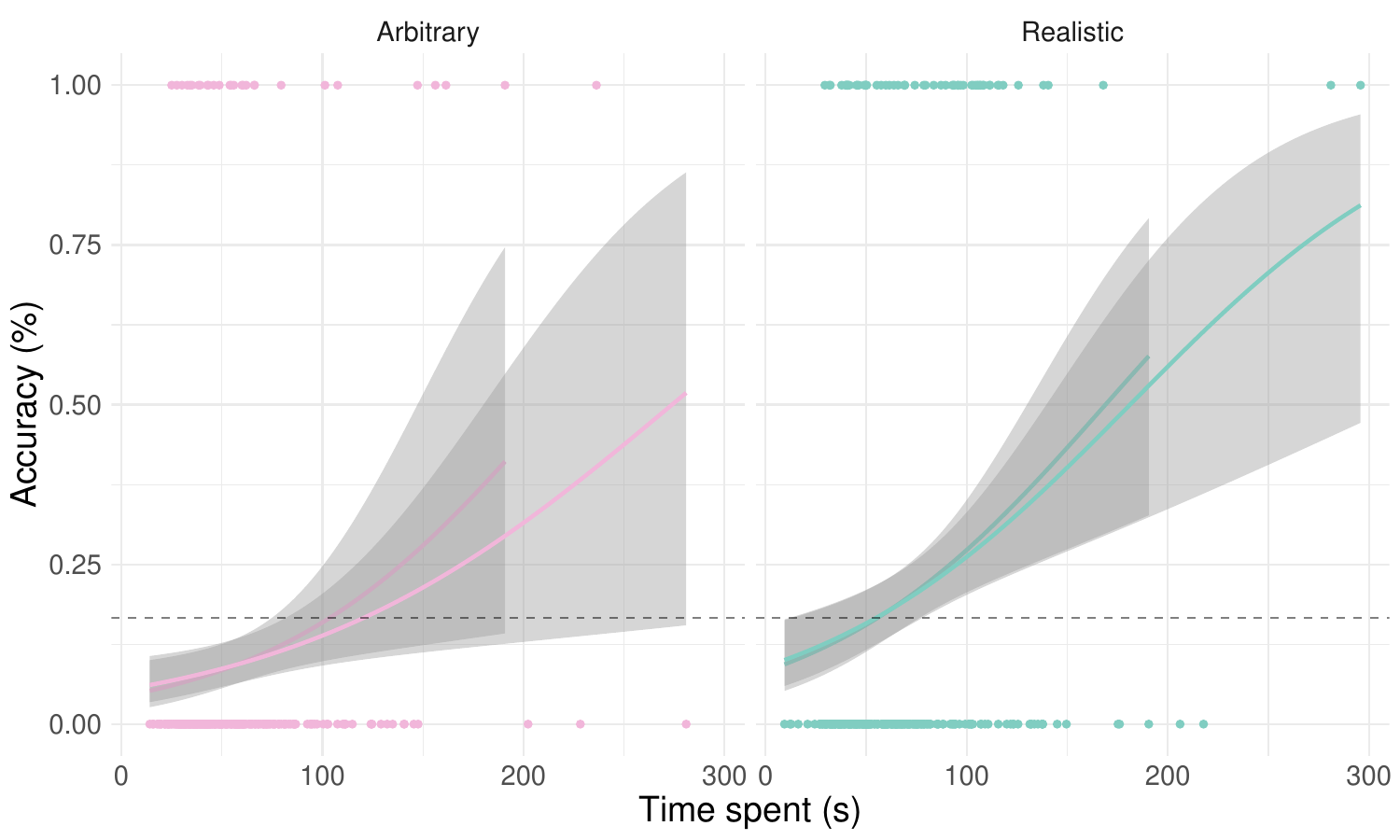}
    \caption{There is a strong relationship between response time and answer accuracy in the Wason tasks; subjects who take longer to answer are more accurate on average. Participants who take sufficiently long to answer perform above chance in the Realistic tasks. There are hints of a similar effect in the Arbitrary condition, but we do not have the power to detect it. (Curves are logistic regression fits, with 95\% CIs. We also plot regressions dropping outliers with time greater than 180 seconds, to show that the effect is not driven solely by outliers.)}
    \label{fig:wason_human_continuous_time}
\end{figure}

Our human participants struggled with this task, as in prior research, and did not achieve significantly higher-than-chance performance overall --- although their behavior is not random, as we discuss below, where we analyze answer choices in more detail. However, spending longer on logical tasks can improve performance \citep{evans2005rapid,evans1994debiasing}, and thus many studies split analyses by response time to isolate participants who spend longer, and therefore show better performance \citep{wickelgren1977speed, wise2005response}. Indeed, we found that human accuracy was significantly associated with response time (\(z = 4.44\), \(p < 0.001\); Appx. \ref{app:statistical_analyses:wason:human}). We depict this relationship in Fig. \ref{fig:wason_human_continuous_time}. To visualize the the performance of discrete subjects in our Figures \ref{fig:summary:wason} and \ref{fig:zero_shot:wason}, we split subjects into `slow' and `fast' groups. The distribution of times taken by subjects is quite skewed, with a long tail. We separate out the top 15\% of subjects that take the longest, who spent more than 80 seconds on the problem, as the slow group. These subjects showed above chance performance in the Realistic condition, but still performed near chance in the other conditions.
We also dig further into the predictive power of human response times in the other tasks in the following sections (and Appx. \ref{app:analyses:nli_syl_rt}).

We collected the data for the Wason task in two different experiments; after observing the lower performance in the first sample, we collected a second sample where we offered a performance bonus for this task. We did not observe significant differences in overall performance or content effects between these subsets, so we collapse across them in the main analyses; however, we present results for each experiment and some additional analyses in Appx. \ref{app:analyses:wason_human_replication}.

\paragraph{Robustness of results to factors like removing instructions, few-shot prompting and scoring methods}
Language model behavior is frequently sensitive to details of the evaluation. Thus, we performed several experiments to confirm that our results were robust to details of the methods used. We present these results in full in Appx.\ \ref{app:analyses:robustness}, but we outline the key experiments here. First, we show that removing the pre-question instructions does not substantially alter the overall results (Appx.\ref{app:analyses:robustness:no_instructions}). Next, we show that our use of the DC-PMI correction for scoring is not the primary driver of content effects (Appx.\ \ref{app:analyses:dcmpi_vs_raw}). On the syllogisms tasks, raw likelihood scoring with the instruction prompt yields strong answer biases --- several models say every argument is valid irrespective of actual logical validity or content. However, the models that don't uniformly say valid show content effects as expected. Furthermore, if the instructions before the question are removed, raw likelihood scoring results in less validity bias, and again strong content effects. For the Wason task, raw likelihood scoring actually improves the accuracy of some models; however, again the content effects are as found with the DC-PMI scoring. Thus, although overall model accuracy and response biases change with uncorrected likelihood scoring, the content effects are similar. Finally, we consider few-shot evaluation, and show that giving few-shot examples yields some mild improvements in accuracy (with greater improvement in the simpler tasks), but does not eliminate the content effects (Appx.\ \ref{app:analyses:robustness:few_shot}). Together, these results suggest that our findings are not strongly driven by idiosyncratic details of our evaluation, and thus support the robustness of our findings.

\paragraph{Variability across different language models}
While we generally find similar content effects across the various models we evaluate, there are some notable differences among them. First, across tasks the larger models tend to be more accurate overall (e.g., comparing the large vs. the medium variants of PaLM 2); however, this does not necessarily mean they show weaker content effects. While it might be expected that instruction-tuning would affect performance, the instruction-tuned models (Flan-PaLM 2 and GPT-3.5-turbo-instruct) do not show consistent differences in overall accuracy or content effects across tasks compared to the base language models---in particular, Flan-PaLM 2 performs quite similarly to PaLM 2-L overall. (However, there are some more notable differences in the distributions of log-probabilities the instruction-tuned models produce; Appx. \ref{app:analyses:answer_logprob_dists}.)

On the syllogisms task in particular, there are some noticeable difference among the models. GPT-3.5, and the larger PaLM 2 models, have quite high sensitivity for identifying valid arguments (they generally correctly identify valid arguments) but relatively less specificity (they also consider several invalid conclusions valid). 
By contrast, PaLM 2-M and Chinchilla models answer more based on content rather than logical validity i.e. regularly judging consistent conclusions as more valid than violating ones, irrespective of their logical validity. The sensitivity to logical structure in the nonsense condition also varies across models -- the PaLM models are fairly sensitive, while GPT 3.5 and Chinchilla both having a strong bias toward answering valid to all nonsense propositions irrespective of actual logical validity.

On the Wason task, the main difference of interest is that the PaLM 2 family of models show generally greater accuracy on the Nonsense problems than the other models do, comparable to their performance on the Realistic condition in some cases.

\subsection*{Model confidence is related to content, correctness, and human response times} \label{sec:results:rt_confidence}

Language models do not produce a single answer; rather, they produce a probability distribution over the possible answers. This distribution can provide further insight into their processing. For example, the probability assigned to the top answers, relative to the others, can be interpreted as a kind of confidence measure. By this measure, language models are often somewhat calibrated, in the sense that the probability they assign to the top answer approximates the probability that their top answer is correct \citep[e.g.][]{kadavath2022language}. Furthermore, human Response Times (RTs) relate to many similar variables, such as confidence, surprisal, or task difficulty; thus, many prior works have related language model confidence to human response or reaction times for linguistic stimuli \citep[e.g.][]{goodkind2018predictive,futrell-etal-2019-neural}.
In this section, we correspondingly analyze how the language model confidence relates to the task content and logic, the correctness of answers, and the human response times.

\begin{figure}[t]
    \centering
    \begin{subfigure}[t]{0.41\textwidth}
    \includegraphics[width=\textwidth]{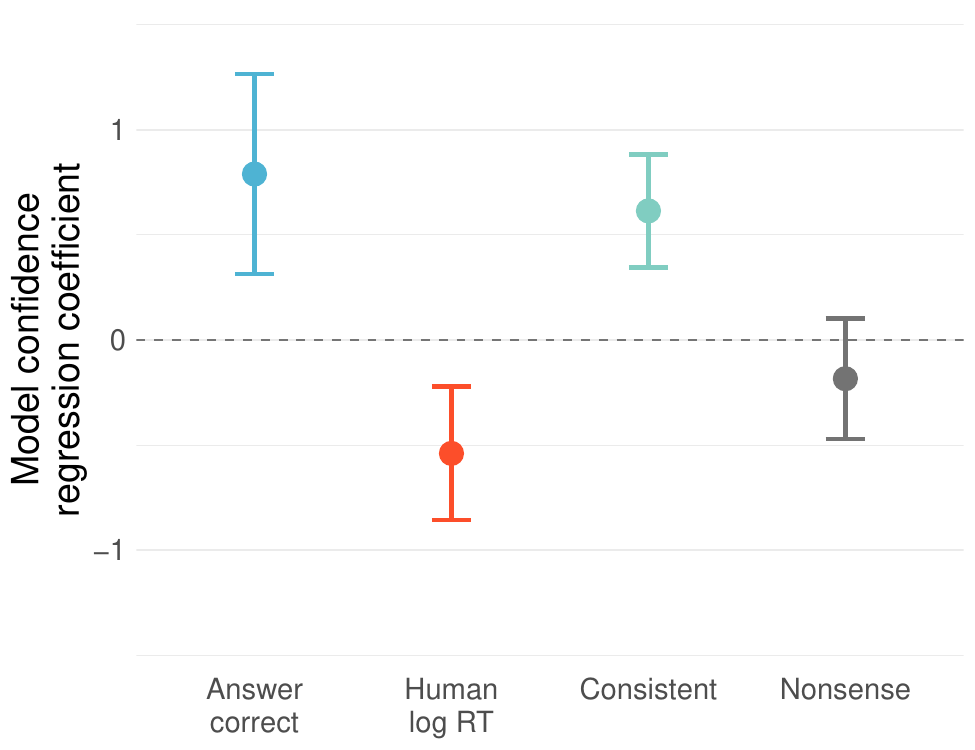}
    \caption{Natural language inference.} \label{fig:rt_logprob:nli_reg}
    \end{subfigure}%
    \begin{subfigure}[t]{0.473\textwidth}
    \includegraphics[width=\textwidth]{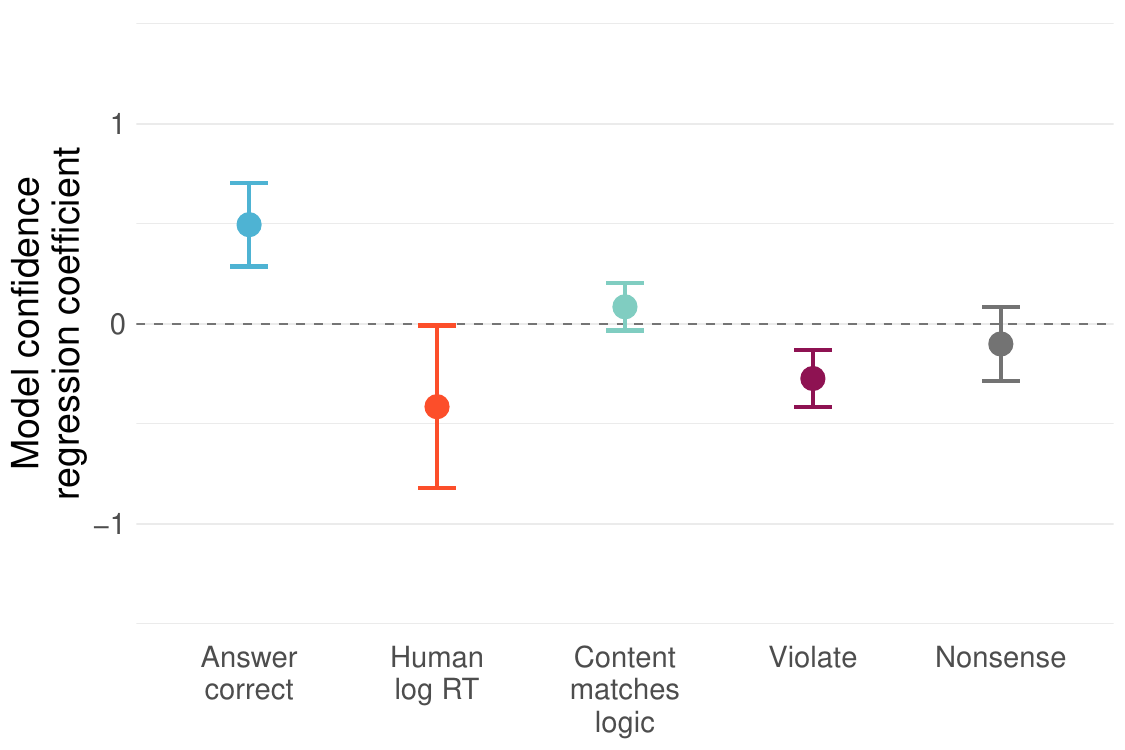}
    \caption{Syllogisms.} \label{fig:rt_logprob:syl_reg}
    \end{subfigure}\\
    \begin{subfigure}[t]{0.5\textwidth}
    \includegraphics[width=\textwidth]{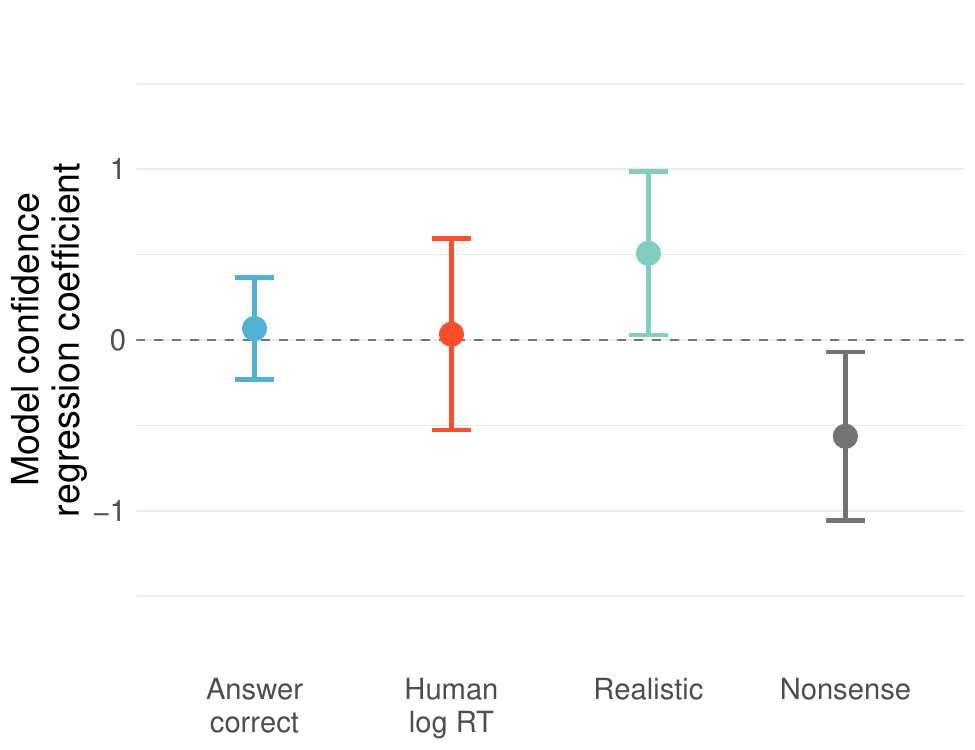}
    \caption{Wason selection task.} \label{fig:rt_logprob:wason_reg}
    \end{subfigure}
    \caption{Language model confidence---as measured by the difference in(prior-corrected) log-probability between the chosen answer and the next most probable---is associated with correct answers, task variables, and human average response times. (\subref{fig:rt_logprob:nli_reg}-\subref{fig:rt_logprob:syl_reg}) On the NLI and syllogism tasks, models are generally more confident in correct answers and belief-consistent conditions, less confident in belief-violating conditions, and less confident on problems that humans take longer to answer.  (\subref{fig:rt_logprob:wason_reg}) On the Wason task, effects are weaker. Human RT and correct answers are not associated with confidence; however, the models do show more confidence on Realistic problems, and less on Nonsense ones. (Effects are calculated from a mixed-effects regression predicting the difference in log-probability between the top and second-highest answer, z-scored within each model, and controlling for all other significant predictors. Errorbars are parametric 95\%-CIs. Note that human RT is calculated across all human subjects for the Wason task, not just slow subjects.)}
    \label{fig:rt_logprob_regression}
\end{figure}

\begin{figure}[t]
    \centering
    \begin{subfigure}[t]{\textwidth}
    \includegraphics[width=\textwidth]{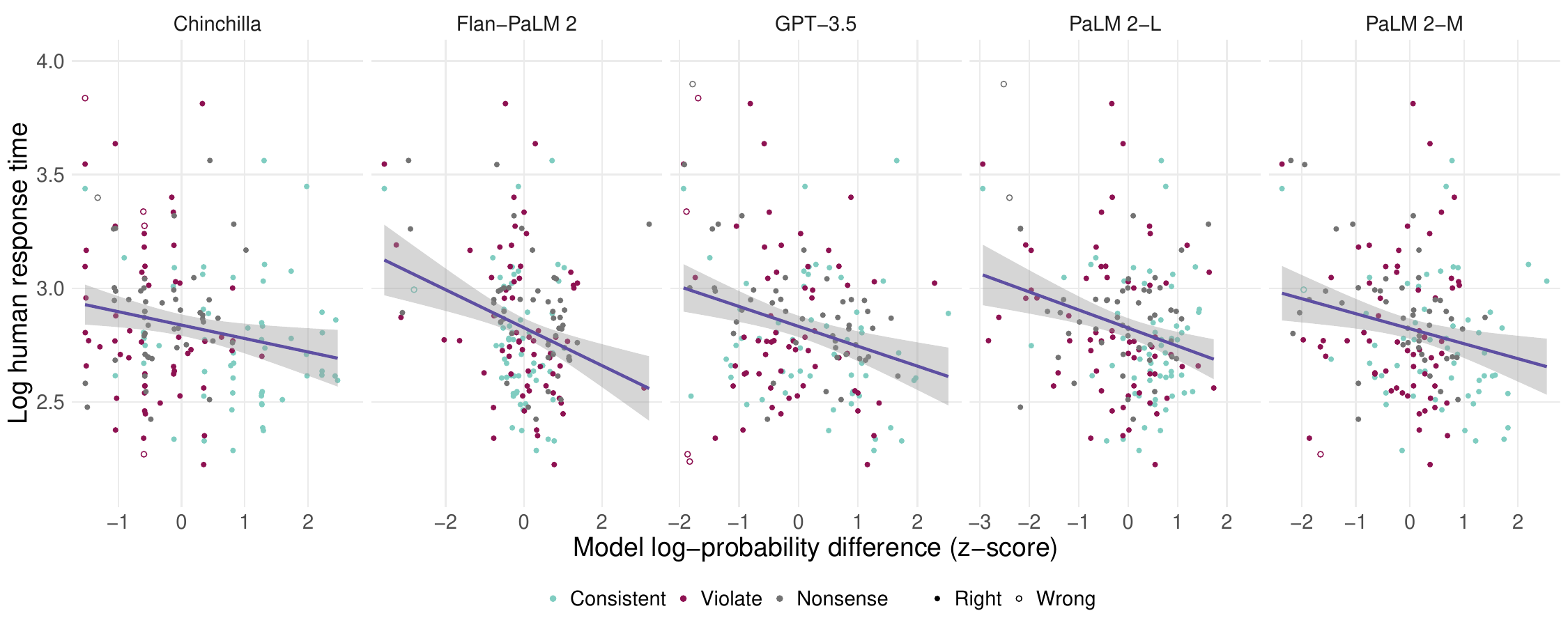}
    \caption{Natural language inference raw results.} \label{fig:rt_logprob:nli}
    \end{subfigure}\\
    \begin{subfigure}[t]{\textwidth}
    \includegraphics[width=\textwidth]{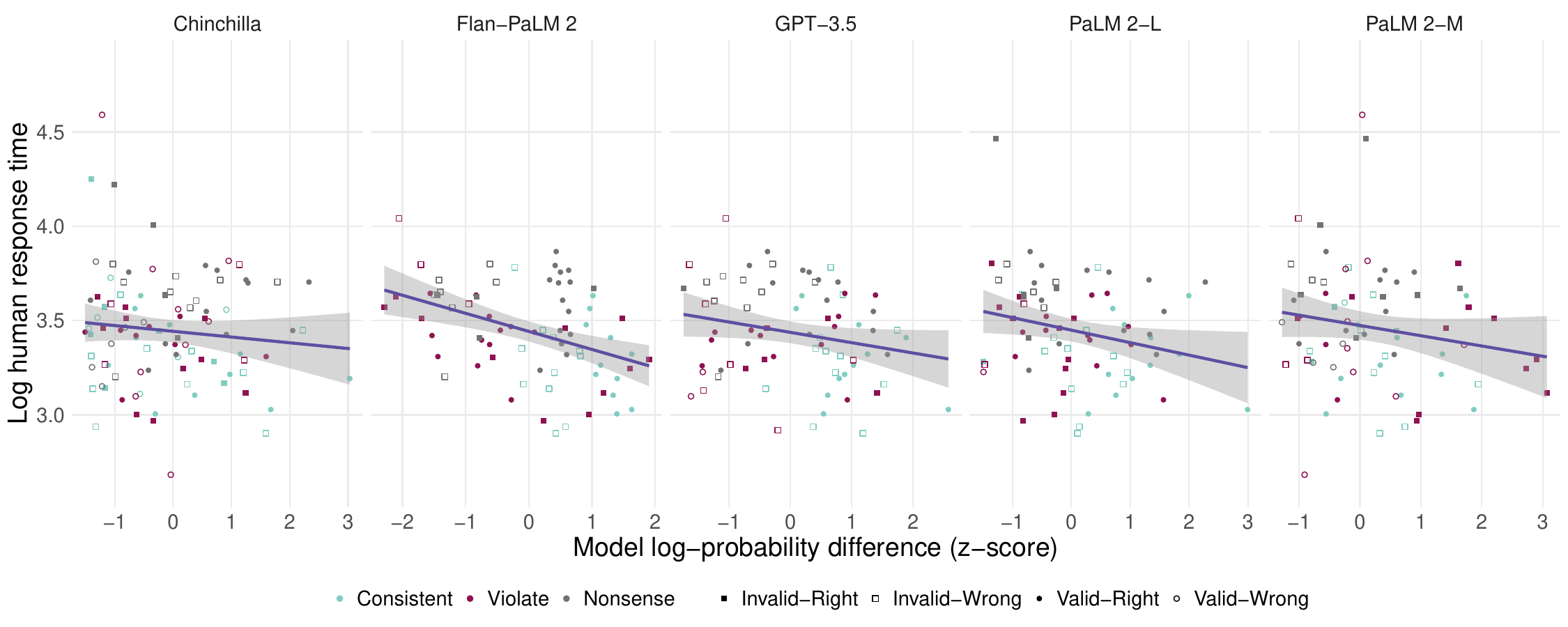}
    \caption{Syllogisms raw results.} \label{fig:rt_logprob:syllogisms}
    \end{subfigure}\\%%
    \caption{Human response times are generally negatively related to model confidence (measured as the difference in log-probabilities between the correct answer and the incorrect answer). That is, on problems for which the model displays greater confidence, humans tend to respond more quickly. This relationship holds on both (\subref{fig:rt_logprob:nli}) the NLI tasks, and (\subref{fig:rt_logprob:syllogisms}) the syllogism tasks. (Points show average response times for individual problems, broken down by whether the humans or models answered correctly or not; see Appx. \ref{app:statistical_analyses:logprob_rt} for details.)}
    \label{fig:rt_logprob_correlation}
\end{figure}

We summarize these results in Fig. \ref{fig:rt_logprob_regression}. We measure model confidence as the difference in prior-corrected log-probability between the top answer and the second highest---thus, if the model is almost undecided between several answers, this confidence measure will be low, while if the model is placing almost all its probability mass on a single answer, the confidence measure will be high.  In mixed-effects regressions predicting model confidence from task variables and average human RTs on the same problem, we find a variety of interesting effects. First, language models tend to be more confident on correct answers (that is, they are somewhat calibrated). Task variables also affect confidence; models are generally less confident when the conclusion violates beliefs, and more confident for the realistic rules on the Wason task.%
Furthermore, even when controlling for task variables and accuracy, there is a statistically-significant negative association with human response times on the NLI and syllogisms tasks (respectively \(t(655) = -3.39\), \(p < 0.001\); and \(t(353) = -2.03\), \(p < 0.05\); Appx. \ref{app:statistical_analyses:logprob_rt})---that is, models tend to show more confidence on problems where humans likewise respond more rapidly. We visualize this relationship in Fig. \ref{fig:rt_logprob_correlation}. 

\subsection*{Analyzing components of the Wason responses} \label{sec:results:wason_components}

\begin{figure}[htbp]
    \centering
    \includegraphics[width=0.66\textwidth]{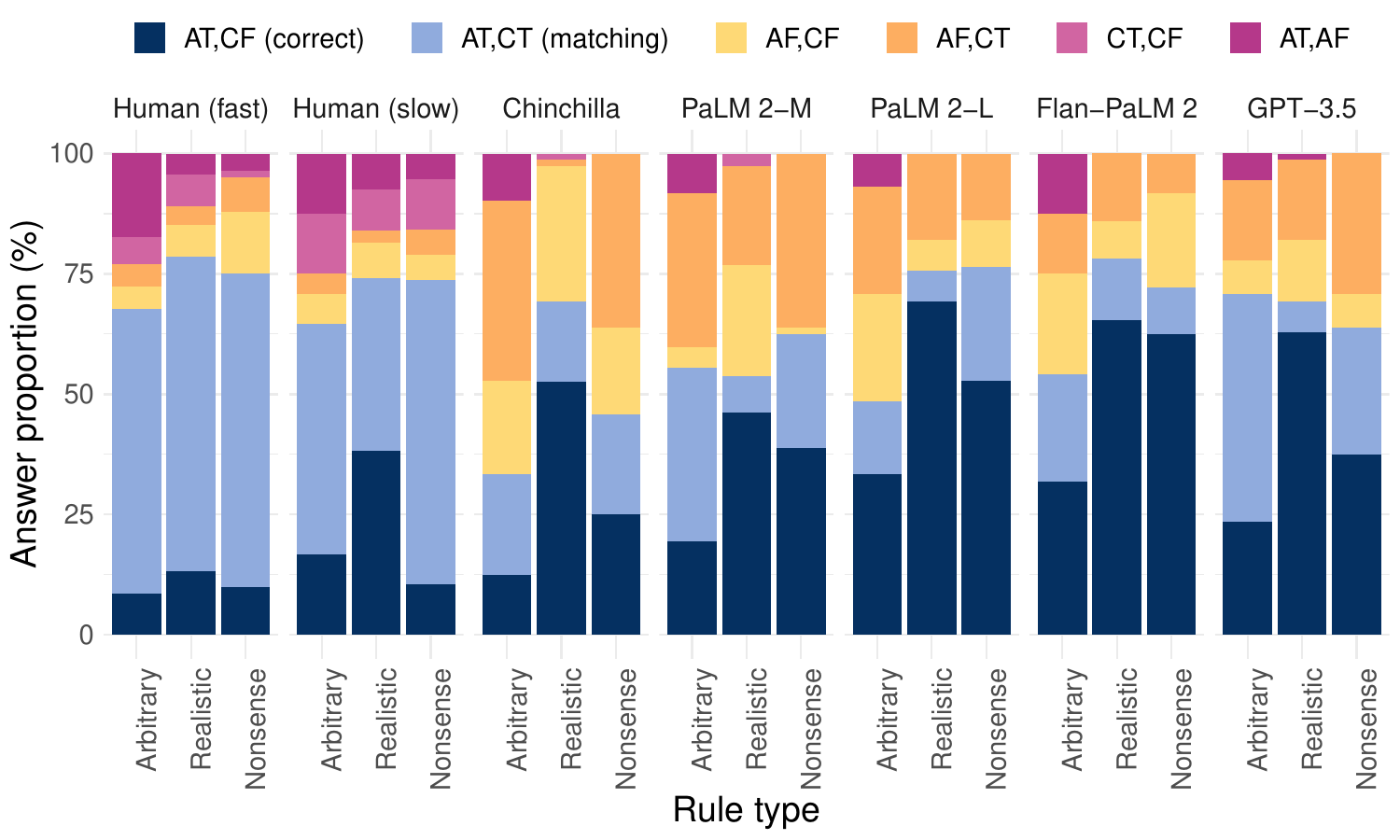}
    \caption{Answer patterns for the Wason tasks, broken down into components: the pairings of individual cards that each participant chose (AT = Antecedent True, CF = Consequent False, etc.). Behavior is not random, even when performance is near chance. As above, humans do not consistently choose the correct answer (AT, CF; dark blue); instead, humans more frequently exhibit the matching bias (AT, CT; light blue). Humans also show other errors, however, including a surprisingly high rate of choosing two cards corresponding to a single rule component (dark/light pink). Language models answer correctly more often than humans, but intriguingly choose options with the antecedent false and a consequent card (yellow/orange) more frequently. (Note that all participants and language models were forced to choose exactly two cards.)}
    \label{fig:wason_choice_details}
\end{figure}

Because each answer to the Wason problems involves selecting a pair of cards, we further analyzed the individual cards chosen. The card options presented each problem are designed so that two cards respectively match and violate the antecedent, and similarly for the consequent.  The correct answer is to choose one card for the antecedent and one for the consequent; more precisely, the card for which the antecedent is true (AT), and the card for which the consequent is false (CF). In Fig. \ref{fig:wason_choice_details} we examine human and model choices; we quantitatively analyze these choices using a multinomial logistic regression model in Appx. \ref{app:statistical_analyses:wason:answer_choices_multinomial}. 

 Even in conditions when performance is close to chance, behavior is generally not random. As in prior work, humans do not consistently choose the correct answer (AT, CF). Instead, humans tend to exhibit the matching bias; that is, they tend to choose each of the two cards that match each component of the rule (AT, CT). However, in the Realistic condition, slow humans answer correctly somewhat more frequently. Humans also exhibit errors besides the matching bias; including an increased rate of choosing the two cards corresponding to a single component of the rule --- either both antecedent cards, or both consequent cards. Language models tend to give more correct responses than humans, and to show facilitation in the realistic rules compared to arbitrary ones. Relative to humans, language models show fewer matching errors, fewer errors of choosing two cards from the same rule component, but more errors of choosing the antecedent false options. These differences in error patterns may indicate differences between the response processes engaged by the models and humans. (Note, however, that while the models accuracies do not change too substantially with alternate scoring methods, the particular errors the models make are somewhat sensitive to scoring method --- without the DC-PMI correction the model errors more closely approximate the human ones in some cases; Appx. \ref{app:analyses:wason_choices_scoring_order}.)

\FloatBarrier
\section*{Discussion}

Humans are imperfect reasoners. We reason most effectively about entities and situations that are consistent with our understanding of the world. Even in these familiar cases, we often make mistakes. Our experiments show that language models mirror these patterns of behavior. Language models likewise perform imperfectly on logical reasoning tasks, but this performance depends on content and context. Most notably, such models often fail in situations where humans fail --- when stimuli become too abstract or conflict with prior expectations about the world. 

Beyond these simple parallels in accuracy across different conditions and items, we also observed more subtle parallels in language model confidence. The model's confidence tends to be higher for correct answers, and for cases where prior expectations about the content are consistent with the logical structure. Even when controlling for these effects, model confidence is related to human response times. Thus, language models reflect human content effects on reasoning at multiple levels. Furthermore, these core results are generally robust across different language models with different training and tuning paradigms, different prompts, etc., suggesting that they are a fairly general phenomenon of predictive models that learn from human-generated text.

\paragraph{Prior research on language model reasoning.}
Since Brown et al. \citep{brown2020language} showed that large language models could perform moderately well on some reasoning tasks, there has been a growing interest in language model reasoning \citep{binz2022using}. Typical methods focus on prompting for sequential reasoning \citep{nye2021show,wei2022chain,kojima2022large}, altering task framing~\citep{khashabi2022reframing,lampinen2022can} or iteratively sampling answers~\citep{wang2022self}. 

In response, some researchers have questioned whether these language model abilities qualify as ``reasoning''. The fact that language models sometimes rely on ``simple heuristics'' \citep{patel2021nlp}, or reason more accurately about frequently-occurring numbers \citep{razeghi2022impact}, have been cited to ``rais[e] questions on the extent to which these models are \emph{actually reasoning}'' (ibid, emphasis ours). The implicit assumption in these critiques is that reasoning should be a purely algebraic, syntactic computations over symbols from which ``all meaning had been purged'' (\citealp{newell1980physical}; cf. \citealp{marcus2003algebraic}). In this work, we emphasize how \emph{both} humans and language models rely on content when answering  reasoning problems --- using simple heuristics in some contexts, and answering more accurately about frequently-occurring situations \citep{mercier2017enigma,dasgupta2020theory}. Thus, abstract reasoning may be a graded, content-sensitive capacity in both humans and models.

\paragraph{Dual systems?} The idea that humans possess dual reasoning systems --- an implicit, intuitive system ``system 1'', and an explicit reasoning ``system 2' --- was motivated in large part by belief bias and Wason task effects \citep{evans1984heuristic,evans2003two,oaksford2003optimal}.
The dual system idea has more recently become popular \citep{kahneman2011thinking,evans2013rationality}, including in machine learning \citep[e.g.][]{bengio2017consciousness}. It is often claimed that current ML (including large language models) behave like system 1, and that we need to augment this with a classically-symbolic process to get system 2 behaviour \citep[e.g.][]{nye2021improving}. These calls to action usually advocate for an explicit duality; with a neural network based system providing the system 1 and a system with more explicit symbolic or otherwise structured system being the system 2.

Our results show that a unitary system --- a large transformer language model --- can mirror this dual behavior in humans, demonstrating both biased and consistent reasoning depending on the context and task difficulty. In the NLI tasks, a few examples takes Chinchilla from highly content-biased performance to near ceiling performance, and even a simple instructional prompt can substantially reduce bias. These findings integrate with prior works showing that language models can be prompted to exhibit sequential reasoning, and thereby improve their performance in domains like mathematics \citep{nye2021show,wei2022chain,kojima2022large}.

These observations suggest the possibility that the unitary language model may have implicitly learned a context-dependent control mechanism that arbitrates between conflicting responses (such as more intuitive answers vs. logically correct ones). This perspective suggests several possible directions for future research. First, it would be interesting to seek out mechanistic evidence of such as conflict-arbitration process within language models.
Furthermore, it suggests that augmenting language models with a second system might not be necessary to achieve relatively reliable performance. Instead, it might be sufficient to further develop the control mechanisms within these models by altering their context and training, as we discuss below.

From a human cognitive neuroscience perspective, these issues are more complex. The idea of context-dependent arbitration between conflicting responses has been influential in the literature on human cognitive control \citep{cohen1990control, botvinick2014computational}, and has been implicated in humans reasoning successfully in tasks that require following novel, arbitrary reasoning procedures or over-riding pre-existing response tendencies \citep{duncan2020integrated,li2022weighted}. However, these control processes are generally believed to principally reside in frontal regions outside the language areas, or in a network of control-related brain areas that interface with the language regions and other domain-specific brain areas but is not housed therein \citep{duncan2020integrated}. Thus, understanding the full detail of human cognition in such language-based logical tasks may require incorporating a control network into the architecture more explicitly. Nevertheless, our results with language models suggest that this controller could be more intertwined with the statistical inference system than it would be in a classic dual-systems model; moreover, that the controller does not need to be implemented as a classical symbol system to achieve human-competitive logical reasoning performance.

\paragraph{Neural mechanisms of human reasoning.} Deep learning models are increasingly used as models of neural processing in biological systems \citep[e.g.][]{yamins2014performance,yamins2016using}, as they often develop similar patterns of representation. These findings have led to proposals that deep learning models capture \emph{mechanistic} details of neural processing at an appropriate level of description \citep{cao2021explanatory,cao2021explanatoryb}, despite the fact that aspects of their information processing clearly differ from biological systems. More recently, large language models have been similarly shown to accurately predict neural representations in the human language system --- large language models ``predict nearly 100\% of the explainable variance in neural responses to sentences'' (\citealp{schrimpf2021neural}; see also \citealp{kumar2022reconstructing, goldstein2022shared}). Language models also predict low-level behavioral phenomena; e.g. surprisal predicts reading time \citep{goodkind2018predictive,wilcox2020predictive}.
In the context of these works, our observation of behavioral similarities in reasoning patterns between humans and language models raise important questions about possible similarities of the underlying reasoning processes between humans and language models, and the extent of overlap between neural mechanisms for language and reasoning in humans. This is particularly exciting because neural models for these phenomena in humans are currently lacking or incomplete. Indeed, even prior high-level explanations of these phenomena have often focused on only a single task, such as explaining \emph{only} the Wason task content effects with appeals to evolved social-reasoning mechanisms \citep{cosmides1989logic}. Our results suggest that there could be a more general explanation.

\paragraph{Towards a normative account of content effects?}
Various accounts of human cognitive biases frame them as `normative' according to some objective. Some explain biases as the application of processes --- such as information gathering or pragmatics --- that are broadly rational under a different model of the world \citep[e.g.][]{oaksford2003optimal,tessler2022logic}. Others interpret them as a rational adaptation to reasoning under constraints such as limited memory or time \citep[e.g.][]{lieder2020resource, gershman2015computational, simon1990bounded} --- where content effects actually support fast and effective reasoning in commonly encountered tasks \citep{mercier2017enigma,dasgupta2020theory}.
Our results show that content effects can emerge from simply training a large transformer to imitate language produced by human culture, without explicitly incorporating any human-specific internal mechanisms.

This observation suggests two possible origins for these content effects. First, the content effects could be directly learned from the humans that generated the data used to train the language models. Under this hypothesis, poor logical inferences about nonsense or belief-violating premises come from \textit{copying} the incorrect inferences made by humans about these premises. Since humans also learn substantially from other humans and the cultures in which we are immersed, it is plausible that both humans and language models could acquire some of these reasoning patterns by imitation.

The other possibility is that (like humans) the model's exposure to the world reflects semantic truths and beliefs and that language models and humans both converge on these content biases that reflect this semantic content for more task-oriented reasons: because it helps humans to draw more accurate inferences in the situations they encounter (which are mostly familiar and believable), and helps language models to more accurately predict the (mostly believable) text that they encounter.
In either case, humans and models acquire surprisingly similar patterns of behavior, from seemingly very different architectures, experiences, and training objectives. A promising direction for future enquiry would be to causally manipulate features of language model's training objective and experience, to explore which features contribute to the emergence of content biases in language models. These investigations could offer insights into the origins of human patterns of reasoning, and into what data we should use to train language models.

\paragraph{Why might model response patterns differ from human ones?} The language model response patterns do not perfectly match all aspects of the human data. For example, on the Wason task several models outperform humans on the Nonsense condition, and the error patterns on the Wason tasks are somewhat different than those observed in humans (although human error patterns also vary across populations; \citealp{inglis2004mathematicians,cresswell2020does}). Similarly, not all models show the significant interaction between believability and validity on the syllogism tasks that humans do \citep{evans1983conflict}, although it is present in most models (and the human interaction similarly may not appear in all cases; \citealp{dube2010assessing}). Various factors could contribute to differences between model and human behaviors.

First, while we attempted to align our evaluation of humans and models as closely as possible \citep[cf.][]{lampinen2022can}, it is difficult to do so perfectly. In some cases, such as the Wason task, differences in the form of the answer are unavoidable --- humans had to select answers individually by clicking on cards to select them, and then clicking continue, while models had to jointly output both answers in text, without a chance to revise their answer before continuing. Moreover, it is difficult to know how to prompt a language model in order to evaluate a particular task. Language model training blends many tasks into a homogeneous soup, which makes controlling the model difficult. For example, presenting task instructions might not actually lead to better performance \citep[cf.][]{webson2021prompt}. Similarly, presenting negative examples can help humans learn, but is generally detrimental to model performance \citep[e.g.][]{mishra2021cross} --- presumably because the model infers that the task is to sometimes output wrong answers, while humans might understand the communicative intent behind the use of negative examples. Thus, while we tried to match instructions between humans and models, it is possible that idiosyncratic details of our task framing may have caused the model to infer the task incorrectly. To minimize this risk, we tried various different prompting strategies, and where we varied these details we generally observed similar overall effects. Nevertheless, it is possible that some aspect of the problem instructions or framing contributes to the response patterns.%

More fundamentally, language models do not directly experience the situations to which language refers \citep{mcclelland2020placing}; grounded experience (for instance the capacity to simulate the physical turning of cards on a table) presumably underpins some human beliefs and reasoning. Furthermore, humans sometimes use physical or motor processes such as gesture to support logical reasoning \citep{alibali2014gesture,nathan2020embodied}. Finally, language models experience language passively, while humans experience language as an active, conventional system for social communication \citep[e.g.][]{clark1996using}; active participation may be key to understanding meaning as humans do \citep{santoro2021symbolic,schlangen2022norm}. Some differences between language models and humans may therefore stem from differences between the rich, grounded, interactive experience of humans and the impoverished experience of the models.

\paragraph{How can we achieve more abstract, context-independent reasoning?}
If language models exhibit some of the same reasoning biases as humans could some of the factors that reduce content dependency in human reasoning be applied to make these models less content-dependent? In humans, formal education is associated with an improved ability to reason logically and consistently  \citep{luria1971towards,lehman1990longitudinal,attridge2016does,inglis2004mathematicians,cresswell2020does,nam2021underlies}. However, causal evidence is scarce, because years of education are difficult to experimentally manipulate; thus the association may be partly due to selection effects, e.g. continuing in formal education might be more likely in individuals with stronger prior abilities. Nevertheless, the association with formal education raises an intriguing question: could language models learn to reason more reliably with targeted formal education?

Several recent results suggest that this may indeed be a promising direction. Pretraining on synthetic logical reasoning tasks can improve model performance on reasoning and mathematics problems \citep{clark2020transformers,wu2021lime}. In some cases language models can either be prompted or can learn to verify, correct, or debias their own outputs \citep{schick2021self,cobbe2021training,saunders2022self,kadavath2022language}. Finally, language model reasoning can be bootstrapped through iterated fine-tuning on successful instances \citep{zelikman2022star}. These results suggest the possibility that a model trained with instructions to perform logical reasoning, and to check and correct the results of its work, might move closer to the logical reasoning capabilities of formally-educated humans. Perhaps logical reasoning is a graded competency that is supported by a range of different environmental and educational factors \citep{santoro2021symbolic, wang2021meta}, rather than a core ability that must be built in to an intelligent system.

\paragraph{Limitations}
In addition to the limitations noted above --- such as the challenges of perfectly aligning comparisons between humans and language models --- there are several other limitations to our work. First, our human participants exhibited relatively low performance on the Wason task. However, as noted above, there are well-known individual differences in these effects that are associated with factors like depth of mathematical education. We were unfortunately unable to examine these effects in our data, but in future work it would be interesting to explicitly explore how educational factors affect performance on the more challenging Wason conditions, as well as more general patterns like the relationship between model confidence and human response time. Furthermore while our experiments suggest that content effects in reasoning can emerge from predictive learning on naturalistic data, they do not ascertain precisely which aspects of the large language model training datasets contribute to this learning. Other research has used controlled training data distributions to systematically investigate the origin of language model capabilities \citep{chan2022data,prystawski2023think}; it would be an interesting future direction to apply analogous methods to investigate the origin of content effects.

\section*{Materials and Methods}

%\paragraph{Experimental design}

\paragraph{Creating datasets}
While many of these tasks have been extensively studied in cognitive science, the stimuli used in cognitive experiments are often online in articles and course materials, and thus may be present in the training data of large language models, which could compromise results \citep[e.g.][]{emami2020analysis,dodge-etal-2021-documenting}. To reduce these concerns, we generate new datasets, by following the design approaches used in prior work. We briefly outline this process here; see Appx.\ \ref{app:methods:datasets} for full details.

For each of the three tasks above, we generate multiple versions of the task stimuli. Throughout, the logical structure of the stimuli remains fixed, we simply manipulate the entities over which this logic operates (Fig. \ref{fig:manipulating_structure}). We generate propositions that are:\\ \textbf{\textit{Consistent}} with human beliefs and knowledge (e.g. ants are smaller than whales).\\
\textbf{\textit{Violate}} beliefs by inverting the consistent statements (e.g. whales are smaller than ants).\\
\textbf{\textit{Nonsense}} tasks about which the model should not have strong beliefs, by swapping the entities out for nonsense words (e.g. kleegs are smaller than feps). 

For the Wason tasks, we slightly alter our approach to fit the different character of the tasks. We generate questions with:\\
\textbf{\textit{Realistic}} rules involving plausible relationships (e.g. ``if the passengers are traveling outside the US, then they must have shown a passport'').\\
\textbf{\textit{Arbitrary}} rules (e.g. ``if the cards have a plural word, then they have a positive emotion'').\\
\textbf{\textit{Nonsense}} rules relating nonsense words (``if the cards have more bem, then they have less stope''). Note that for the Wason task, this change alters the kinds of inferences that need to be made; while for the basic Wason task, matching each card to the antecedent or consequent is nontrivial (e.g. realizing that ``shoes'' is a plural word, not singular), it is difficult to match these inferences with Nonsense words that have no prior associations. As shown in the examples, we use a format where the cards either have more or less of a nonsense attribute, which makes the inferences perhaps more direct than other conditions (although models perform similarly on the basic inferences across conditions; Appx.\ \ref{app:fig:wason_premise_match}).\\

In Appx.\ \ref{app:analyses:human priors} we validate the semantic content of our datasets, by showing that participants find the propositions and rules from our Consistent and Realistic stimuli much more plausible than those from other conditions.

We attempted to create these datasets in a way that could be presented to the humans and language models in precisely the same manner (for example, prefacing the problems with the same instructions for both the humans and the models).\footnote{N.B. this required adapting some of the problem formats and prompts compared to an earlier preprint of this paper that did not evaluate humans; see Appx.\ \ref{app:methods:datasets:differences_from_previous_version}.} 

\paragraph{Models \& evaluation}
We evaluate several different families of language models. First, we evaluate several base LMs that are trained only on language modeling: including Chinchilla \citep{hoffmann2022training} a large model (with 70 billion parameters) trained on causal language modeling, and PaLM 2-M and -L \citep{anil2023palm}, which are trained on a mixture of language modeling and infilling objectives \citep{tay2022ul2}. 
We also evaluate two instruction-tuned models: Flan-PaLM 2 (an instruction-tuned version of Palm 2-L), and GPT-3.5-turbo-instruct \citep{gpt35}, which we generally refer to as GPT-3.5 for brevity.\footnote{We fortuitously performed this evaluation during the short window of time in which scoring was available on GPT-3.5-turbo-instruct.} We observe broadly similar content effects across all types of models, suggesting that these effects are not too strongly affected by the particular training objective, or by standard instruction-tuning. 

For each task, we present the model with brief instructions that approximate the relevant portions of the human instructions.
We then present the question, which ends with ``Answer:'' and assess the model by evaluating the likelihood of continuing this prompt with each of a set of possible answers. We apply the DC-PMI correction proposed by Holtzman et al. \citep{holtzman2021surface} --- i.e., we measure the change in likelihood of each answer in the context of the question relative to a baseline context, and choosing the answer that has the largest increase in likelihood in context. This scoring approach is intended to reduce the possibility that the model would simply phrase the answer differently than the available choices; for example, answering ``this is not a valid argument'' rather than ``this argument is invalid''. This approach can also be interpreted as correcting for the prior over utterances. For the NLI task, however, the direct answer format means that the DC-PMI correction would therefore control for the very bias we are trying to measure. Thus, for the NLI task we simply choose the answer that receives the maximum likelihood among the set of possible answers.
We also report syllogism and Wason results with maximum likelihood scoring in Appx.\ \ref{app:analyses:dcmpi_vs_raw}; while overall accuracy changes (usually decreases, but with some exceptions), the direction of content effects is generally preserved under alternative scoring methods.

\paragraph{Human experiments}

The human experiments were conducted in 2023 using an online crowd-sourcing platform, and recruiting only participants from the UK who spoke English as a first language, and who had over a 95\% approval rate. We did not further restrict participation. We offered pay of {\textsterling}2.50 for our task. Our intent was to pay at a rate exceeding {\textsterling}15/h, and we exceeded this target, as most participants completed the task in less than 10 minutes.

The human participants were first presented with a consent form detailing the experiment and their ability to withdraw at any time. If they consented to participate, they then proceeded to an instructions page. After the instructions they were presented with one question from each of our three tasks, one at a time. Each participant saw the tasks in a randomized order, and with randomized conditions. Subsequently, the participants were presented with three rating questions, rating the believability of a rule from one of the Wason tasks (not the one they had completed), and their degree of agreement with a concluding proposition from a syllogism task, and a concluding proposition from the NLI task. In each case, the ratings were provided on a continuous scale from 0 to 100 (with 50\% indicated as neither agree nor disagree). On each question and rating, the participants had to answer in less than a time limit of 5 minutes (to ensure they were not abandoning the task entirely). This time limit was reset on the next question. See Appx. \ref{app:methods:human_experiments} for further detail on the experimental methods.

We first collected a dataset of responses from 625 participants. After observing the low accuracy in the Wason tasks, we collected an additional dataset from 360 participants in which we offered an additional performance bonus of {\textsterling}0.50 for answering the Wason question correctly, to motivate subjects. In this replication, we collected data only on the Realistic and Abstract Wason conditions. In our main analyses, we collapse across these two subsets, but we present the results for each experiment separately in Appx.\ \ref{app:fig:wason_human_replication}.

Due to infrastructure restrictions in the framework used to create the human tasks, we assigned participants to conditions and items randomly rather than with precise balancing. Furthermore, a few participants timed out on some questions, and there were a handful of instances of data not saving properly due to server issues. Thus, the exact number of participants for which we have data varies slightly from task to task and item to item.

\paragraph{Statistical analyses}

Our main analyses are quantified with mixed-effects logistic regression models that include task condition variables as predictors, and control for random effects of items, and, where applicable, models. The key results of these models are reported in the main text. The full model specifications and full results are provided in Appx. \ref{app:statistical_analyses}.

\bibliography{main.bib}
\bibliographystyle{ScienceAdvances}

\noindent \textbf{Acknowledgements:} 
We thank Michiel Bakker, Michael Henry Tessler, Matt Botvinick, and Adam Santoro for helpful comments and suggestions.\\
\noindent \textbf{Funding:} This work was not supported by external funding.\\
\noindent \textbf{Author Contributions} 
{AKL} formulated and lead the project, developed the syllogisms and Wason datasets, performed the model experiments, performed the human experiments, performed the main analyses and created the figures, and wrote the paper.
{ID} initiated the investigation into interactions between language model knowledge and reasoning, formulated and lead the project, performed the model experiments, and wrote the paper.
{SCYC} created the NLI dataset, contributed ideas, contributed to experiments, and contributed to writing the paper. 
{HS} created the human experiments infrastructure, assisted with the human experiments and contributed to the paper.
{AC} contributed ideas, created technical infrastructure, and contributed to the paper.
{DK, JLM} and {FH} advised throughout and contributed to writing the paper.\\
\noindent \textbf{Competing Interests} The authors declare that they have no competing financial interests.\\
\noindent \textbf{Data and materials availability:} Additional data and materials are available upon request.

\clearpage
\appendix

\noindent
In Appx.\ \ref{app:methods} we provide more details of the methods and datasets, in Appx.\ \ref{app:supp_analyses} we provide supplemental analyses, and in Appx.\ \ref{app:statistical_analyses} we provide full results of statistical models for the main results.

\section{Supplemental methods} \label{app:methods}

\subsection{Datasets} \label{app:methods:datasets}

\begin{figure}[H]
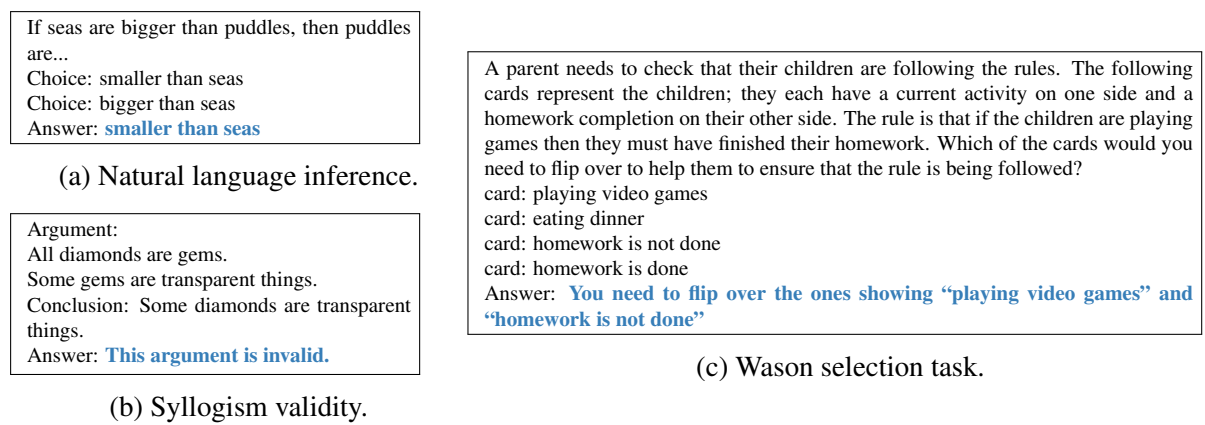

    \centering
    \begin{minipage}{0.38\textwidth}
    \begin{subfigure}[t]{\textwidth}
    \fbox{ \parbox{0.84\textwidth}{
    \scriptsize
    If seas are bigger than puddles, then puddles are...\\
    Choice: smaller than seas\\
    Choice: bigger than seas\\
    Answer: {\color{bblue}\bf smaller than seas}
    }}
    \caption{Natural language inference.} \label{fig:tasks:nli}
    \end{subfigure}\\[0.46em]
    \begin{subfigure}[t]{\textwidth}
    \fbox{ \parbox{0.84\textwidth}{
    \scriptsize
    Argument:\\
    All diamonds are gems.\\
    Some gems are transparent things.\\
    Conclusion: Some diamonds are transparent things.\\
    Answer: {\color{bblue}\bf This argument is invalid.}
    }}
    \caption{Syllogism validity.} \label{fig:tasks:syllogisms}
    \end{subfigure}
    \end{minipage}%
    \begin{minipage}{0.62\textwidth}
    \begin{subfigure}[t]{\textwidth}
    \fbox{ \parbox{0.95\textwidth}{
    \scriptsize
A parent needs to check that their children are following the rules. The following cards represent the children; they each have a current activity on one side and a homework completion on their other side. The rule is that if the children are playing games then they must have finished their homework. Which of the cards would you need to flip over to help them to ensure that the rule is being followed?\\
card: playing video games\\
card: eating dinner\\
card: homework is not done\\
card: homework is done\\
Answer: {\color{bblue}\bf You need to flip over the ones showing ``playing video games'' and ``homework is not done''}
    }}
    \caption{Wason selection task.} \label{fig:tasks:wason}
    \end{subfigure}%
    \end{minipage}
    \caption{Examples of the three logical reasoning tasks we evaluate, as they were presented to the models: (\subref{fig:tasks:nli}) simple single-step natural language inferences, (\subref{fig:tasks:syllogisms}) assessing the validity of logical syllogisms, and (\subref{fig:tasks:wason}) the Wason selection task. In each case, the model must choose the answer (blue and bold) from a set of possible answer choices.}
    \label{fig:tasks}
\end{figure}

As noted in the main text, we generated new datasets for each task to avoid problems with training data contamination. In this section we present further details of dataset generation.

\subsubsection{NLI task generation} In the absence of existing cognitive literature on generating belief-aligned stimuli for this task, we used a larger language model \citep[Gopher, 280B parameters, from][]{rae2021scaling} to generate 100 comparison statements automatically, by prompting it with 6 comparisons that are true in the real world. The exact prompt used was:

\begin{Verbatim}[breaklines,fontsize=\footnotesize]
The following are 100 examples of comparisons:

1. mountains are bigger than hills 

2. adults are bigger than children 

3. grandparents are older than babies 

4. volcanoes are more dangerous than cities 

5. cats are softer than lizards
\end{Verbatim} 

We prompted the LLM multiple times, until we had generated 100 comparisons that fulfilled the desired criteria. The prompt completions were generated using nucleus sampling \citep{holtzman2019curious} with a probability mass of 0.8 and a temperature of 1. We filtered out comparisons that were not of the form ``[entity] is/are [comparison] than [other entity]''. We then filtered these comparisons manually to remove false and subjective ones, so the comparisons all respect real-world facts. An example of the generated comparisons includes ``puddles are smaller than seas''. 

We generated a natural inference task derived from these comparison sentences as follows. We began with the \textit{consistent} version, by taking the the raw output from the LM, ``puddles are smaller than seas'' as the hypothesis and formulating a premise ``seas are bigger than puddles'' such that the generated hypothesis is logically valid. We then combine the premise and hypothesis into a prompt and continuations. For example:
\begin{Verbatim}[breaklines,fontsize=\footnotesize]
If seas are bigger than puddles, then puddles are
A. smaller than seas
B. bigger than seas
\end{Verbatim}
where the logically correct (A) response matches real-world beliefs (that  `puddles are smaller than seas'). Similarly, we can also generate a \textit{violate} version of the task where the logical response violates these beliefs. For example, 
\begin{Verbatim}[breaklines,fontsize=\footnotesize]
If seas are smaller than puddles, then puddles are
A. smaller than seas
B. bigger than seas
\end{Verbatim}
here the correct answer, (B), violates the LM's prior beliefs.
Finally, to generate a \textit{nonsense} version of the task, we simply replace the nouns (`seas' and `puddles') with nonsense words. For example:
\begin{Verbatim}[breaklines,fontsize=\footnotesize]
If vuffs are smaller than feps, then feps are
A. smaller than vuffs
B. bigger than vuffs
\end{Verbatim}
Here the logical conclusion is B. For each of these task variations, we evaluate the log probability the language model places on the two options and choose higher likelihood one as its prediction.

\subsubsection{Syllogisms data generation} 
\begin{figure}
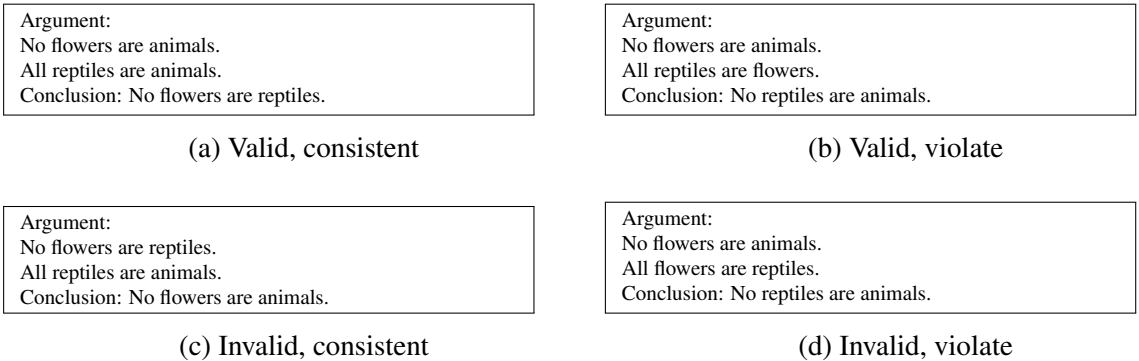

    \centering
    \begin{subfigure}{0.5\textwidth}
    \fbox{ \parbox{0.84\textwidth}{
    \scriptsize
    Argument:\\
    No flowers are animals.\\
    All reptiles are animals.\\
    Conclusion: No flowers are reptiles.
    }}
    \caption{Valid, consistent}
    \end{subfigure}%
    \begin{subfigure}{0.5\textwidth}
    \fbox{ \parbox{0.84\textwidth}{
    \scriptsize
    Argument:\\
    No flowers are animals.\\
    All reptiles are flowers.\\
    Conclusion: No reptiles are animals.
    }}
    \caption{Valid, violate}
    \end{subfigure}\\[1em]
    \begin{subfigure}{0.5\textwidth}
    \fbox{ \parbox{0.84\textwidth}{
    \scriptsize
    Argument:\\
    No flowers are reptiles.\\
    All reptiles are animals.\\
    Conclusion: No flowers are animals.
    }}
    \caption{Invalid, consistent}
    \end{subfigure}%
    \begin{subfigure}{0.5\textwidth}
    \fbox{ \parbox{0.84\textwidth}{
    \scriptsize
    Argument:\\
    No flowers are animals.\\
    All flowers are reptiles.\\
    Conclusion: No reptiles are animals.
    }}
    \caption{Invalid, violate}
    \end{subfigure}\\
\caption{Example syllogism cluster, showing \(2 \times 2\) design of valid (top row), invalid (bottom row), and consistent (left column) and violate (right column) arguments.} \label{app:fig:example_sylls}
\end{figure}

We generated a new set of problems for syllogistic reasoning. Following the approach of Evans et al. \citep{evans1983conflict}, in which the syllogisms were written based on the researchers' intuitions of believability, we hand-authored these problems based on beliefs that seemed plausible to the authors. See Fig. \ref{fig:tasks:syllogisms} for an example problem. We built the dataset from clusters of 4 arguments that use the same three entities, in a \(2 \times 2\) combination of valid/invalid, and belief-consistent/violate. For example, in Fig. \ref{app:fig:example_sylls} we present a full cluster of arguments about reptiles, animals, and flowers.

By creating the arguments in this way, we ensure that the low-level properties (such as the particular entities referred to in an argument) are approximately balanced across the relevant conditions. In total there are twelve clusters. We avoided using the particular negative form (``some X are not Y'') to avoid substantial negation, which complicates behavior both for language models and humans \citep[cf.][]{hosseini2021understanding,evans1996role}. We then sampled an identical set of nonsense arguments by simply replacing the entities in realistic arguments with nonsense words.  

We present the arguments to the model, and give a forced choice between ``The argument is valid.'' or ``The argument is invalid.'' Where example shots are used, they are sampled from distinct clusters, and are separated by a blank line. We also tried some minor variations in preliminary experiments (such as changing the prompt or prefixing the conclusion with ``Therefore:'' or omitting the prefix before the conclusion), but observed qualitatively similar results so we omit them here.

\begin{figure}
    \centering
    \begin{subfigure}{0.5\textwidth}
    \begin{Verbatim}[breaklines,fontsize=\tiny]
Some librarians are happy people
All happy people are healthy people
Conclusion: Some librarians are healthy people

All guns are weapons
All weapons are dangerous things
Conclusion: All guns are dangerous things

Some electronics are computers
All computers are expensive things
Conclusion: Some electronics are expensive things

All trees are plants
Some trees are tall things
Conclusion: Some plants are tall things

No flowers are animals
All reptiles are animals
Conclusion: No flowers are reptiles

All diamonds are gems
Some diamonds are transparent things
Conclusion: Some gems are transparent things
    \end{Verbatim}
    \end{subfigure}%
    \begin{subfigure}{0.5\textwidth}
    \begin{Verbatim}[breaklines,fontsize=\tiny]
All dragons are mythical creatures
No mythical creatures are things that exist
Conclusion: No dragons are things that exist

Some politicians are dishonest people
All dishonest people are people who lie
Conclusion: Some politicians are people who lie

All whales are mammals
Some whales are big things
Conclusion: Some mammals are big things

All vegetables are foods
Some vegetables are healthy things
Conclusion: Some foods are healthy things

All famous actors are wealthy people
Some famous actors are old people
Conclusion: Some old people are wealthy people

All vehicles are things that move
No buildings are things that move
Conclusion: No buildings are vehicles
    \end{Verbatim}
    \end{subfigure}%
    \caption{One argument (valid, consistent) from each of the 12 argument clusters we used for the syllogisms tasks, showing the entities and argument forms covered.}
    \label{app:fig:syll_problems}
\end{figure}

\subsubsection{Wason data generation}
As above, we generated a new dataset of Wason problems to avoid potential for dataset contamination (see Fig. \ref{fig:tasks:wason} for an example). The final response in a Wason task does not involve a declarative statement (unlike completing a comparison as in NLI), so answers do not directly `violate' beliefs. Rather, in the cognitive science literature, the key factor affecting human performance is whether the entities are `realistic' and follow `realistic' rules (such as people following social norms) or consist of arbitrary relationships between abstract entities such as letters and numbers. We therefore study the effect of realistic and arbitrary scenarios in the language models.

We created 12 realistic rules and 12 arbitrary rules. Each rule appears with four instances, respectively matching and violating the antecedent and consequent. Each realistic rule is augmented with one sentence of context for the rule, and the cards are explained to represent the entities in the context. The model is presented with the context, the rule, and is asked which of the following instances it needs to flip over, then the instances. The model is then given a forced choice between sentences of the form ``You need to flip over the ones showing ``X'' and ``Y''.'' for all subsets of two items from the instances. There are two choices offered for each pair, in both of the possible orders, to eliminate possible biases if the model prefers one ordering or another. (Recall that the model scores each answer independently; it does not see all answers at once.)

See Figs. \ref{app:fig:wason_rules_realistic} and \ref{app:fig:wason_rules_arbitrary} for the realistic and arbitrary rules and instances used --- but note that problems were presented to the model with more context and structure, see Fig. \ref{fig:tasks:wason} for an example. We demonstrate in Appx. \ref{app:analyses:wason_propositions} that the difficulty of basic inferences about the propositions involved in each rule type is similar across conditions.

We also created 12 rules using nonsense words. Incorporating nonsense words is less straightforward in the Wason case than in the other tasks, as the model needs to be able to reason about whether instances match the antecedent and consequent of the rule. We therefore use nonsense rules of the form ``If the cards have less gluff, then they have more caft'' with instances being more/less gluff/caft. The more/less framing makes the instances roughly the same length regardless of rule type, and avoids using negation which might confound results \citep{hosseini2021understanding}.

Finally, we created two types of control rules based on the realistic rules, which we present here. First, we created shuffled realistic rules by combining the antecedents and consequents of different realistic rules, while ensuring that there is no obvious rationale for the rule. For example, one shuffled-realistic rule is ``If they are doctors, then they must have a parachute.''

We then created violate-realistic rules by taking each realistic rule and reversing its consequent. For example, the realistic rule ``If the clients are skydiving, then they must have a parachute'' is transformed to the violate rule ``If the clients are skydiving, then they must have a wetsuit'', but ``parachute'' is still included among the cards. The violate condition is designed to make the rule especially implausible in context of the examples (viz. requiring the item that is \emph{not} a parachute to skydive), while the rule in the shuffled condition is somewhat more arbitrary/belief neutral.

To rule out a possible specific effect of cards (which were used in the original tasks) we also sampled versions of each problem with sheets of paper or coins, but results are similar so we collapse across these conditions in the main analyses.

\begin{figure}
    \centering
    \begin{Verbatim}[breaklines,fontsize=\tiny]
An airline worker in Chicago needs to check passenger documents. The rule is that if the passengers are traveling outside the US then they must have showed a passport.
Buenos Aires / San Francisco / passport / drivers license

A chef needs to check the ingredients for dinner. The rule is that if the ingredients are meat then they must not be expired.
beef / flour / expires tomorrow / expired yesterday

A lawyer for the Innocence Project needs to examine convictions. The rule is that if the people are in prison then they must be guilty.
imprisoned / free / committed murder / did not commit a crime

A medical inspector needs to check hospital worker qualifications. The rule is that if the workers work as a doctor then they must have received an MD.
surgeon / janitor / received an MD / received a GED

A museum curator is examining the collection. The rule is that if the artworks are in the museum then they must be genuine.
displayed in the museum / not in the museum / genuine / forgery

An adventure trip organizer needs to ensure their clients have the appropriate gear. The rule is that if the clients are going skydiving then they must have a parachute.
skydiving / mountain biking / parachute / wetsuit

A parent needs to check that their children are following the rules. The rule is that if the children are playing games then they must have finished their homework.
playing video games / eating dinner / homework is done / homework is not done

A priest needs to check if people are ready for marriage. The rule is that if the people are engaged then they must be adults.
engaged / single / 25 years old / 7 years old

A traffic enforcement officer needs to check that people are following the law. The rule is that if the people in vehicles are driving then they must have a driver license.
driver / passenger / has a license / does not have a license

A gardener needs to take care of their plants. The rule is that if the plants are flowers then they must be fertilized.
rose / oak / fertilized / not fertilized

A farmer is getting equipment ready for the day. The rule is that if the pieces of equipment have an engine then they must have fuel.
tractor / shovel / has gasoline / does not have gasoline

A person is cleaning out and organizing his closet. The rule is that if the clothes are going to the thrift store then they must be old.
thrift store / keep / worn out / brand new

An employer needs to check that their business is following health regulations. The rule is that if the employees are working then they must not be sick.
working / on vacation / healthy / has a cold
    \end{Verbatim}
    \caption{Realistic Wason rules and instances used.}
    \label{app:fig:wason_rules_realistic}
\end{figure}

\begin{figure}
    \centering
    \begin{Verbatim}[breaklines,fontsize=\tiny]
The rule is that if the cards have a plural word then they must have a positive emotion.
crises / dog / happiness / anxiety

The rule is that if the cards have a soft texture then they must have a polygon.
soft / rough / hexagon / circle

The rule is that if the cards have a French word then they must have a positive number.
chapeau / sombrero / 4 / -1

The rule is that if the cards have a prime number then they must have a secondary color.
11 / 12 / purple / red

The rule is that if the cards have a European country then they must have something hot.
Germany / Brazil / furnace / ice cube

The rule is that if the cards have the name of a famous book then they must have the name of an elementary particle.
Moby Dick / Citizen Kane / neutrino / atom

The rule is that if the cards have a type of plant then they must have the name of a philosopher.
cactus / horse / Socrates / Napoleon

The rule is that if the cards have the name of a web browser then they must have a type of pants.
Internet Explorer / Microsoft Word / jeans / sweatshirt

The rule is that if the cards have a beverage containing caffeine then they must have a material that conducts electricity.
coffee / orange juice / copper / wood

The rule is that if the cards have something electronic then they must have a hairy animal.
flashlight / crescent wrench / bear / swan

The rule is that if the cards have a verb then they must have a Fibonacci number.
walking / slowly / 13 / 4

The rule is that if the cards have a text file extension then they must have a time in the morning.
.txt / .exe / 11:00 AM / 8:00 PM
    \end{Verbatim}
    \caption{Arbitrary Wason rules and instances used.}
    \label{app:fig:wason_rules_arbitrary}
\end{figure}

\subsubsection{Differences from an earlier preprint of this paper:} \label{app:methods:datasets:differences_from_previous_version}

Readers of an earlier preprint of this paper (\url{https://arxiv.org/abs/2207.07051v1}) may notice some differences in task format and performance of Chinchilla, especially on the NLI task. These differences are due to our attempts to adapt the tasks in order to present them to human participants. In order to align comparisons between humans and the language models \citep[cf.][]{lampinen2022can}, we then ported the human-oriented changes back into the format used for language model evaluation.

For example, in the original paper we did not show the model the two possible choices for the NLI task; we simply evaluated the model's likelihood of each continuation. However, because we presented the tasks multiple choice to the humans in multiple choice format, we showed them the two possible answers. Thus, in the current version of the paper we also included the two answer choices in the prompt when evaluating the language models, followed by ``Answer:'', and only then evaluate the model (see Fig. \ref{fig:tasks:nli}).  Likewise, in the original version of the paper we did not provide instructions before the tasks; in this version we attempted to match the relevant portions of the human instructions.

These changes mean that the results in the current version of the paper cannot be directly compared to the results in the earlier version.

\subsection{Evaluation}

\textbf{DC-PMI correction:} We use the DC-PMI correction \citep{holtzman2021surface} for the syllogisms and Wason tasks; i.e., we choose an answer from the set of possible answers (\(\mathcal{A}\)) as follows:
\[
\text{argmax}_{a \in \mathcal{A}} \,\,\,p\!\left(a \,| \,\text{question}\right) - p\!\left(a \,| \, \text{baseline prompt}\right)
\]
Where the baseline prompt is the task instruction prompt, followed by ``Answer:'' and \(p(x \,|\, y)\) denotes the model's evaluated likelihood of continuation \(x\) after prompt \(y\).

\textbf{Instruction prompt:} We prefixed each question with a two-part instruction prompt that attempted to match the human generic and task-specific instructions (see below). We began each of these prompts with the performance relevant generic instructions that preceded our human experiment:

\begin{Verbatim}[breaklines,fontsize=\tiny]

In this task, you will have to answer a series of questions. You will have to choose the best answer to complete a sentence, paragraph, or question. Please answer them to the best of your ability.\n\n
\end{Verbatim}

After two linebreaks, a task-specific instruction was appended:

\noindent
\textit{NLI}:
\begin{Verbatim}[fontsize=\tiny]

Please choose the best completion for the following sentence:\n
\end{Verbatim}

\noindent
\textit{Syllogisms}:
\begin{Verbatim}[fontsize=\tiny,breaklines]

Please assume that the first two sentences in the argument are true. Determine whether the argument is valid, that is, whether the conclusion follows from the first two sentences:\n'
\end{Verbatim}

\noindent
\textit{Wason}:
\begin{Verbatim}[fontsize=\tiny]

Please answer the following question carefully:\n
\end{Verbatim}

Finally, the question was appended to this prompt.

\subsection{Human experiments} \label{app:methods:human_experiments}

The exact text seen by the participants before each question was as follows:

\begin{lstlisting}[basicstyle=\footnotesize]
NLI_DEFAULT_PREFACE = (
    "Please choose the best completion for the following sentence:")
SYLLOGISMS_DEFAULT_PREFACE = (
    "Please assume that the first two sentences in the argument are true. "
    "Determine whether the argument is valid, that is, whether the conclusion "
    "follows from the first two sentences:")
WASON_DEFAULT_PREFACE = (
    "Please answer the following question carefully:"
)
WASON_BONUS_PREFACE = (
    "Please answer the following question carefully; <font color='#bb0044'>we "
    "will pay you an additional performance bonus of 0.5 GBP if you answer "
    "this question correctly</font>:"
)
PRIOR_AGREEMENT_PREFACE = (
    "Please rate how much you agree with the following statement, on a scale "
    "from 0% (disagree completely) to 50% (neither agree nor disagree) to 100% "
    "(agree completely)."
)
WASON_BELIEVABLE_PREFACE = (
    "Please rate how believable the following rule is, on a scale from 0% "
    "(completely unbelievable) to 50% (neither believable nor unbelievable) to "
    "100% (completely believable)."
)
\end{lstlisting}

%%%%%%%%%%%%%%%%%%%%%%%%%%%%%
\section{Supplemental analyses} \label{app:supp_analyses}

\subsection{Believability of the propositions and rules} \label{app:analyses:human priors}

\begin{figure}[H]
    \centering
    \begin{subfigure}{0.33\textwidth}
    \includegraphics[width=\textwidth]{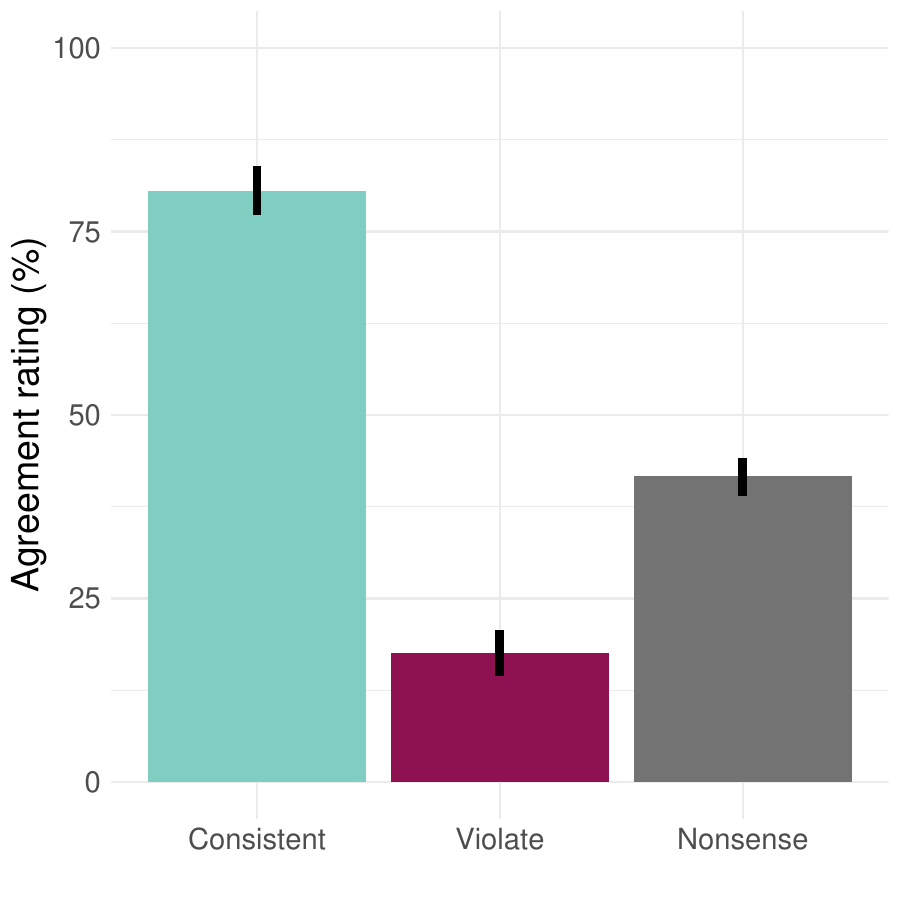}
    \caption{NLI} \label{app:fig:human_priors:nli} 
    \end{subfigure}%
    \begin{subfigure}{0.33\textwidth}
    \includegraphics[width=\textwidth]{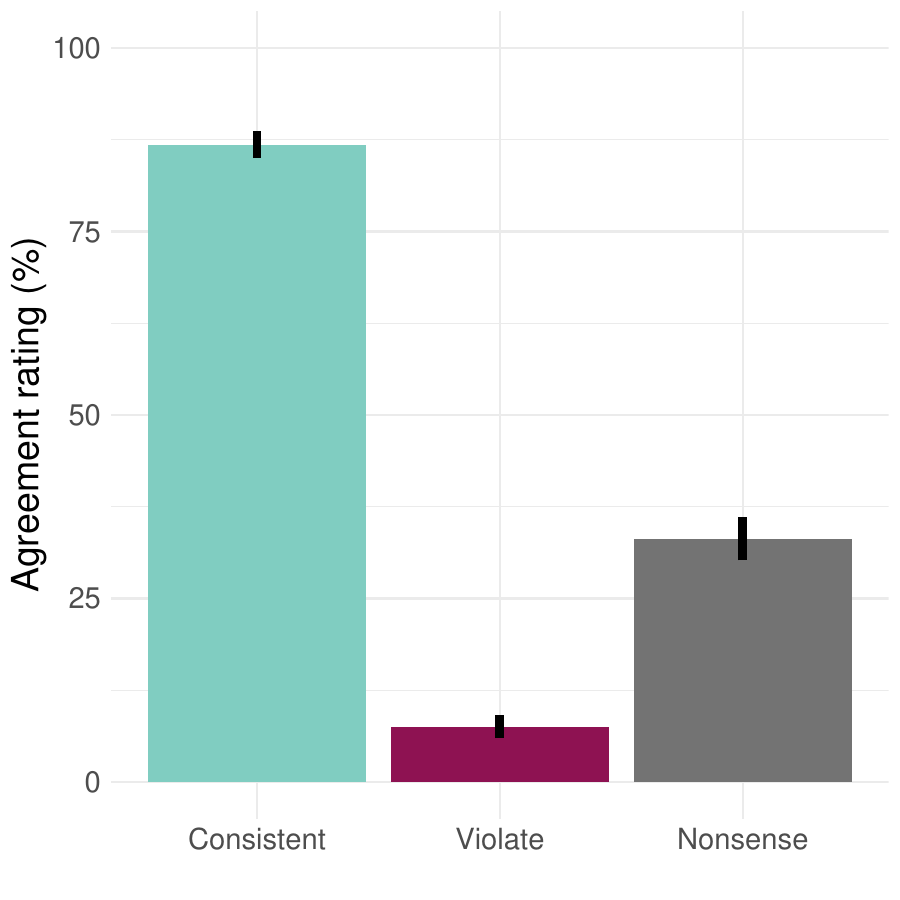}
    \caption{Syllogisms} \label{app:fig:human_priors:syllogisms} 
    \end{subfigure}%
    \begin{subfigure}{0.33\textwidth}
    \includegraphics[width=\textwidth]{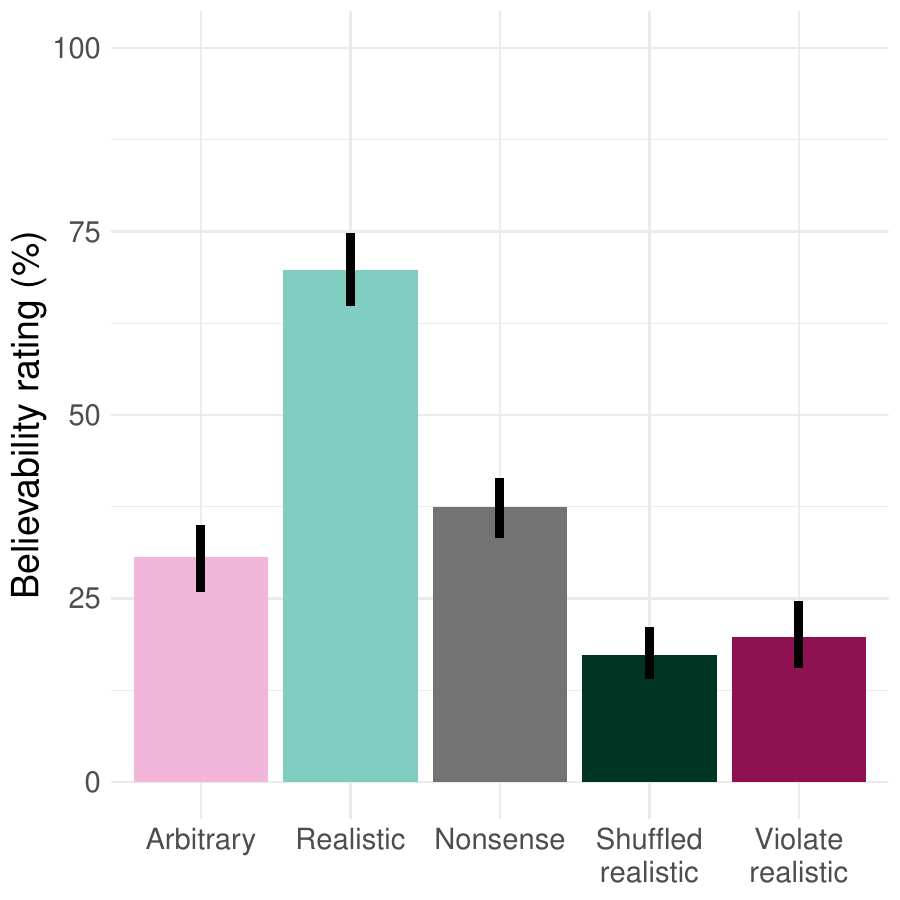}
    \caption{Wason} \label{app:fig:human_priors:wason} 
    \end{subfigure}%
    \caption{Our datasets align with human beliefs. When participants were asked how much they believed propositions or rules from our three tasks (\subref{app:fig:human_priors:nli}-\subref{app:fig:human_priors:wason}), they rated the Consistent or Realistic conditions as much more believable than the Violate ones, with Nonsense in between.}
    \label{app:fig:human_priors}  
\end{figure}

In order to assess the validity of our new datasets, we collected believability ratings from each subject, after they had completed the three tasks tasks, on one stimulus from each task type (not the version they had seen). Specifically, we asked the participants how believable a Wason rule was, and how much they agreed or disagreed with a proposition. In Fig. \ref{app:fig:human_priors} we show that participants found Consistent and Realistic stimuli much more believable than those in other conditions. 

\subsection{Robustnesss of the main language model results to raw-likelihood scoring and few-shot prompting} \label{app:analyses:robustness}

In this section, we show that the content effects we observe are robust to various manipulations of the evaluation context.

\subsubsection{Removing instruction prompts} \label{app:analyses:robustness:no_instructions}

\begin{figure}
    \centering
    \begin{subfigure}{0.42\textwidth}
    \centering
    \includegraphics[width=\textwidth]{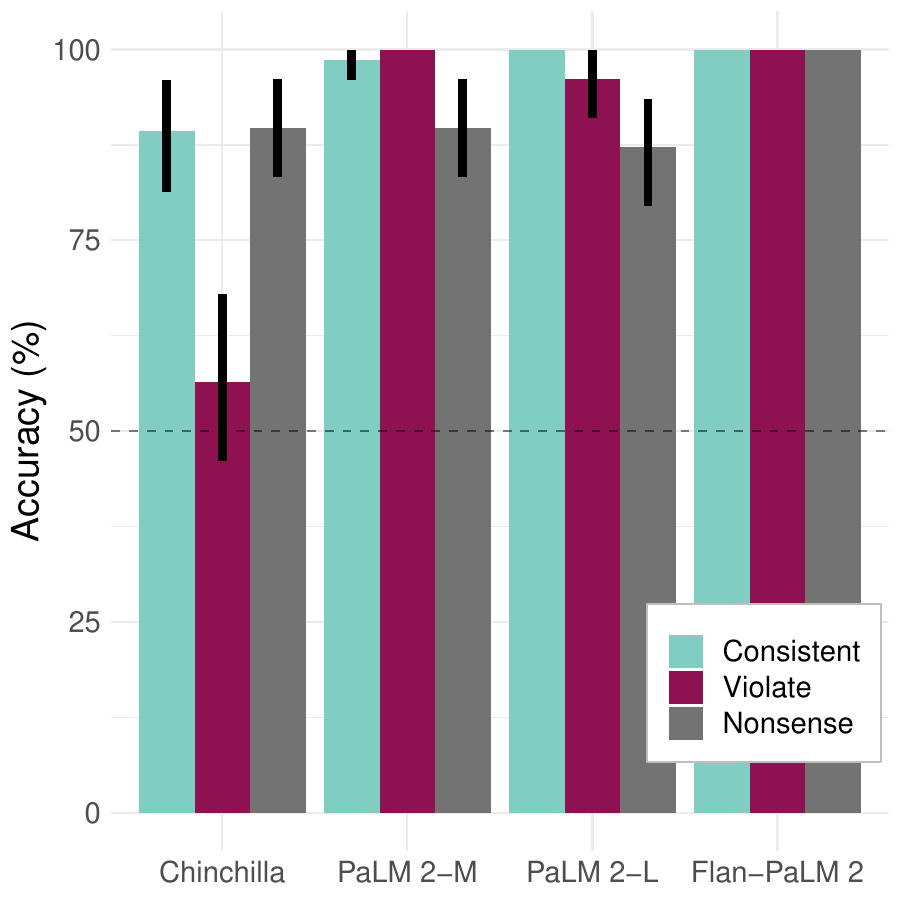}
    \caption{NLI.} \label{app:fig:no_instructions:nli}
    \end{subfigure}%
    \begin{subfigure}{0.42\textwidth}
    \centering
    \includegraphics[width=\textwidth]{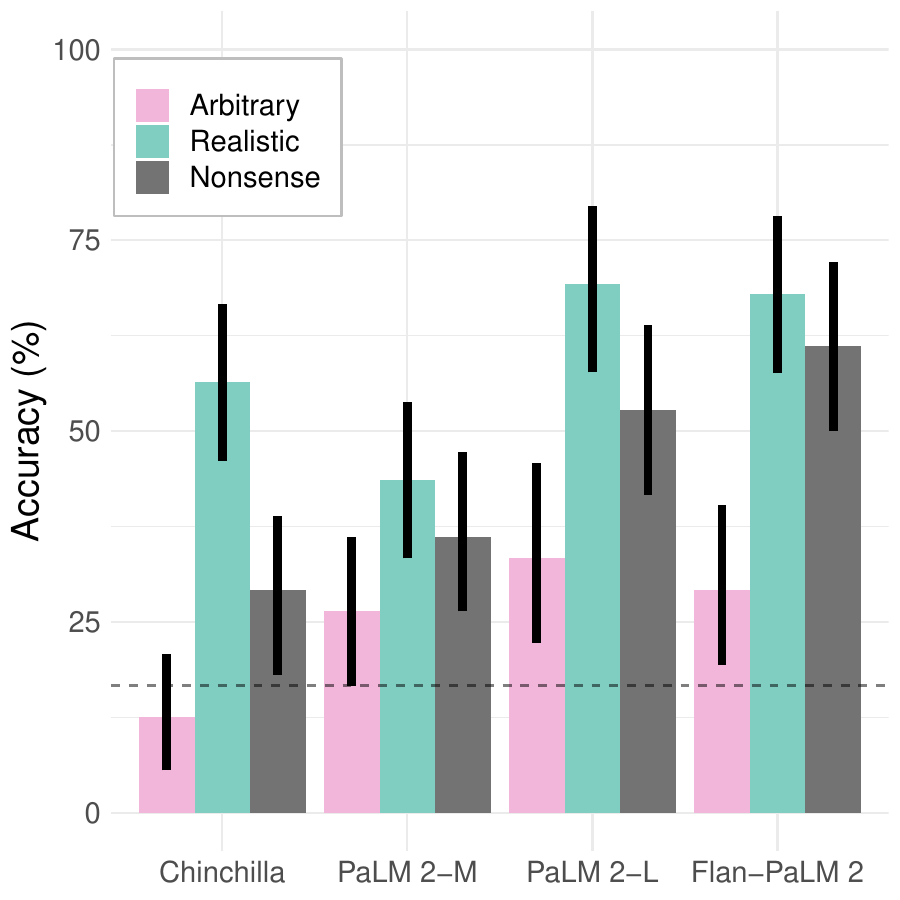}
    \caption{Wason.} \label{app:fig:no_instructions:wason}
    \end{subfigure}\\
    \begin{subfigure}{0.57\textwidth}
    \centering
    \includegraphics[width=\textwidth]{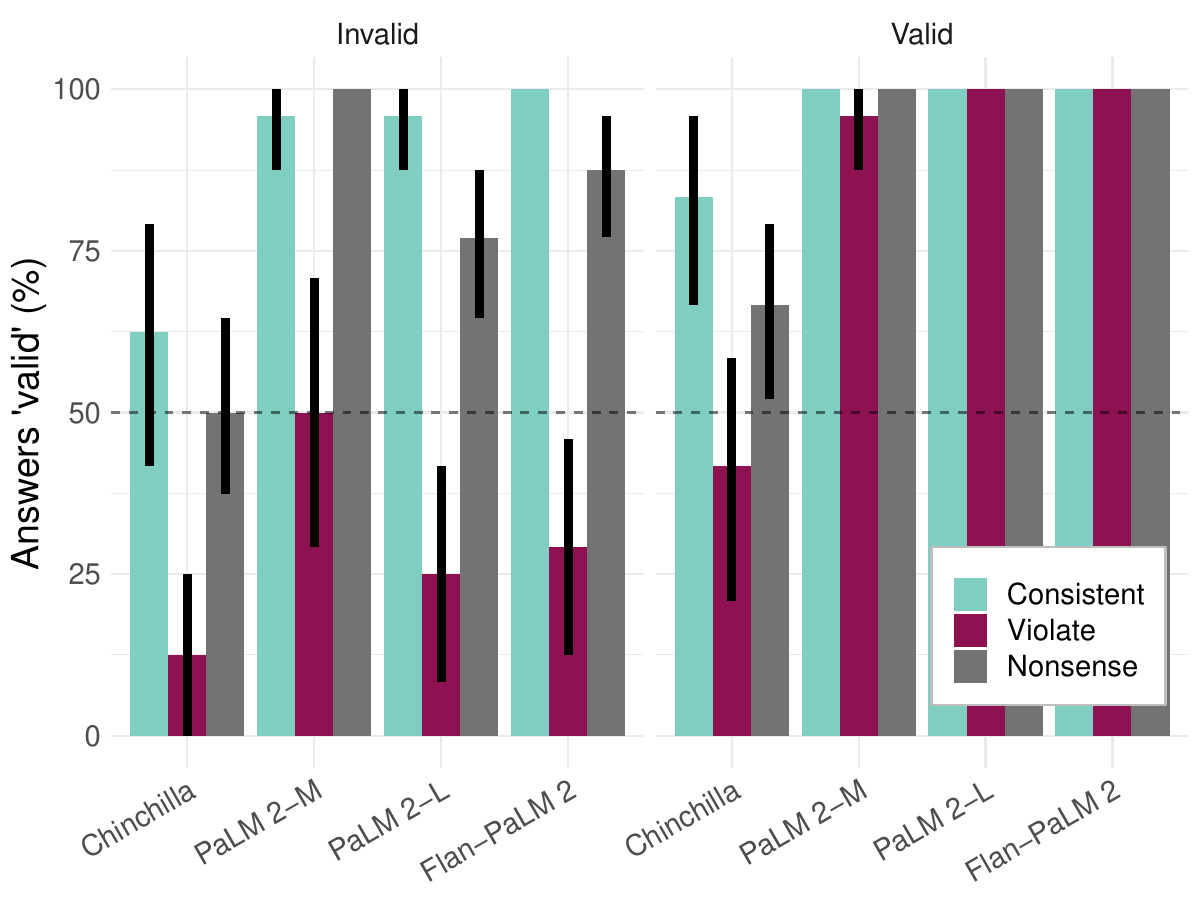}
    \caption{Syllogisms.} \label{app:fig:no_instructions:syl}
    \end{subfigure}
    \caption{Performance of a subset of the models when evaluated without an instruction prompt. Overall results and content effects are similar; however, in a few cases performance is noticeably impaired, particularly for Chinchilla on the Violate condition of NLI.}
    \label{app:fig:no_instructions}
\end{figure}

In the main text experiments, we provided models with an instruction prompt that roughly matched the human instructions \citep[cf.][]{lampinen2022can}. However, it is unclear how substantial a role this prompt played in performance, and human-likeness of the content effects. In Fig. \ref{app:fig:no_instructions} we show performance of a subset of the models when removing this instruction prompt; in most cases, results are similar, with a few notable exceptions. In particular Chinchilla shows much stronger content effects on the NLI tasks without instructions.

\subsubsection{Using raw likelihoods rather than Domain-Conditional PMI on the Syllogisms and Wason tasks} \label{app:analyses:dcmpi_vs_raw}

In the main results for the Syllogisms and Wason tasks, we scored the model using the Domain-Conditional PMI \citep{holtzman2021surface}. However, it is also common to score language models using raw likelihood comparisons. Would we observe the same content effects in that case?

\begin{figure}[H]
    \centering
    \begin{subfigure}{0.56\textwidth}
    \includegraphics[width=\textwidth]{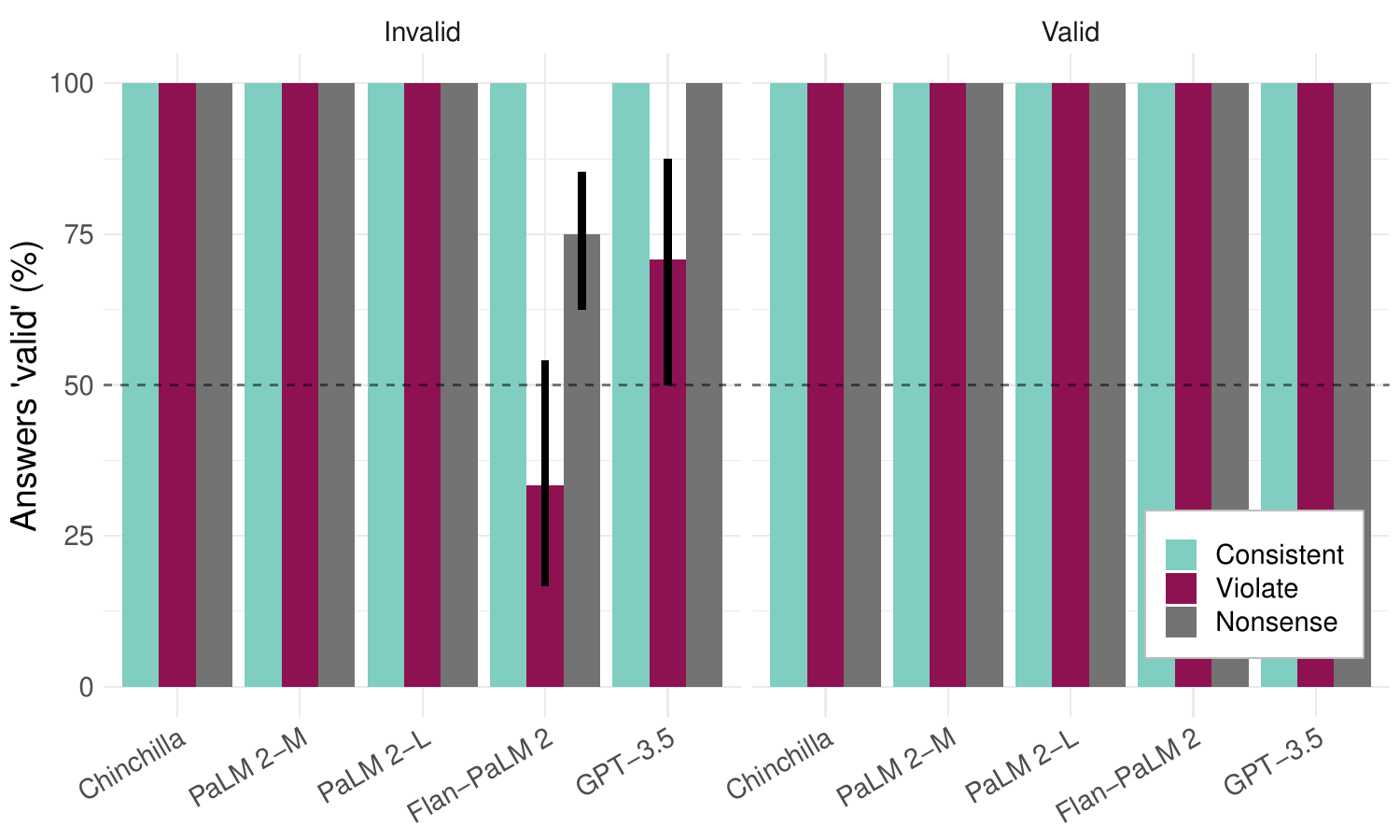}
    \caption{Syllogisms} \label{app:fig_syll_wason_nocorrection:syll} 
    \end{subfigure}%
    \begin{subfigure}{0.44\textwidth}
    \includegraphics[width=\textwidth]{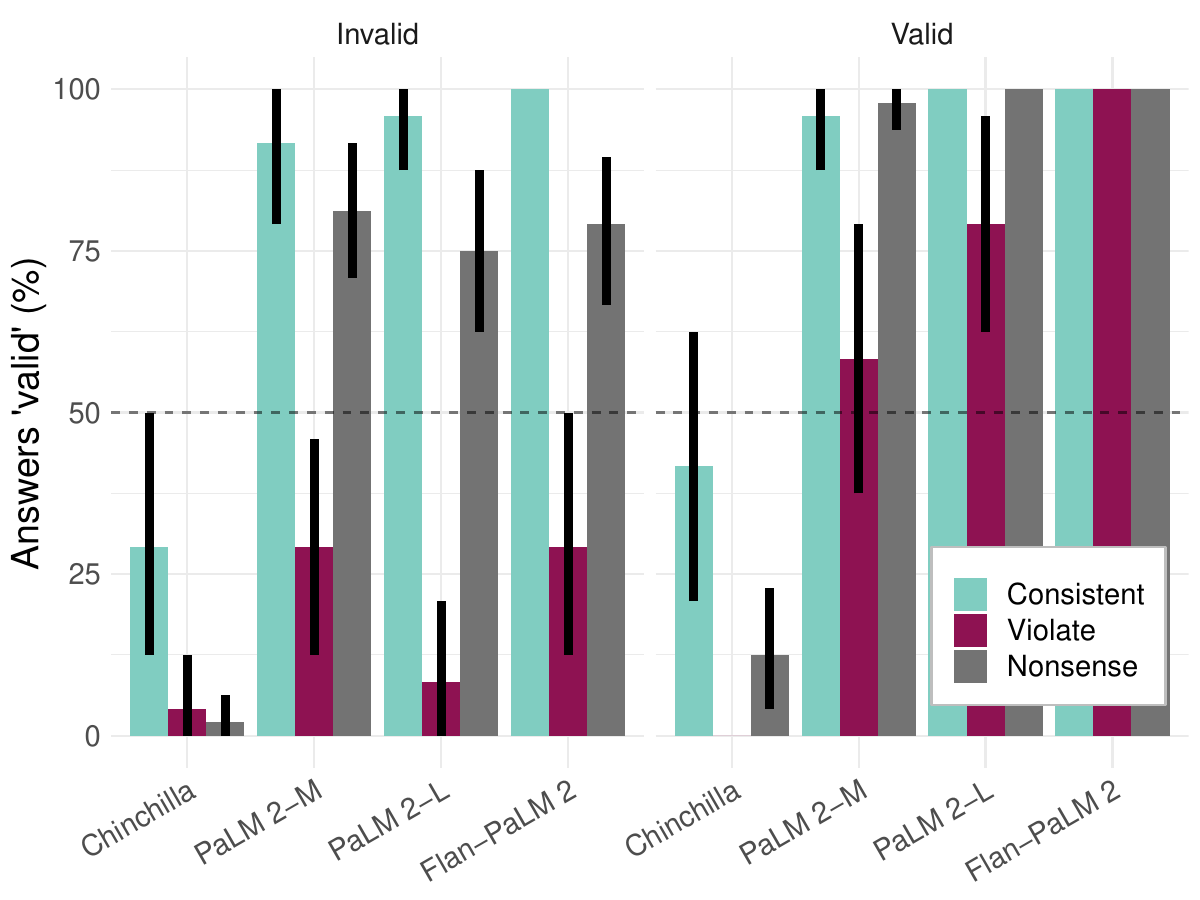}
    \caption{Syllogisms, no instruction prompt} \label{app:fig_syll_wason_nocorrection:syll_no_prompt} 
    \end{subfigure}\\
    \begin{subfigure}{0.56\textwidth}
    \includegraphics[width=\textwidth]{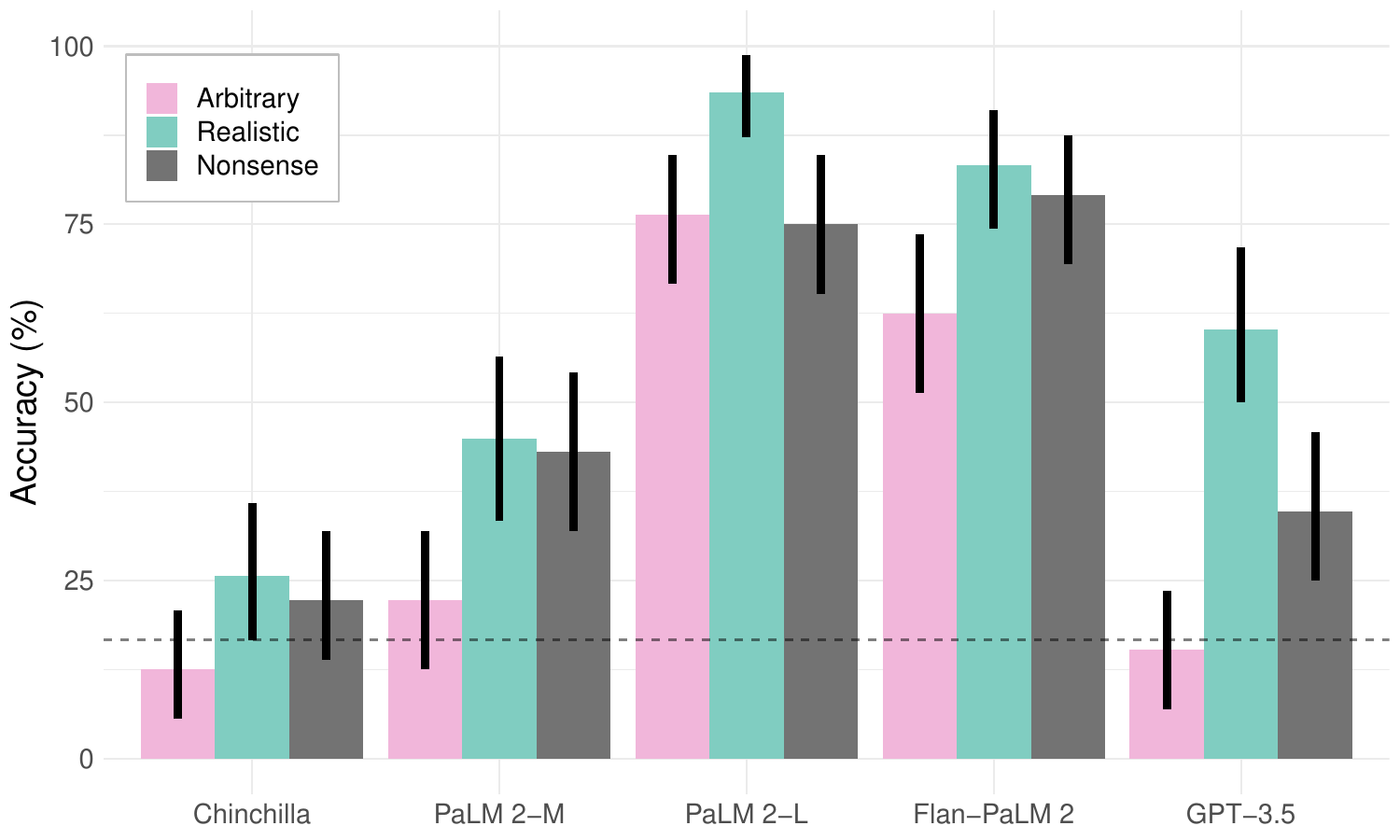}
    \caption{Wason} \label{app:fig_syll_wason_nocorrection:wason} 
    \end{subfigure}%
    \begin{subfigure}{0.44\textwidth}
    \centering
    \includegraphics[width=0.74\textwidth]{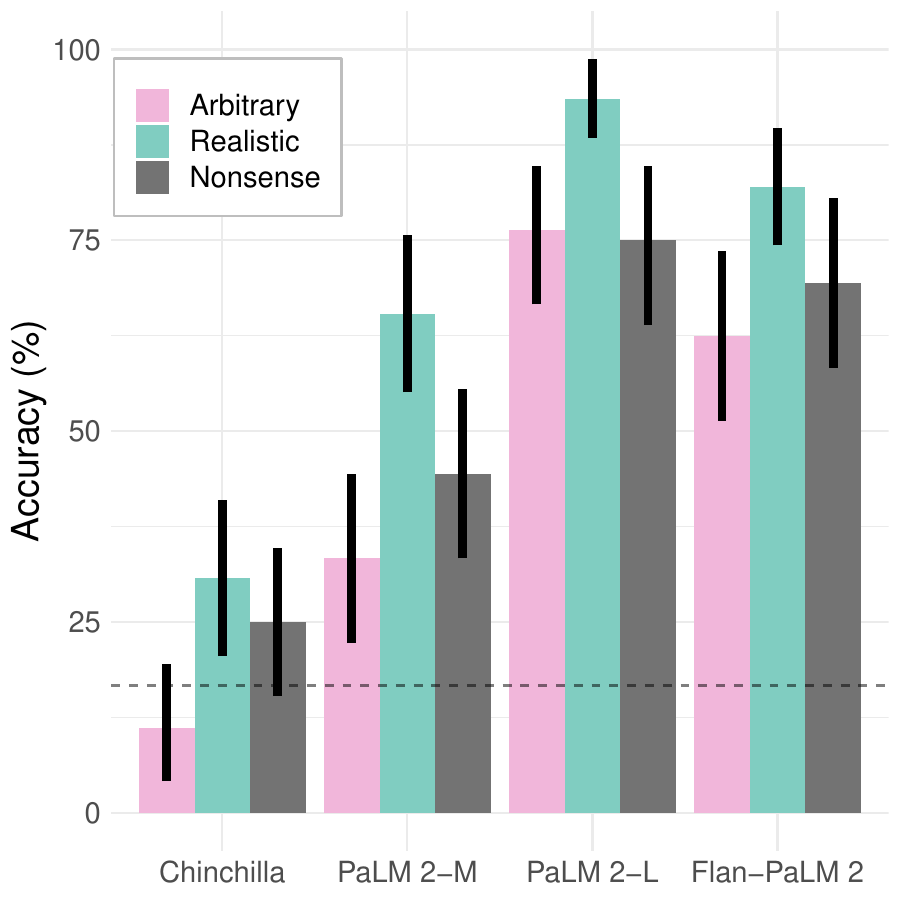}
    \caption{Wason, no instruction prompt} \label{app:fig_syll_wason_nocorrection:wason_no_prompt} 
    \end{subfigure}\\
    \caption{Scoring using the raw answer likelihoods --- rather than the Domain-Conditional PMI prior correction --- for the Syllogisms and Wason tasks. (\subref{app:fig_syll_wason_nocorrection:syll}) On the syllogisms tasks, removing the prior correction results in substantial answer biases for many models: much greater likelihood to say ``valid'' than ``invalid.'' Overall performance is much worse due to this bias; indeed, several models answer ``valid'' for every argument in every condition. However, for those that do not --- Flan-PaLM 2 and GPT-3.6 --- the direction of the content effects is as in the main text: the models are more likely to answer ``valid'' if the conclusion is belief-consistent.
    (\subref{app:fig_syll_wason_nocorrection:syll_no_prompt}) However, the answer bias on the syllogisms with raw-likelihood scoring seems to be strongly driven by the instruction prompt; without the prompt, the raw likelihoods yield less biased responses, and strong overall content effects.
    (\subref{app:fig_syll_wason_nocorrection:wason}-\subref{app:fig_syll_wason_nocorrection:wason_no_prompt}) On the Wason tasks, with or without the instruction prompt, removing the prior correction improves performance from some models, but hurts performance from others. Regardless, all models show the same pattern of content effects: facilitation in the Realistic rules compared to Arbitrary. (Compare to Figs. \ref{fig:zero_shot:syllogisms} and \ref{fig:zero_shot:wason}, respectively, which use DC-PMI scoring.))}
    \label{app:fig:syll_wason_nocorrection}  
\end{figure}

In Fig. \ref{app:fig:syll_wason_nocorrection} we show the results of raw-likelihood scoring. On the syllogisms tasks, this scoring method results in substantially more answer bias --- several of the models say valid in response to every problem, regardless of the content or logical structure. Thus, performance is much worse overall. However, for the models that do show any variability with content, the content effects are broadly similar to those observed in the main text: the models are more likely to say an argument is valid if the conclusion is belief-consistent than if the conclusion violates beliefs. Furthermore, if the instruction prompt is removed, the bias is substantially reduced, and stronger content effects are revealed.

In the Wason tasks, the effects on accuracy are more complex. While some models perform worse without the prior correction (e.g. Chinchilla), others perform much better. In particular, PaLM-2 L achieves over 75\% performance in every condition (including Arbitrary and Nonsense). However, all models that perform above chance show the same content effects observed in the main text: better performance on Realistic than Arbitrary rules. (In Appx. \ref{app:analyses:wason_choices_scoring_order} we also explore the effect of scoring with raw likelihoods on the individual card choices on the Wason task.)

\subsubsection{Effects of scoring method and answer order on the Wason answer choices} \label{app:analyses:wason_choices_scoring_order}

\begin{figure}[H]
    \centering
    \includegraphics[width=0.7\textwidth]{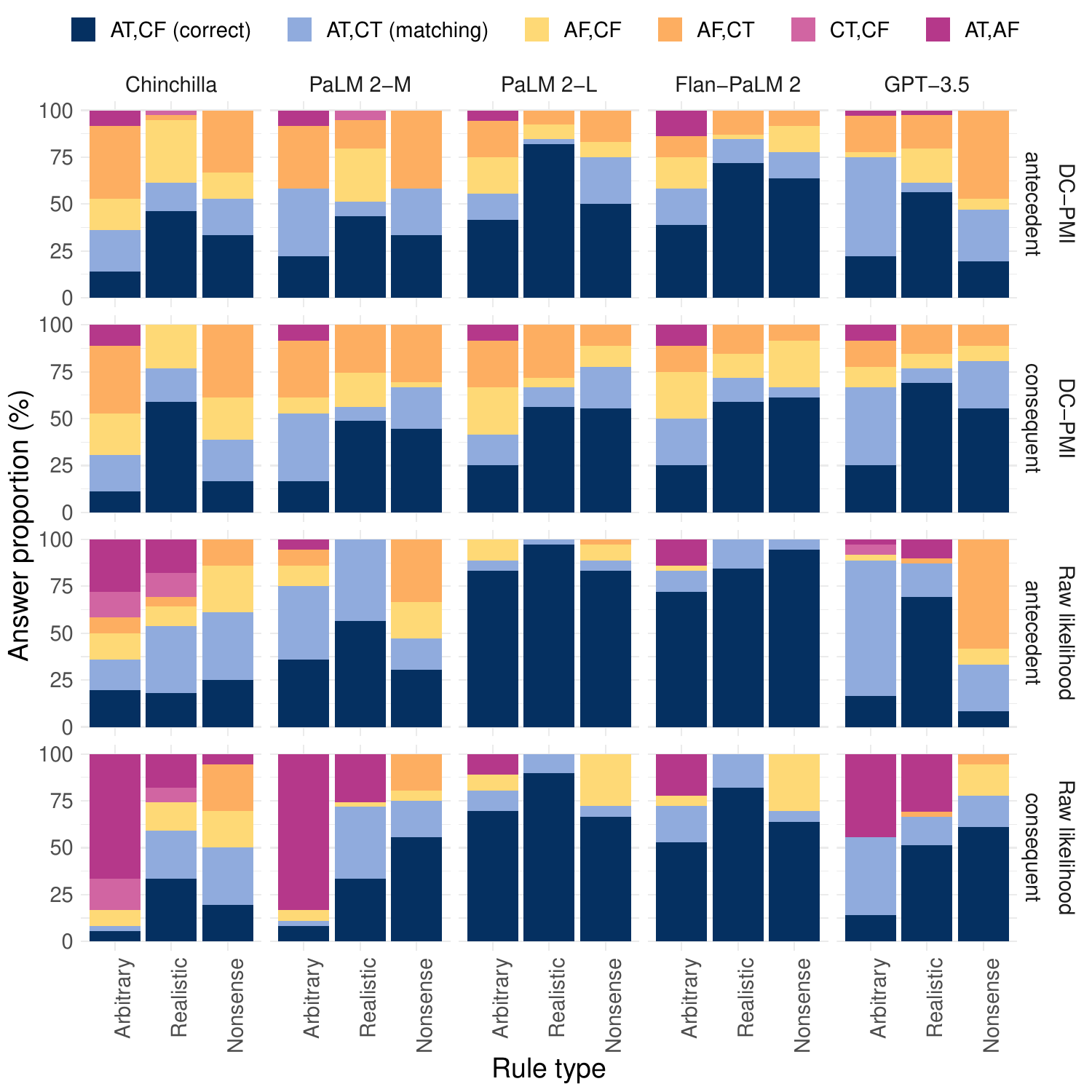}
    \caption{Effect of scoring method (DC-PMI in the top two rows vs. raw likelihoods in the bottom two) and ordering of the cards (antecedent cards first or consequent cards first; respectively in rows 1 and 3, and 2 and 4) on model choices. The DC-PMI prior correction does shift error patterns somewhat, and the models commit relatively more of the AT,AF answers with raw likelihood scoring, while with the DC-PMI scoring, the humans commit more of these errors than the models. The ordering of the cards does not have too substantial an effect, particularly with DC-PMI scoring. Generally, content effects --- that is, the advantage of the Realistic rules over arbitrary ones --- persists regardless of scoring method or order.}
    \label{app:fig:wason_answer_choices_scoring_order}
\end{figure}

In Fig. \ref{app:fig:wason_answer_choices_scoring_order} we show the effect of scoring method (DC-PMI vs. raw likelihoods) and the order in which the cards were presented (antecedent cards first or consequent cards first) on the models' answer choices on the Wason task. Scoring method does affect the error distribution fairly substantially, even where accuracy is similar; answer order has smaller effects.

\FloatBarrier
\subsubsection{Few-shot prompting of Chinchilla} \label{app:analyses:robustness:few_shot}

In all the main text experiments, we evaluated the models zero-shot, with only instructions. However, language model performance is generally improved by few-shot prompting \citep[e.g.][]{brown2020language}. We therefore evaluated whether few shot prompting with different kinds of prompt examples would alter the content effects we observed. (Note that, for computational reasons, we restrict these analyses to the Chinchilla model.) 
When we present a few-shot prompt of examples of the task to the model, the examples are presented with correct answers, and each example (as well as the final probe) is separated from the previous example by a single blank line. 

\begin{figure}
    \centering
    \includegraphics[width=0.66\textwidth]{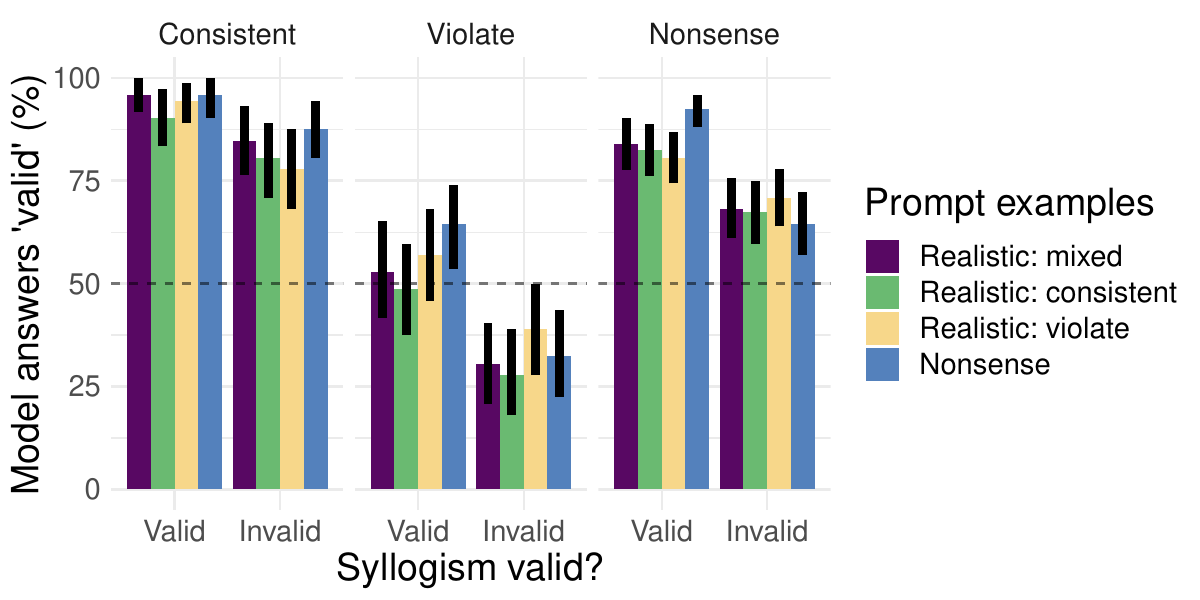}
    \caption{Chinchilla evaluated 5-shot on the syllogisms task, with different types of prompt examples. Content effects are very slightly reduced relative to the original experiments, but remain robust. The particular type of problems used in the prompt examples do not strongly affect performance. (The ``Realistic: mixed'' condition includes realistic examples from both the consistent and violate subsets.)}
    \label{app:fig:varied_few:syllogisms}
\end{figure}

In Fig. \ref{app:fig:varied_few:syllogisms} we show 5-shot prompting results for Chinchilla on the Syllogisms tasks. Content effects are slightly weaker than without the examples, but remain robust. 

\begin{figure}
    \centering
    \includegraphics[width=0.66\textwidth]{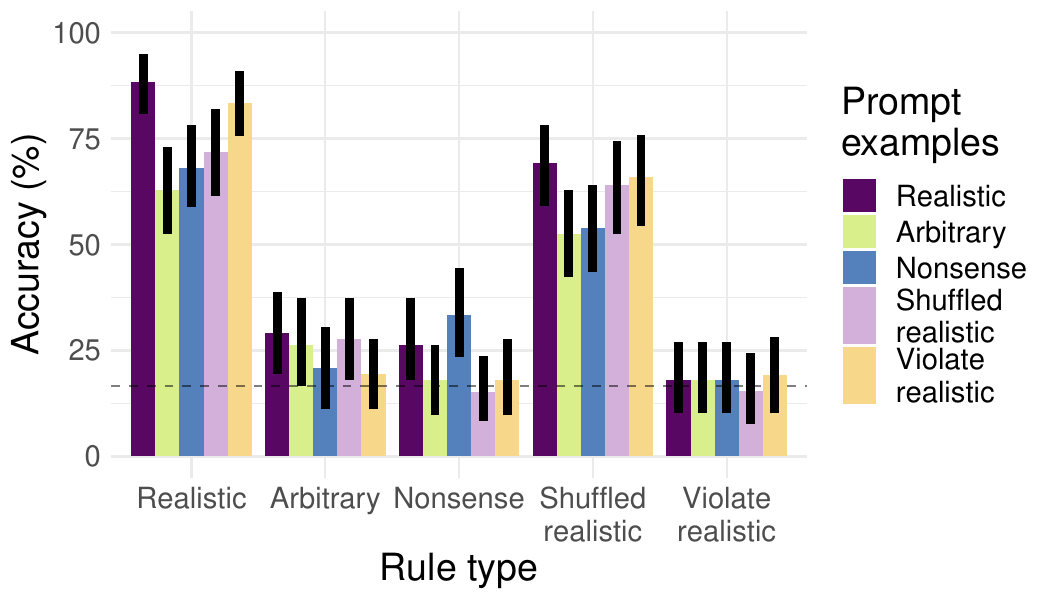}
    \caption{Chinchilla evaluated 5-shot on the Wason task, with different types of prompt examples. Again, content effects remain strong --- or are even amplified --- with few-shot prompts. Realistic prompt examples appear to be most beneficial overall, but especially for realistic and shuffled realistic probes, thus they actually enhance content effects. Other types of prompts are generally helpful in a more limited set of conditions; there may be an overall benefit to prompts matching probes.}
    \label{fig:varied_few:wason}
\end{figure}

In Fig. \ref{fig:varied_few:wason} we show 5-shot prompting results for Chinchilla on the Wason selection tasks. Content effects are exaggerated with the 5-shot prompts, because the model improves noticeably at Realistic rules, but improves less (if at all) on Arbitrary ones. We also see a noticeable effect of the type of examples used in the prompt, with Realistic examples offering optimal benefits.

\subsection{The Wason rule propositions have similar difficulty across conditions} \label{app:analyses:wason_propositions}

One possible confounding explanation for our Wason results would be that the base propositions that form the antecedents and consequents of the rules have different difficulty across conditions---this could potentially explain why the realistic rules and shuffled realistic rules are both easier than abstract or nonsesnse ones. To investigate this possibility, we tested the difficulty of identifying which of the options on the cards matched the corresponding proposition. Specifically, for the antecedent of the rule ``if the workers work as a doctor then they must have received an MD'' we prompted Chinchilla with a question like:
\begin{Verbatim}
Which choice better matches "work as a doctor"?
choice: surgeon
choice: janitor
Answer:
\end{Verbatim}
And then gave a two-alternative forced choice between `surgeon' and `janitor'. To avoid order biases, we repeated this process for both possible answer choice orderings in the prompt, and then aggregated likelihoods across these and chose the highest-likelihood answer.

\begin{figure}
    \centering
    \includegraphics[width=0.5\textwidth]{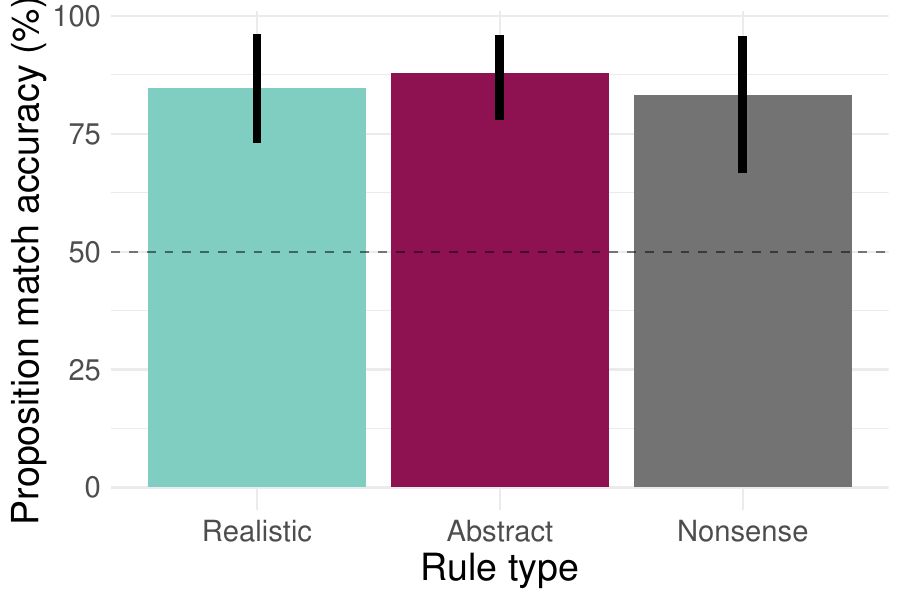}
    \caption{The component propositions (antecedents and consequents) of the Wason rules have similar difficulty across conditions. This plot shows Chinchilla's accuracy on forced choices of which instance matches a proposition, across conditions. (Note that the shuffled realistic rules use the same component propositions as the realistic rules.)}
    \label{app:fig:wason_premise_match}
\end{figure}

By this metric, we find that there are no substantial differences in difficulty across the rule types (Fig. \ref{app:fig:wason_premise_match})---in fact, arbitrary rule premises are numerically slightly easier, though the differences are not significant. Thus, the effects we observed are not likely to be explained by the base difficulty of verifying the component propositions.

\subsection{Additional recombined realistic conditions for the Wason tasks} \label{app:ablation_control:wason_other_conditions}

The Wason task rules can be realistic or unrealistic in multiple ways. For example, the component propositions can be realistic even if the relationship between them is not. We therefore generate two variations on realistic rules:\\
\textbf{\textit{Shuffled realistic}} rules, which combine realistic components in nonsensical ways (e.g. ``if the passengers are traveling outside the US, then they must have received an MD'').\\
\textbf{\textit{Violate realistic}} rules, which directly violate the expected relationship (e.g. ``if the passengers are flying outside the US, then they must have shown a drivers license [not a passport]'').\\

We also evaluated models and humans on these rules. For shuffled rules, results are well above chance. Surprisingly, one family of models (PaLM 2) even perform better at shuffled realistic than realistic rules. For violate rules, by contrast, performance is generally close to chance. It appears that the model reasons more accurately about rules formed from realistic propositions, particularly if the relationships between propositions in the rule are also realistic, but even to some degree if they are shuffled in nonsensical ways that do not directly violate expectations. However, if the rules strongly violate beliefs, performance is low. Humans generally perform poorly on either rule variant.

\begin{figure}
    \centering
    \includegraphics[width=0.66\textwidth]{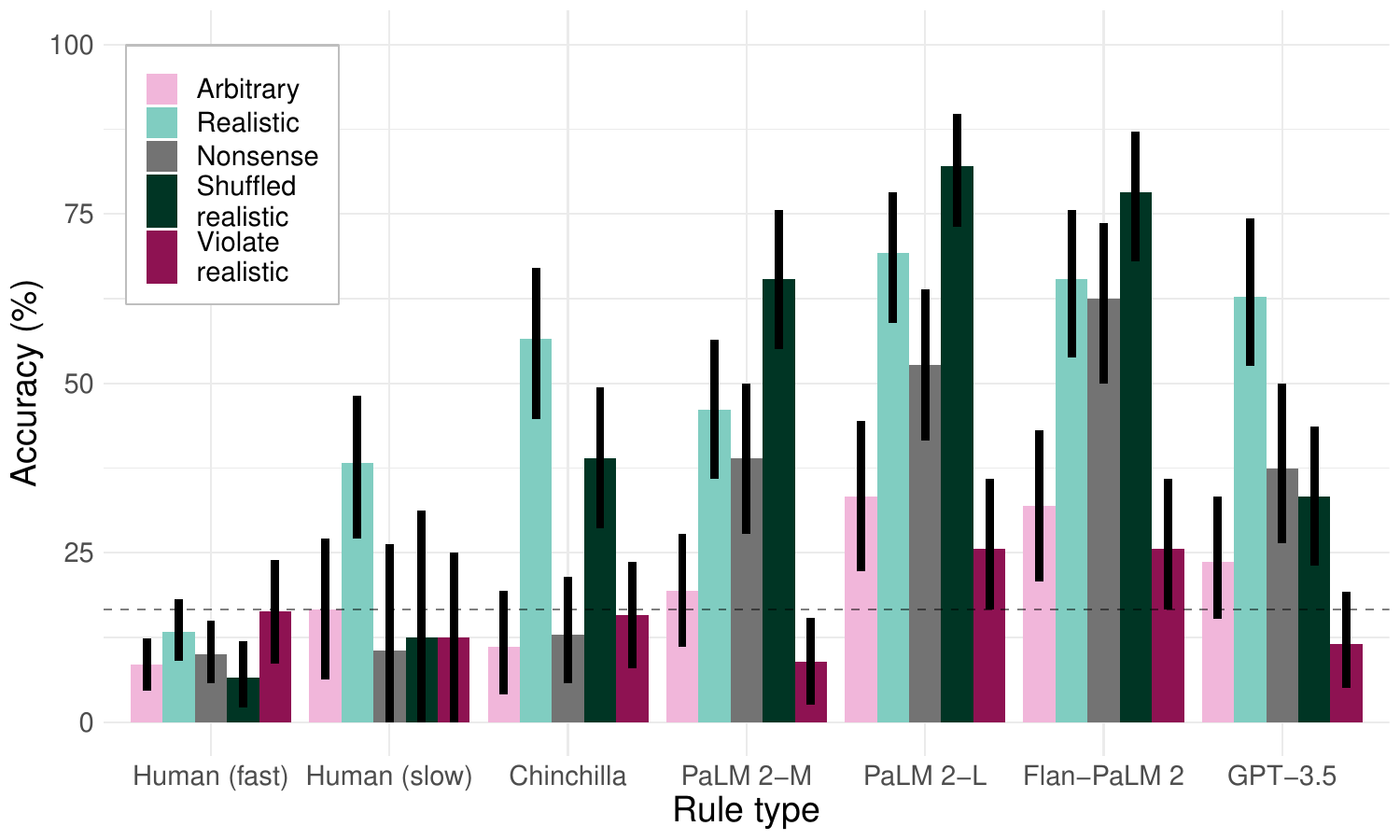}
    \caption{Evaluating models and humans on shuffled realistic and violate versions of the Wason rules. Humans}
    \label{app:fig:wason_shuffled_violate}
\end{figure}

\subsection{Human performance on the Wason tasks, in our original sample \& replication} \label{app:analyses:wason_human_replication}

\begin{figure}[H]
    \centering
    \includegraphics[width=0.66\textwidth]{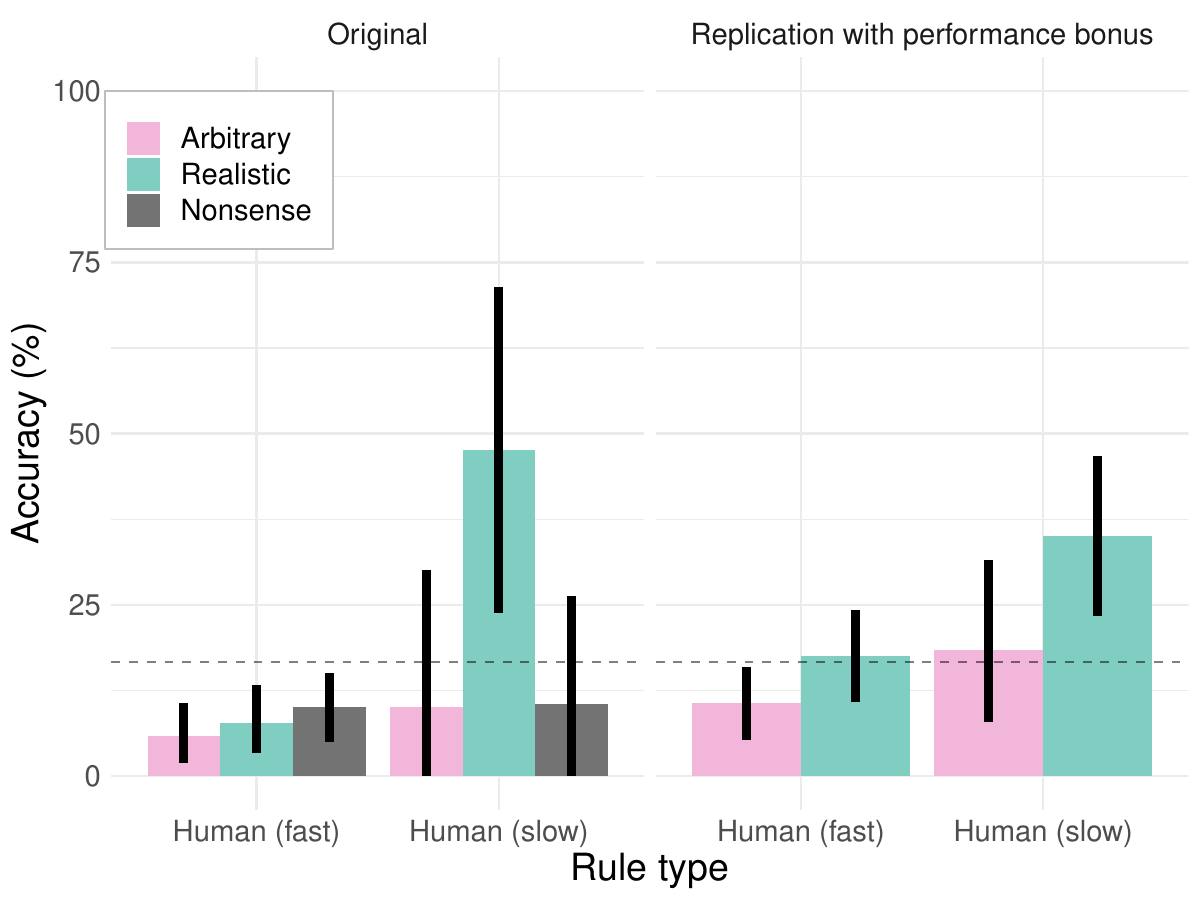}
    \caption{Breakdown of human results in our original experiment, and our replication (where we also added a performance bonus of 0.5 GPB for the Wason question). We observe a significant advantage for the slower humans in the Realistic condition in each case. The performance bonus does not seem to clearly improve performance. }
    \label{app:fig:wason_human_replication}
\end{figure}

As mentioned in the main text, after collecting our original sample on the Wason task, we recruited an additional set of participants to whom we offered a performance bonus on this task, in an attempt to increase performance. We present the results broken down by sample in Fig. \ref{app:fig:wason_human_replication}. We performed mixed-effects logistic regressions (Table \ref{app:tab:wason_human_replication_regression}) to test for an improvement in performance in the sample with a performance bonus; this effect was marginally significant. However, performance remains low overall, and we do not observe a significant difference in the content effect.

\begin{table*}[htbp]
\begin{subtable}{\textwidth}
\begin{Verbatim}[fontsize=\tiny]
Generalized linear mixed model fit by maximum likelihood (Laplace
  Approximation) [glmerMod]
 Family: binomial  ( logit )
Formula: response_correct ~ wason_condition + replication_experiment +  
    (1 | wason_name)
   Data: wason_human_correct_df

     AIC      BIC   logLik deviance df.resid 
   476.1    493.5   -234.0    468.1      570 

Scaled residuals: 
    Min      1Q  Median      3Q     Max 
-0.7247 -0.4879 -0.3449 -0.2718  3.9974 

Random effects:
 Groups     Name        Variance Std.Dev.
 wason_name (Intercept) 0.1704   0.4127  
Number of obs: 574, groups:  wason_name, 25

Fixed effects:
                           Estimate Std. Error z value Pr(>|z|)    
(Intercept)                 -2.6190     0.3029  -8.646  < 2e-16 ***
wason_conditionRealistic     0.8261     0.3066   2.694  0.00706 ** 
replication_experimentTRUE   0.5274     0.2695   1.957  0.05034 .  
---
Signif. codes:  0 ‘***’ 0.001 ‘**’ 0.01 ‘*’ 0.05 ‘.’ 0.1 ‘ ’ 1
\end{Verbatim}
\caption{Additive model.}
\end{subtable}\\
\begin{subtable}{\textwidth}
\begin{Verbatim}[fontsize=\tiny]
Generalized linear mixed model fit by maximum likelihood (Laplace
  Approximation) [glmerMod]
 Family: binomial  ( logit )
Formula: response_correct ~ wason_condition * replication_experiment +  
    (1 | wason_name)
   Data: wason_human_correct_df

     AIC      BIC   logLik deviance df.resid 
   477.7    499.5   -233.9    467.7      569 

Scaled residuals: 
    Min      1Q  Median      3Q     Max 
-0.7198 -0.4788 -0.3544 -0.2523  4.3184 

Random effects:
 Groups     Name        Variance Std.Dev.
 wason_name (Intercept) 0.1762   0.4197  
Number of obs: 574, groups:  wason_name, 25

Fixed effects:
                                                    Estimate Std. Error z value
(Intercept)                                          -2.7723     0.4162  -6.662
wason_conditionRealistic                              1.0550     0.5078   2.078
replication_experimentTRUE                            0.7380     0.4606   1.602
wason_conditionRealistic:replication_experimentTRUE  -0.3262     0.5673  -0.575
                                                    Pr(>|z|)    
(Intercept)                                         2.71e-11 ***
wason_conditionRealistic                              0.0378 *  
replication_experimentTRUE                            0.1091    
wason_conditionRealistic:replication_experimentTRUE   0.5653    
---
Signif. codes:  0 ‘***’ 0.001 ‘**’ 0.01 ‘*’ 0.05 ‘.’ 0.1 ‘ ’ 1
\end{Verbatim}
\caption{Interaction model.}
\end{subtable}
\caption{Mixed-effects linear regressions for differences in human performance on the replication sample on the Wason task. We do observe a marginally-significant effect of the experiment in the additive model (top). However, we do not observe significant differences in the content effect in an interaction model (bottom).} \label{app:tab:wason_human_replication_regression}
\end{table*}

\FloatBarrier
\subsection{Human response time distributions on the Wason tasks} \label{app:analyses:wason_rt}

\begin{figure}[H]
    \centering
    \includegraphics[width=0.66\textwidth]{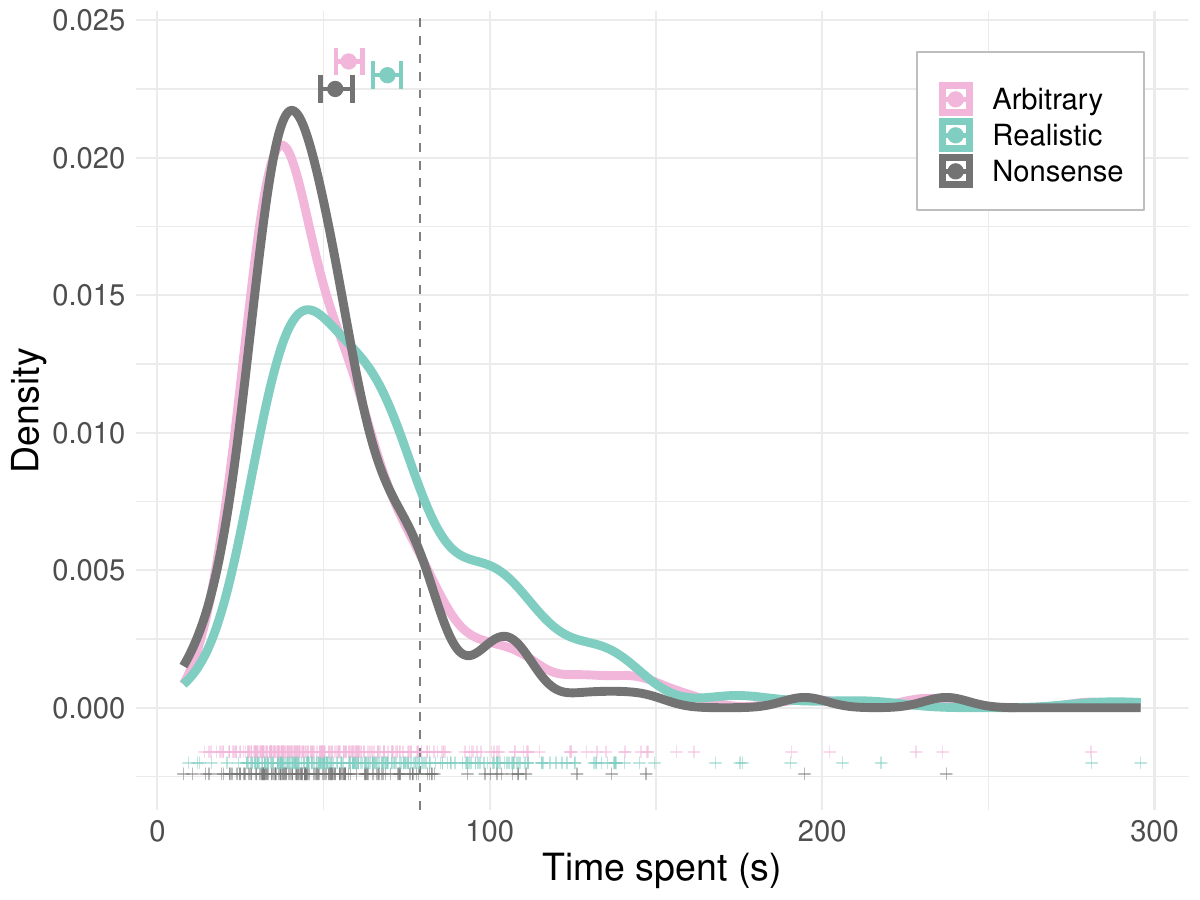}
    \caption{Human response time distributions on the Wason tasks. The Realistic condition results in significantly longer response times. The vertical dashed line indicates the cutoff for ``slow'' subject group; 85\% of the subjects were faster than this in the original experiment.}
    \label{app:fig:wason_human_rt_distributions}
\end{figure}

In Fig. \ref{app:fig:wason_human_rt_distributions} we show the distribution of response times for humans in the Wason tasks. There is a mean difference in response times, with participants spending about 12 seconds longer on Realistic questions on average. This difference may be due to the time needed to read the extra sentences giving the realistic context, or to the participants engaging more deeply with the problems that seem more sensible. However, in Appx. \ref{app:statistical_analyses:wason:human} we show that this difference alone does not explain the advantage of the Realistic conditions.

\subsubsection{Response time effects on NLI and syllogisms} \label{app:analyses:nli_syl_rt}

\begin{figure}[H]
    \centering
    \includegraphics[width=0.66\textwidth]{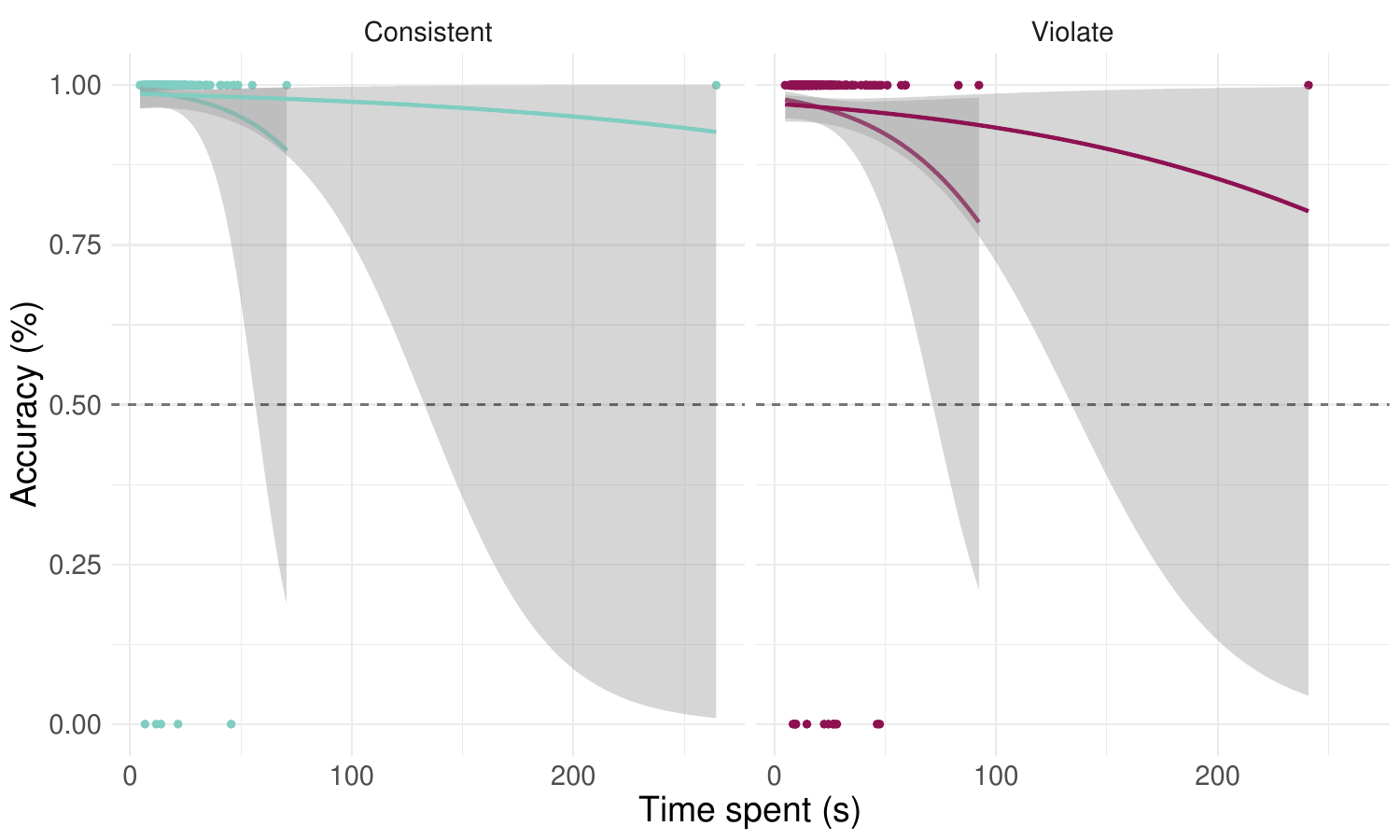}
    \caption{There is little effect of response time on accuracy in the NLI tasks.}
    \label{app:fig:acc_by_rt:nli}
\end{figure}

Given the strong effect of response time on Wason task performance, we also analyzed the effects on the NLI and Syllogism tasks (Figs. \ref{app:fig:acc_by_rt:nli} \&  \ref{app:fig:acc_by_rt:nli}; Table \ref{app:tab:acc_by_rt:syl}). In these tasks we do not see clear effects, though there are hints of an interesting potential interaction in the syllogisms task.

\begin{figure}[H]
    \centering
    \includegraphics[width=0.66\textwidth]{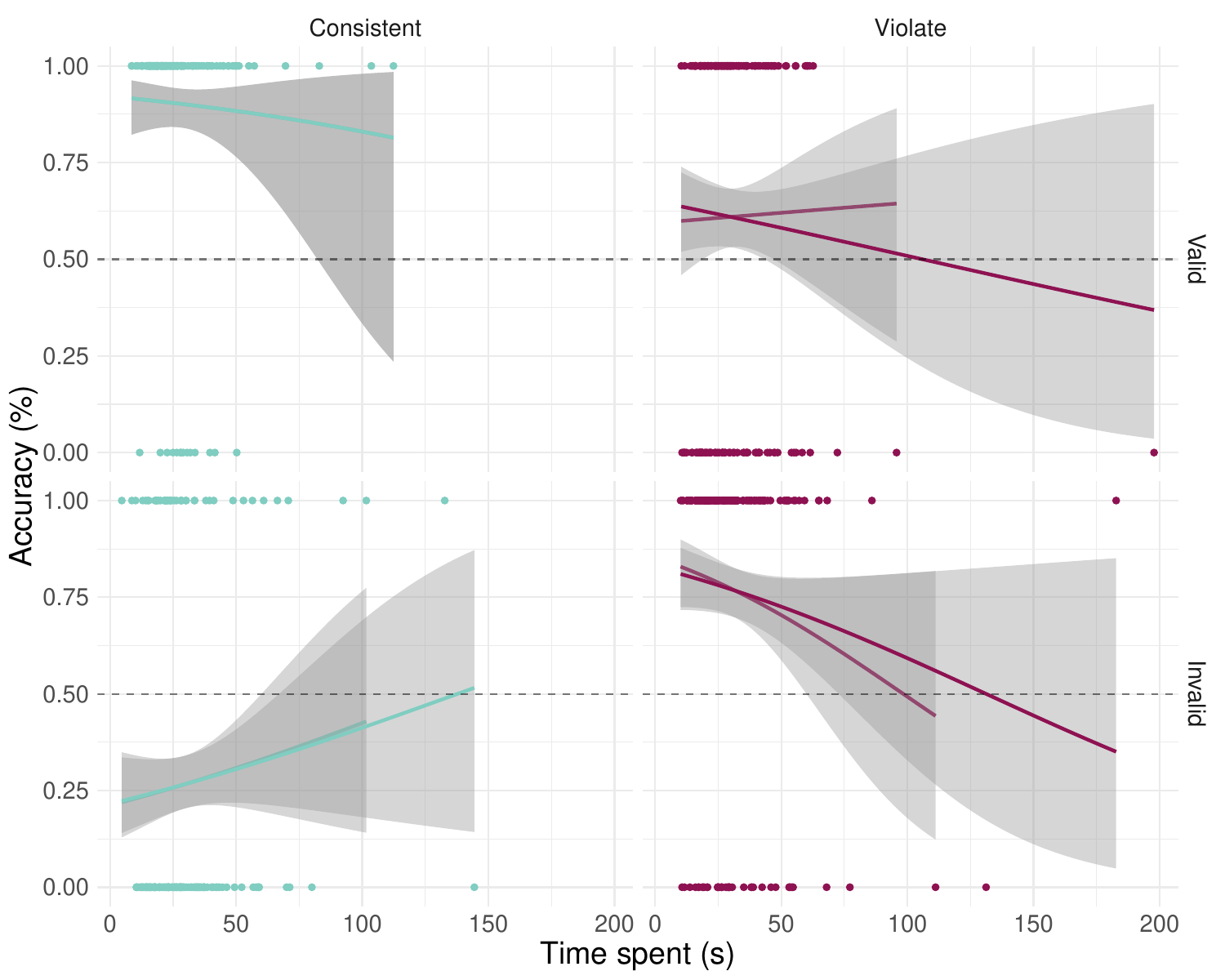}
    \caption{Effects of response time on accuracy on the syllogisms task.}
    \label{app:fig:acc_by_rt:syl}
\end{figure}

\begin{table*}[htbp]
\begin{Verbatim}[fontsize=\tiny]
Generalized linear mixed model fit by maximum likelihood (Laplace
  Approximation) [glmerMod]
 Family: binomial  ( logit )
Formula: response_correct ~ logic_belief_consistent * (scale(log(rt)) +  
    consistent_plottable) + (1 | syllogism_name)
   Data: syllogism_model_df %>% filter(subject == "Human")

     AIC      BIC   logLik deviance df.resid 
   693.4    724.6   -339.7    679.4      631 

Scaled residuals: 
    Min      1Q  Median      3Q     Max 
-4.5854 -0.6137  0.3423  0.6457  2.1597 

Random effects:
 Groups         Name        Variance Std.Dev.
 syllogism_name (Intercept) 0.08975  0.2996  
Number of obs: 638, groups:  syllogism_name, 12

Fixed effects:
                                                     Estimate Std. Error z value Pr(>|z|)    
(Intercept)                                           0.61895    0.19140   3.234  0.00122 ** 
logic_belief_consistent1                              3.16927    0.33673   9.412  < 2e-16 ***
scale(log(rt))                                       -0.08297    0.09858  -0.842  0.39998    
consistent_plottableViolate                           0.20829    0.21271   0.979  0.32746    
logic_belief_consistent1:scale(log(rt))              -0.35038    0.19413  -1.805  0.07109 .  
logic_belief_consistent1:consistent_plottableViolate -2.40845    0.42012  -5.733 9.88e-09 ***
\end{Verbatim}
\caption{Mixed-effects regression examining the continuous effect of RT on the Syllogism tasks. There is no main effect, but there is a marginally-significant interaction with the content effect, such that slower responses are more helpful on problems where content contradicts logic.} \label{app:tab:acc_by_rt:syl}
\end{table*}

\subsection{Item-level effects} \label{app:analyses:item_level} 

In this section, we perform item level analyses for each task. 

\subsubsection{NLI}
\begin{figure}[H]
    \centering
    \includegraphics[width=\textwidth]{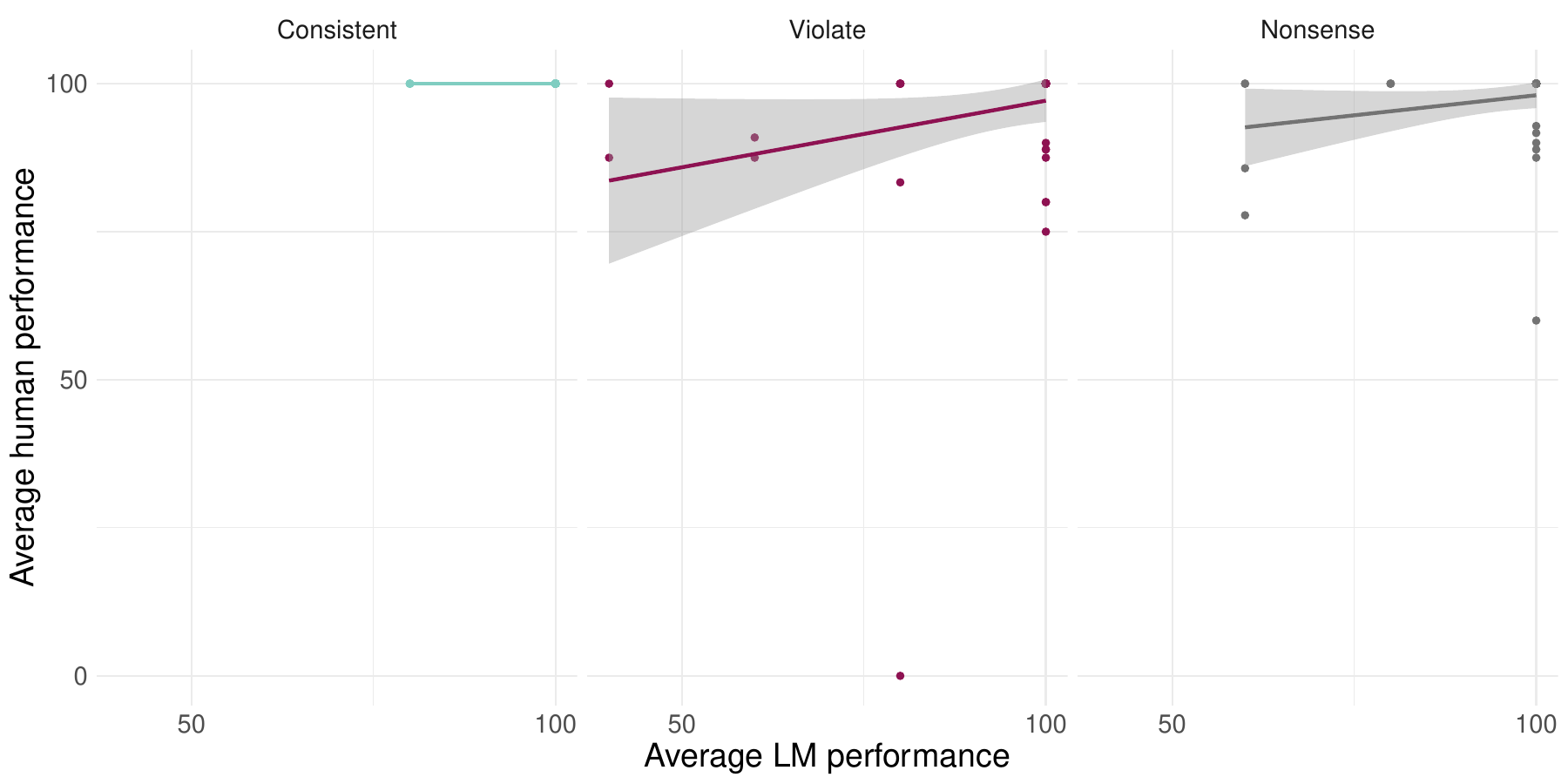}
    \caption{Association of human and average model accuracy on the NLI task.}
    \label{app:fig:item_correlation:nli}
\end{figure}

First, for the NLI task, we plot the item-level correlations in accuracy in Fig. \ref{app:fig:item_correlation:nli}. Surprisingly (given the close-to-ceiling performance), we find that Human success rates are significantly predictive of LM success rates, even when controlling for condition (Table \ref{app:tab:item_level_regression:nli}). 

\begin{table*}[htbp]
\begin{Verbatim}[fontsize=\tiny]
Linear mixed model fit by REML ['lmerMod']
Formula: LM ~ Human + consistent_plottable + (1 | model)
   Data: nli_item_level_df

REML criterion at convergence: -404.8

Scaled residuals: 
    Min      1Q  Median      3Q     Max 
-5.2460  0.0661  0.1428  0.2724  1.6139 

Random effects:
 Groups   Name        Variance  Std.Dev.
 model    (Intercept) 0.0008882 0.0298  
 Residual             0.0350993 0.1873  
Number of obs: 845, groups:  model, 5

Fixed effects:
                             Estimate Std. Error t value
(Intercept)                   0.73890    0.07288  10.139
Human                         0.24681    0.07077   3.488
consistent_plottableViolate  -0.02879    0.01560  -1.846
consistent_plottableNonsense -0.02429    0.01649  -1.473
\end{Verbatim}
\caption{Mixed-effects linear regression for item-level association of human and model accuracy on the NLI task, controlling for consistency.} \label{app:tab:item_level_regression:nli}
\end{table*}

\FloatBarrier
\subsubsection{Syllogisms}
For the Syllogisms task, we plot the item-level correlations in accuracy in Fig. \ref{app:fig:item_correlation:nli}. We again find a significant relationship between human success rates and language model success (\(t = 4.98\), \(p < 0.001\) when controlling for task variables; Table \ref{app:tab:item_level_regression:syl}).

\begin{figure}[H]
    \centering
    \includegraphics[width=\textwidth]{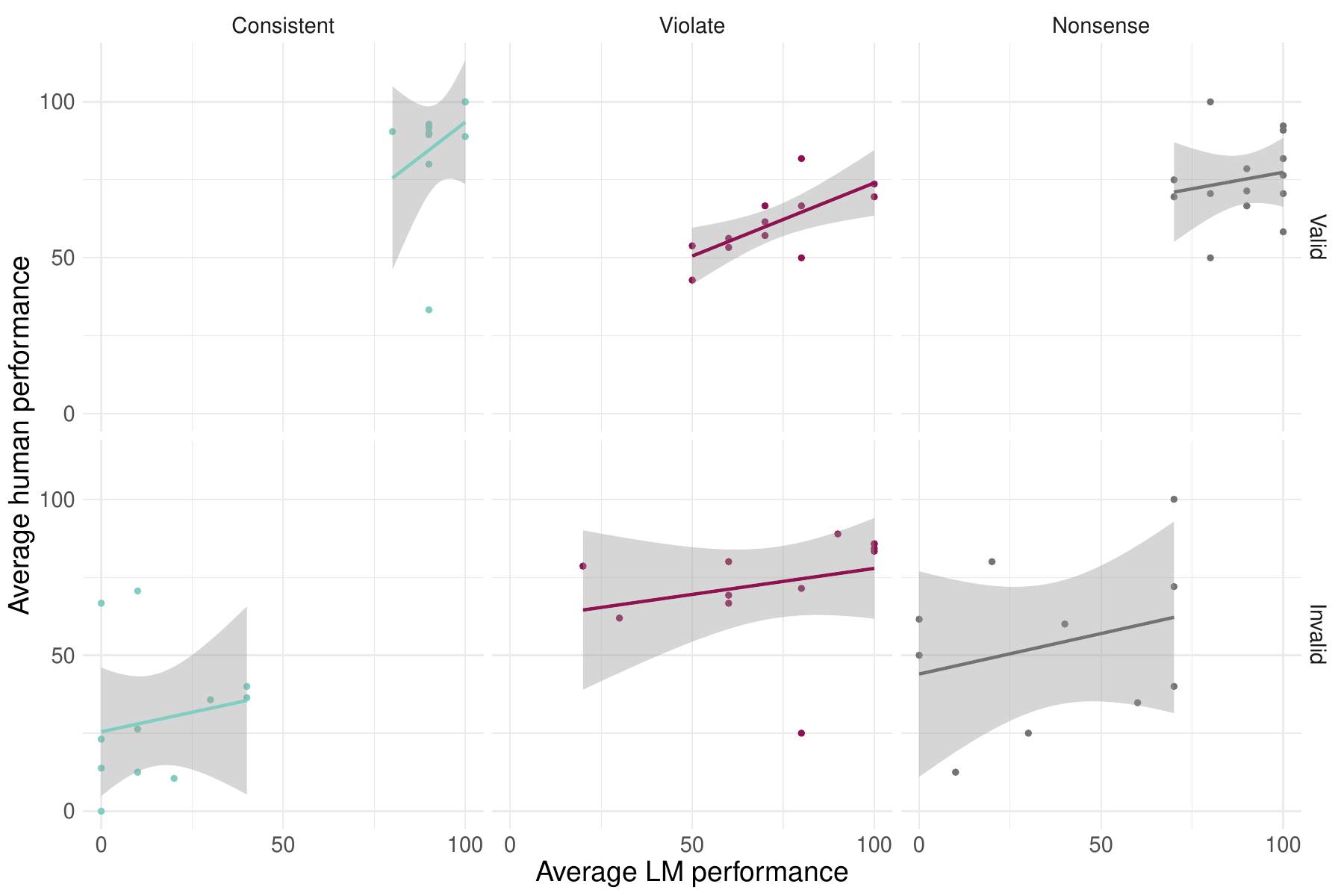}
    \caption{Association of human and average model accuracy on the Syllogisms task.}
    \label{app:fig:item_correlation:syl}
\end{figure}

\begin{table*}[htbp]
\begin{Verbatim}[fontsize=\tiny]
Linear mixed model fit by REML ['lmerMod']
Formula: LM ~ Human + logic_belief_consistent * consistent_plottable +  
    (1 | model)
   Data: syl_item_level_df

REML criterion at convergence: 313.7

Scaled residuals: 
    Min      1Q  Median      3Q     Max 
-2.5622 -0.5048  0.2668  0.6984  2.8012 

Random effects:
 Groups   Name        Variance Std.Dev.
 model    (Intercept) 0.004978 0.07056 
 Residual             0.131391 0.36248 
Number of obs: 355, groups:  model, 5

Fixed effects:
                                                    Estimate Std. Error t value
(Intercept)                                          0.23822    0.07538   3.160
Human                                                0.51089    0.10262   4.978
logic_belief_consistent                              0.24473    0.04505   5.432
consistent_plottableViolate                          0.14737    0.04827   3.053
consistent_plottableNonsense                         0.09873    0.04806   2.054
logic_belief_consistent:consistent_plottableViolate -0.27191    0.05281  -5.148
\end{Verbatim}
\caption{Mixed-effects linear regression for item-level association of human and model accuracy on the Syllogisms task, controlling for content and logic.} \label{app:tab:item_level_regression:syl}
\end{table*}

\FloatBarrier
\subsubsection{Wason} \label{app:analyses:wason_items}

For the Wason task, we plot the item-level correlations in accuracy in Fig. \ref{app:fig:item_correlation:nli}. Perhaps because human performance is low overall, we do not observe a significant relationship between human success rates and language model success (Table \ref{app:tab:item_level_regression:wason}).

\begin{figure}[p]
    \centering
    \includegraphics[width=\textwidth]{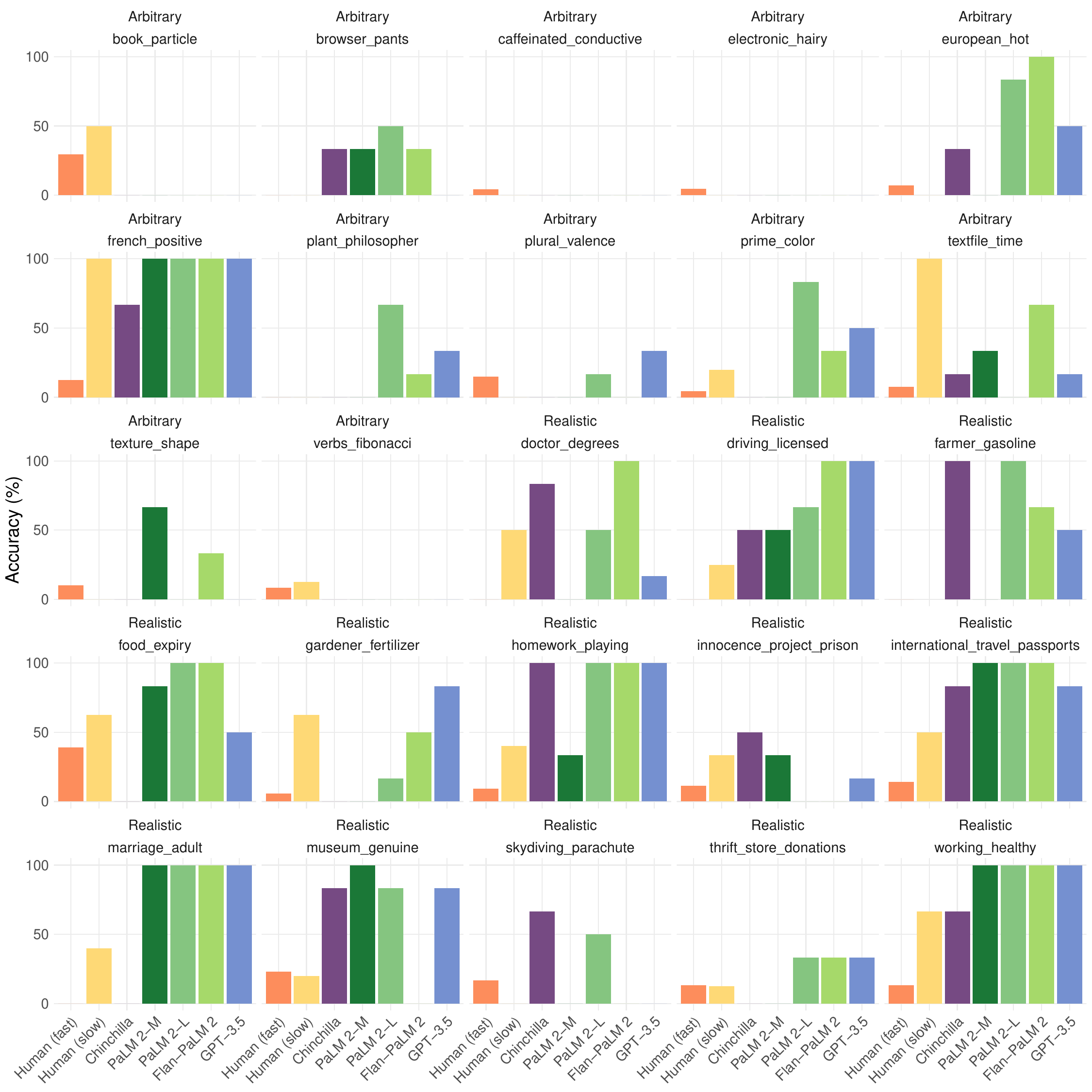}
    \caption{Accuracy of humans and each model on each rule for the Wason tasks. Note that due to sampling variability, the number of human participants who experienced each rule varies, particularly for the slow subjects. There are various suggestive patterns, including an arbitrary rule (\texttt{french\_positive}) that models and slower humans perform quite well on, and realistic rules (like \texttt{skydiving\_parachute}) that all perform surprisingly poorly on. (Note that the variation within a model comes from testing on multiple variations of each problem, with different card orders and card names; see Appx. \ref{app:methods:datasets}.)}
    \label{app:fig:wason_accuracy_per_item}
\end{figure}

Due to the item-level effects observed in some of the main regressions, we also plot performance of each model or human group on each of the Wason rules in Fig. \ref{app:fig:wason_accuracy_per_item}. Overall, the variability seems mostly as expected. However, there are some interesting patterns, including one arbitrary rule that most subject perform well on. That particular rule is:

\begin{Verbatim}[fontsize=\tiny]

The rule is that if the cards have a French word then they must have a positive number.
chapeau / sombrero / 4 / -1
\end{Verbatim}

It is not particularly apparent to us why this rule might be easier.

\begin{figure}[H]
    \centering
    \includegraphics[width=0.66\textwidth]{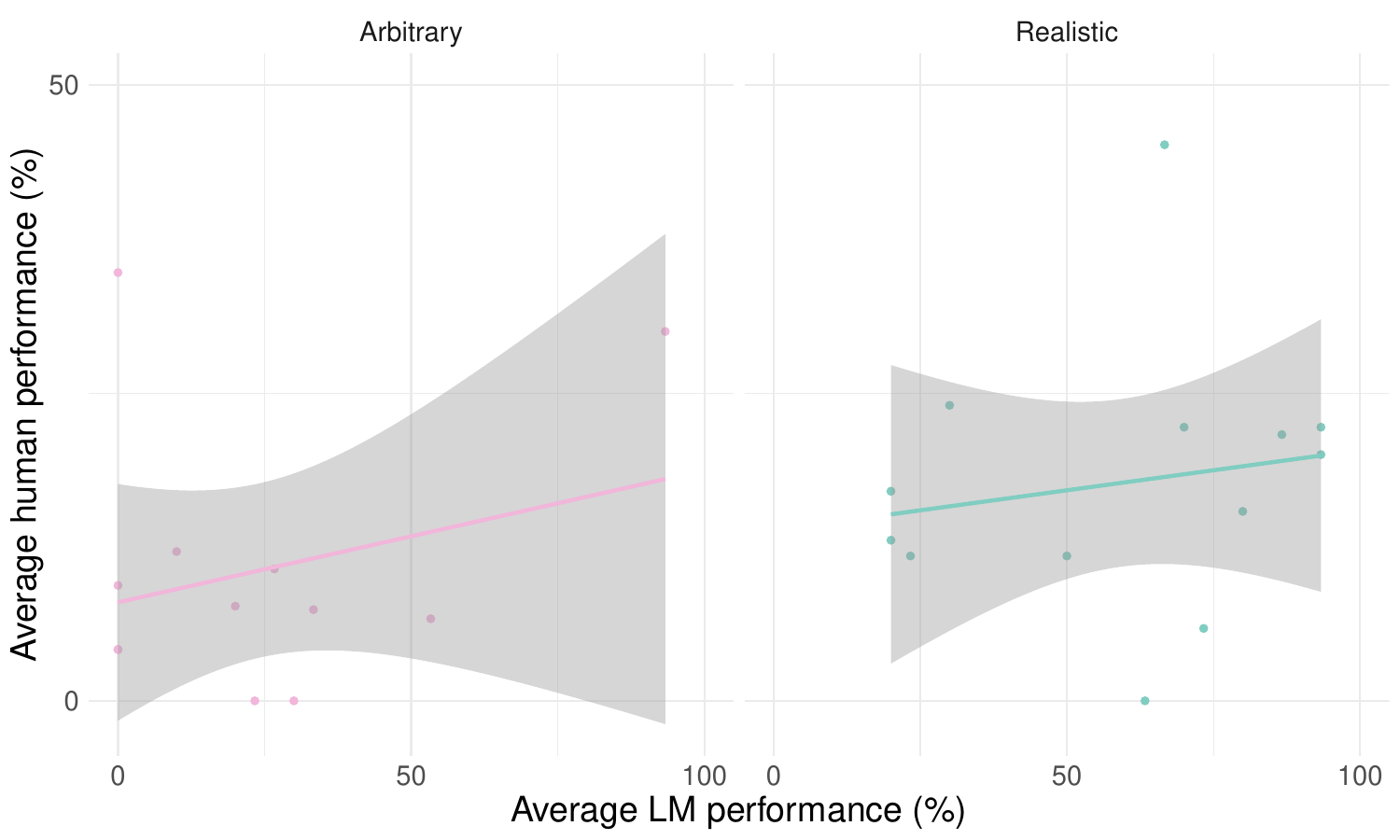}
    \caption{Association of human and average model accuracy on the Wason task. Note the vertical axis scale---human performance is low overall.}
    \label{app:fig:item_correlation:wason}
\end{figure}

\begin{table*}[h]
\begin{Verbatim}[fontsize=\tiny]
Linear mixed model fit by REML ['lmerMod']
Formula: LM ~ Human + wason_condition + (1 | model)
   Data: wason_item_level_df

REML criterion at convergence: 181.3

Scaled residuals: 
    Min      1Q  Median      3Q     Max 
-1.7748 -0.7994 -0.1560  0.8760  1.9239 

Random effects:
 Groups   Name        Variance Std.Dev.
 model    (Intercept) 0.006414 0.08009 
 Residual             0.145352 0.38125 
Number of obs: 185, groups:  model, 5

Fixed effects:
                         Estimate Std. Error t value
(Intercept)               0.20251    0.06860   2.952
Human                     0.36974    0.29860   1.238
wason_conditionRealistic  0.32436    0.07148   4.538
wason_conditionNonsense   0.19533    0.06967   2.804
\end{Verbatim}
\caption{Mixed-effects linear regression for item-level association of human and model accuracy on the Syllogisms task, controlling for content and logic.} \label{app:tab:item_level_regression:wason}
\end{table*}

%%%%%%%%%%%%%%%%%% 
\FloatBarrier
\subsection{Model answer log-probability distributions} \label{app:analyses:answer_logprob_dists}

\begin{figure}[H]
    \centering
    \begin{subfigure}{\textwidth}
    \centering
    \includegraphics[width=0.66\textwidth]{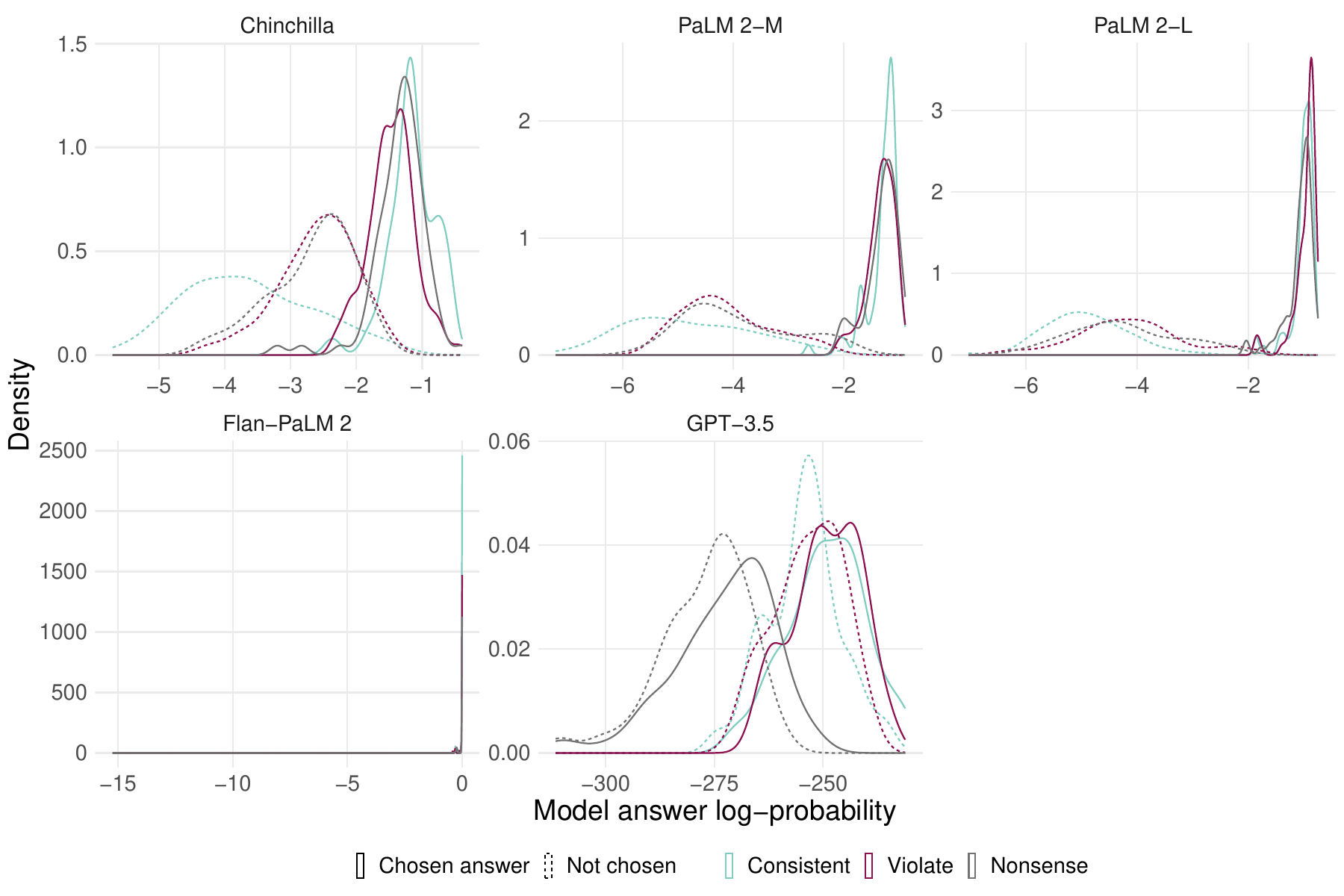}
    \caption{Raw log-probabilities.} \label{app:fig:nli_logprob_dists:raw}
    \end{subfigure}\\
    \begin{subfigure}{\textwidth}
    \centering
    \includegraphics[width=0.66\textwidth]{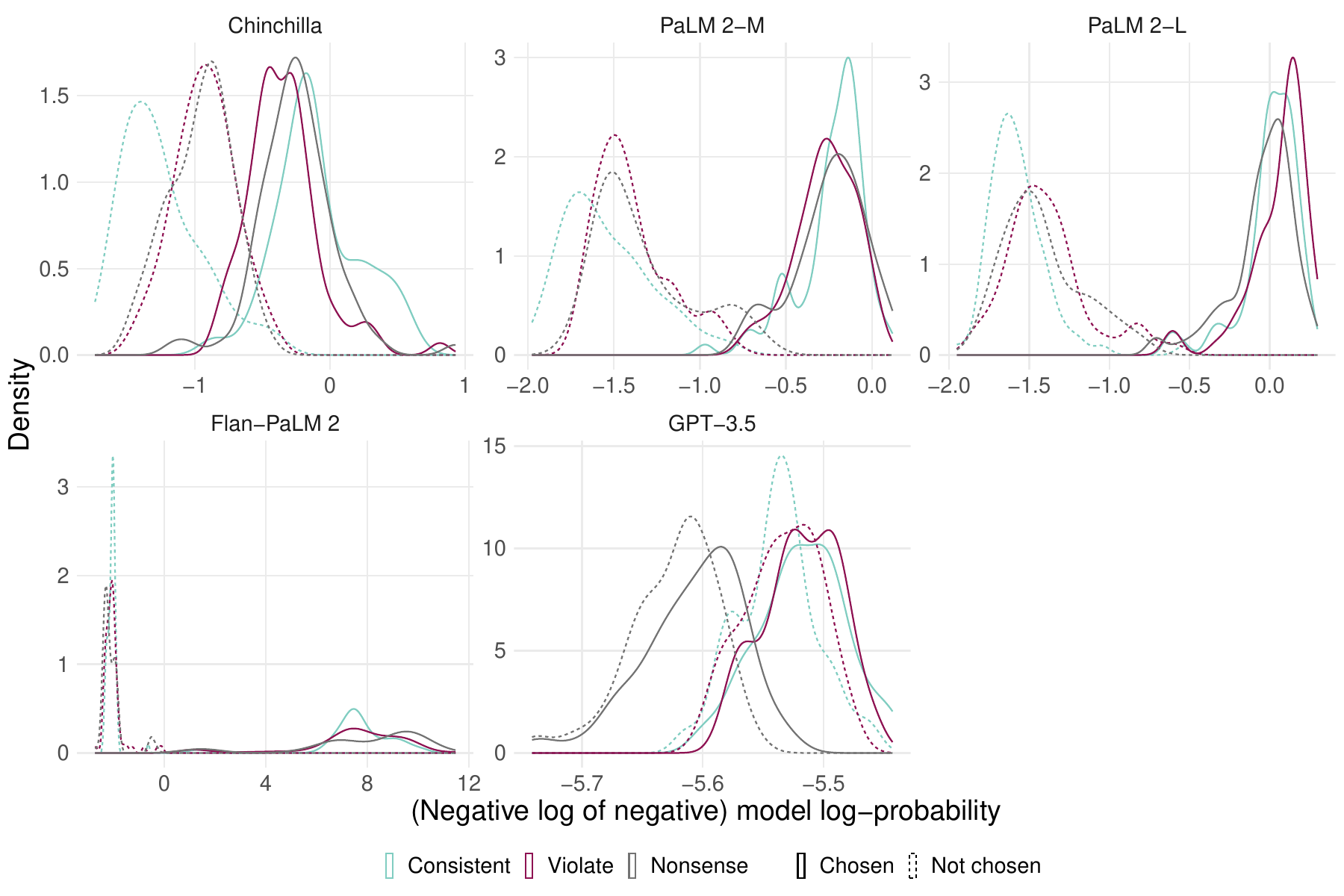}
    \caption{Log transformed.} \label{app:fig:nli_logprob_dists:double_logged}
    \end{subfigure}\\
    \begin{subfigure}{\textwidth}
    \centering
    \end{subfigure}\\
    \caption{Model log-probability distributions for the answer choices on the Natural Language Inference (NLI) task. We visualize these in two ways: (\subref{app:fig:nli_logprob_dists:raw}) the raw log-probabilities, and (\subref{app:fig:nli_logprob_dists:raw}) the negative log of the negative log-probabilities --- this transform makes the distribution for Flan-PaLM 2 clearer. Across both plots, there is fairly clear separation between the distributions of chosen and unchosen answers for most models. There are various interesting effects of content on the log-probabilities, e.g. changes in the mean and variance of the distributions. There are also striking differences among the models, possibly hinting at the effects of different training processes.}
    \label{app:fig:nli_logprob_dists}
\end{figure}

\begin{figure}[H]
    \centering
    \includegraphics[width=\textwidth]{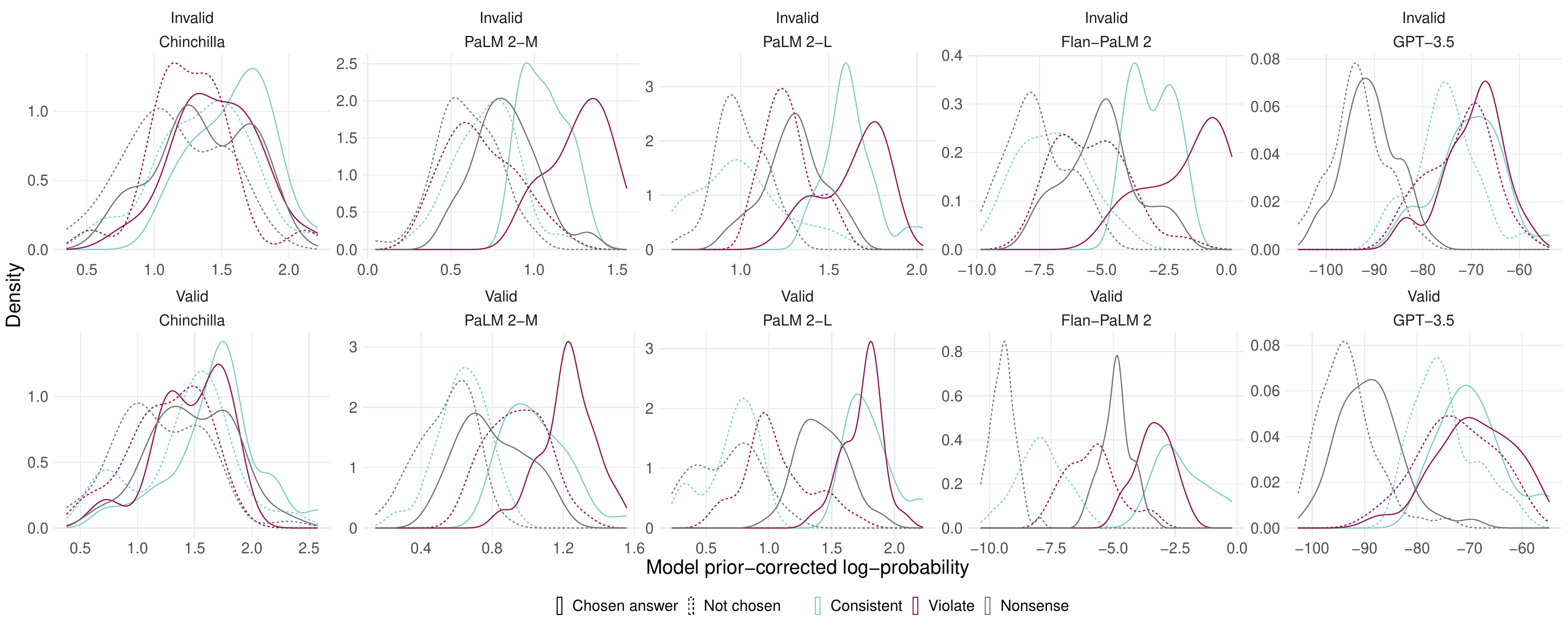}
    \caption{Model prior-corrected log-probability distributions for the answer choices on the syllogisms task. The degree of separation between the distributions depends on the model, validity, and content. Again, there are differences among the models. For example, larger models show more cleanly separated distributions (PaLM 2-L vs. -M), and the instruction tuned models (Flan-PaLM 2 and GPT-3.5) show much larger magnitude prior corrected log probabilities.
    }
    \label{app:fig:syllogism_logprob_dists}
\end{figure}

\begin{figure}[H]
    \centering
    \includegraphics[width=0.66\textwidth]{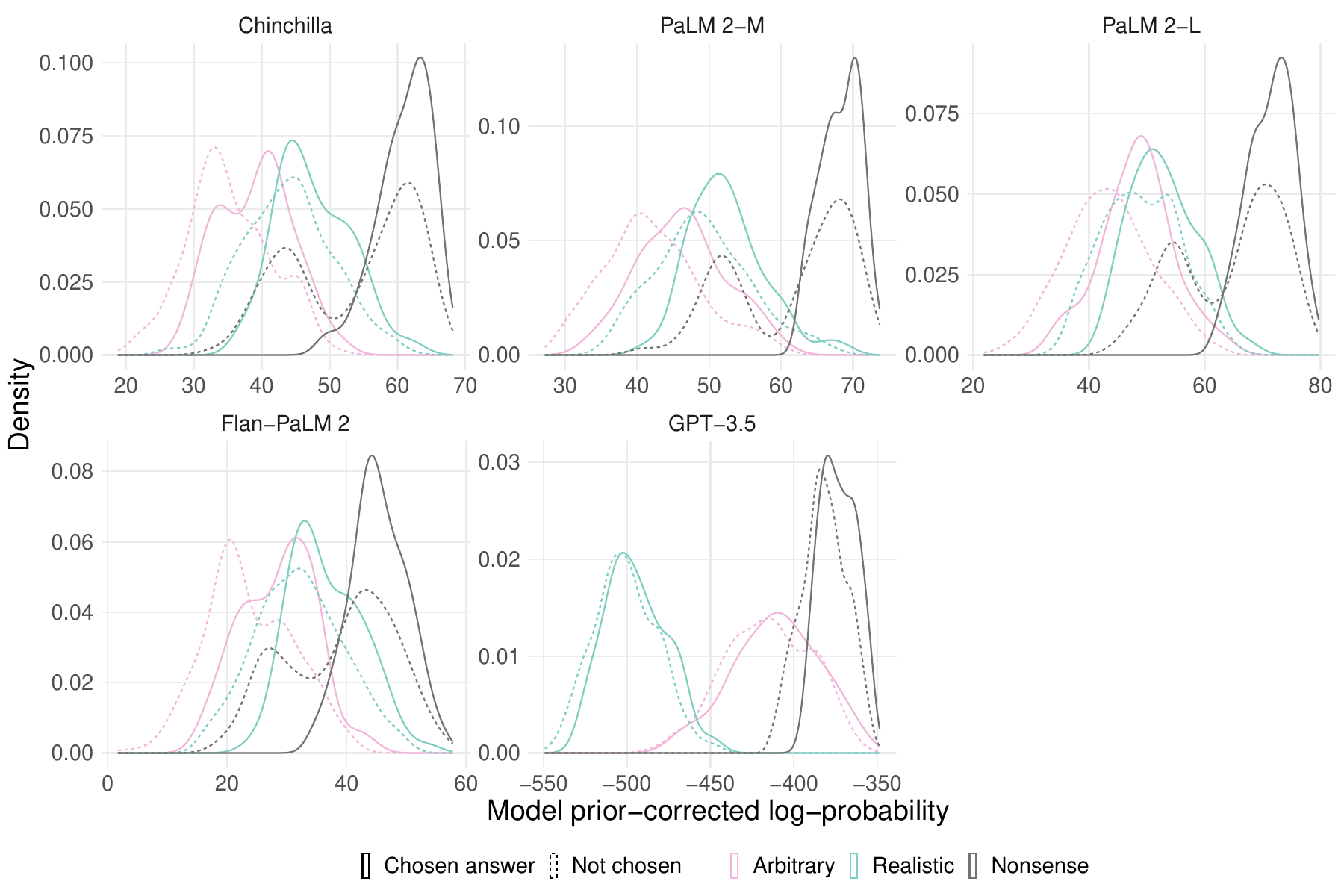}
    \caption{Model prior-corrected log-probability distributions for the answer choices on the Wason selection task. The degree of separation between the chosen and not-chosen answer distributions is generally lower than in other tasks, possibly reflecting the greater difficulty of the Wason task, or the greater problem-to-problem variability. By contrast, the separation by content is striking for some models, e.g. GPT-3.5. 
    }
    \label{app:fig:wason_logprob_dists}
\end{figure}

In this section we plot the log-probability distributions of the models on the different tasks (Figs. \ref{app:fig:nli_logprob_dists}, \ref{app:fig:syllogism_logprob_dists}, \ref{app:fig:wason_logprob_dists}). There are a variety of interesting effects of task variables, and some striking differences among the models.

For example, the instruction-tuned models (Flan-PaLM 2 and GPT-3.5) have numerically much greater magnitude log-probabilities to the answers, especially GPT-3.5. This may be an artifact of the tuning process. Furthermore, the larger models tend to show clearer separation between the chosen answer and the others (e.g., comparing PaLM 2-L to -M).

\subsection{Chinchilla can identify the valid conclusion of a syllogism from among all possible conclusions with high accuracy} \label{app:analyses:syllogsims_conclusions}

\begin{figure}[H]
    \centering
    \begin{subfigure}{0.5\textwidth}
    \includegraphics[width=\textwidth]{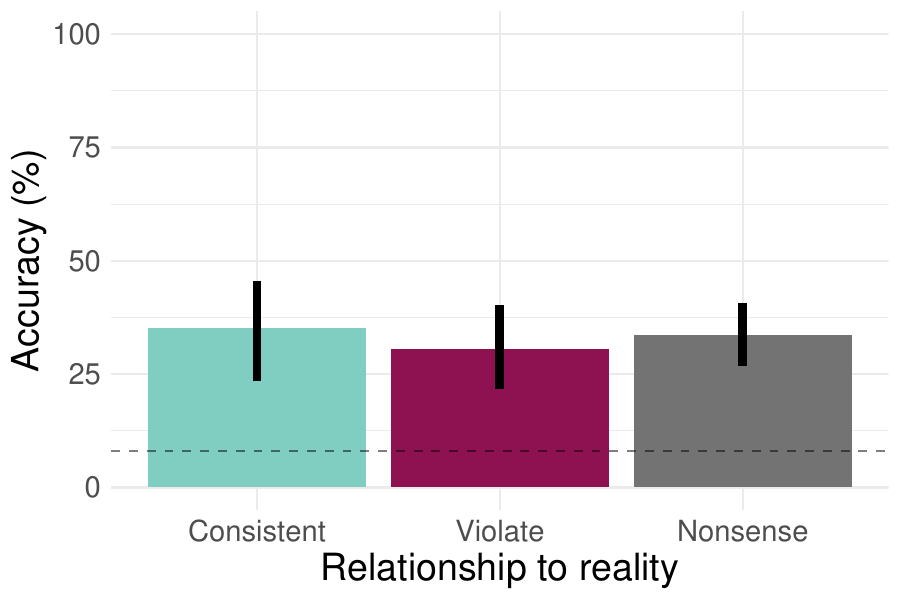}
    \caption{Domain-conditional PMI scoring}
    \end{subfigure}%
    \begin{subfigure}{0.5\textwidth}
    \includegraphics[width=\textwidth]{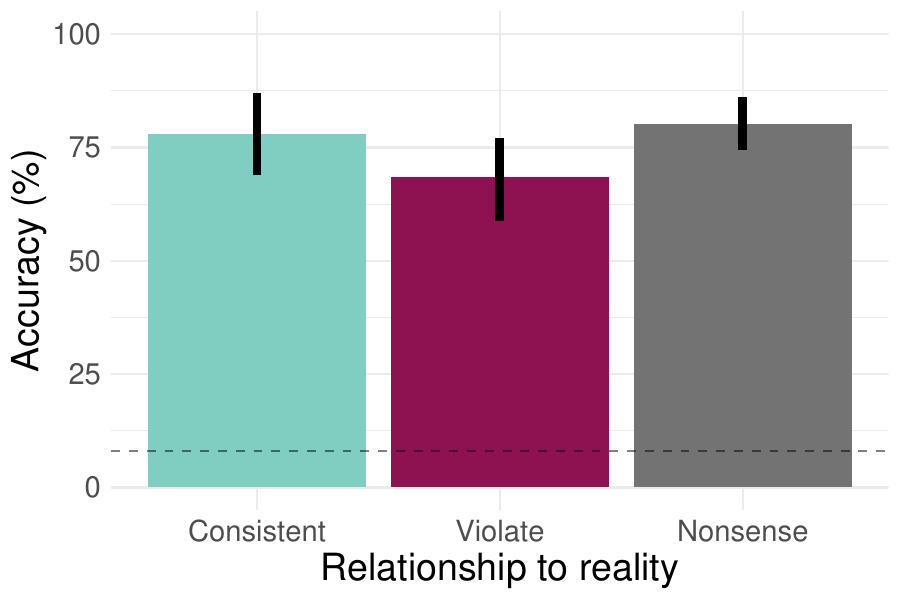}
    \caption{Raw log-likelihood scoring}
    \end{subfigure}%
    \caption{Chinchilla's zero-shot accuracy at identifying the correct conclusion to a syllogism among all possible conclusions. The model exhibits far above chance performance (especially when scoring with raw log-likelihoods), and relatively weaker bias with this task design.}
    \label{app:fig:syll_conc_zero_shot}  
\end{figure}

In Fig. \ref{app:fig:syll_conc_zero_shot} we show the accuracy of Chinchilla when choosing from among all possible predicates containing one of the quantifiers used and two of the entities appearing in the premises of the syllogism. The model exhibits high accuracy across conditions, and relatively little bias (though bias increases few shot). This observation is reminiscent of the finding of Trippas et al. \citep{trippas2014using} that humans exhibit less bias when making a forced choice among two possible arguments (one valid and one invalid) rather than deciding if a single syllogism is valid or invalid.

Note that in this case scoring with the Domain-Conditional PMI \citep{holtzman2021surface}---which we used for the main Syllogisms and Wason results---produces much \emph{lower} accuracy than the raw likelihoods, and minor differences in bias. The patterns are qualitatively similar with or without the correction, but accuracy is lower without (around 35-40\%) regardless of belief consistency.

%%%%%%%%%%%%%%%%%%%%%%%%%%%%%%%%%%%%%%%%%%%%%%%%%%%%%%%%%%%%

\FloatBarrier
\section{Statistical analyses} \label{app:statistical_analyses}

In this section, we provide the full results for all statistical analyses reported in the main text. We generally report results from mixed-effects logistic regressions, controlling for the random effects of the different stimuli used.\footnote{Unless otherwise noted, we conservatively approximate the degrees of freedom for all \(t\)-tests by treating all random effects as though they were fixed effects (i.e. by subtracting the number of levels of each random variable from the residual degrees of freedom), rather than using a variance-based approximation.}

\subsection{NLI} \label{app:statistical_analyses:nli}

We report statistical analyses of content effects on the NLI tasks for humans and all models in Tables \ref{app:tab:nli_stats:Human}-\ref{app:tab:nli_stats:GPT-3.5}. We generally fit mixed effects logistic regressions, but the regressions for PaLM 2-L and Flan-PaLM 2 failed to converge due to ceiling effects. We therefore also report \(\chi^2\) tests of the difference in correct responses across conditions. In all cases, we do not find a significant content effect on this simple task.

\begin{table*}[htbp]
\begin{subtable}{\textwidth}
\begin{Verbatim}[fontsize=\tiny]
Generalized linear mixed model fit by maximum likelihood (Laplace
  Approximation) [glmerMod]
 Family: binomial  ( logit )
Formula: response_correct ~ consistent_plottable + (1 | name)
   Data: 
nli_joint_df %>% filter(subject == "Human", consistent_plottable !=  
    "Nonsense")

     AIC      BIC   logLik deviance df.resid 
   133.2    146.8    -63.6    127.2      677 

Scaled residuals: 
    Min      1Q  Median      3Q     Max 
-3.5119  0.0105  0.0106  0.0282  0.4931 

Random effects:
 Groups Name        Variance Std.Dev.
 name   (Intercept) 25.82    5.081   
Number of obs: 680, groups:  name, 122

Fixed effects:
                            Estimate Std. Error z value Pr(>|z|)    
(Intercept)                    9.082      2.072   4.384 1.17e-05 ***
consistent_plottableViolate   -2.051      1.608  -1.276    0.202    
---
Signif. codes:  0 ‘***’ 0.001 ‘**’ 0.01 ‘*’ 0.05 ‘.’ 0.1 ‘ ’ 1
\end{Verbatim}
\caption{Mixed-effects logistic regression.} \label{app:tab:nli_stats:Human:regression}
\end{subtable}\\
\begin{subtable}{\textwidth}
\begin{Verbatim}[fontsize=\tiny]
Chi-squared test for given probabilities
X-squared = 0.33937, df = 1, p-value = 0.5602
\end{Verbatim}
\caption{\(\chi^2\) test.} \label{app:tab:nli_stats:Human:chisq}
\end{subtable}
\caption{Statistical analyses of human performance on the NLI tasks, using (\subref{app:tab:nli_stats:Human:regression}) a mixed-effects logistic regression or (\subref{app:tab:nli_stats:Human:chisq}) a \(\chi^2\) test. There are no significant content effects.} \label{app:tab:nli_stats:Human}
\end{table*}

\begin{table*}[htbp]
\begin{subtable}{\textwidth}
\begin{Verbatim}[fontsize=\tiny]
Generalized linear mixed model fit by maximum likelihood (Laplace
  Approximation) [glmerMod]
 Family: binomial  ( logit )
Formula: response_correct ~ consistent_plottable + (1 | name)
   Data: 
nli_joint_df %>% filter(subject == "Chinchilla", consistent_plottable !=  
    "Nonsense")

     AIC      BIC   logLik deviance df.resid 
    46.6     55.8    -20.3     40.6      153 

Scaled residuals: 
      Min        1Q    Median        3Q       Max 
-0.084635  0.000969  0.000969  0.001903  0.001903 

Random effects:
 Groups Name        Variance Std.Dev.
 name   (Intercept) 2646     51.43   
Number of obs: 156, groups:  name, 153

Fixed effects:
                            Estimate Std. Error z value Pr(>|z|)    
(Intercept)                   13.877      3.373   4.114  3.9e-05 ***
consistent_plottableViolate   -1.357      3.760  -0.361    0.718    
---
Signif. codes:  0 ‘***’ 0.001 ‘**’ 0.01 ‘*’ 0.05 ‘.’ 0.1 ‘ ’ 1
\end{Verbatim}
\caption{Mixed-effects logistic regression.} \label{app:tab:nli_stats:Chinchilla:regression}
\end{subtable}\\
\begin{subtable}{\textwidth}
\begin{Verbatim}[fontsize=\tiny]
Chi-squared test for given probabilities
X-squared = 0.34266, df = 1, p-value = 0.5583
\end{Verbatim}
\caption{\(\chi^2\) test.} \label{app:tab:nli_stats:Chinchilla:chisq}
\end{subtable}
\caption{Statistical analyses of Chinchilla's performance on the NLI tasks, using (\subref{app:tab:nli_stats:Chinchilla:regression}) a mixed-effects logistic regression or (\subref{app:tab:nli_stats:Chinchilla:chisq}) a \(\chi^2\) test. There are no significant content effects.}
\label{app:tab:nli_stats:Chinchilla}
\end{table*}

\begin{table*}[htbp]
\begin{subtable}{\textwidth}
\begin{Verbatim}[fontsize=\tiny]
Generalized linear mixed model fit by maximum likelihood (Laplace
  Approximation) [glmerMod]
 Family: binomial  ( logit )
Formula: response_correct ~ consistent_plottable + (1 | name)
   Data: 
nli_joint_df %>% filter(subject == "PaLM 2-M", consistent_plottable !=  
    "Nonsense")

     AIC      BIC   logLik deviance df.resid 
    22.0     31.1     -8.0     16.0      153 

Scaled residuals: 
      Min        1Q    Median        3Q       Max 
-0.074976  0.000648  0.000648  0.000804  0.000804 

Random effects:
 Groups Name        Variance Std.Dev.
 name   (Intercept) 3553     59.61   
Number of obs: 156, groups:  name, 153

Fixed effects:
                            Estimate Std. Error z value Pr(>|z|)    
(Intercept)                  14.2497     3.5260   4.041 5.31e-05 ***
consistent_plottableViolate   0.4323     5.5651   0.078    0.938    
---
Signif. codes:  0 ‘***’ 0.001 ‘**’ 0.01 ‘*’ 0.05 ‘.’ 0.1 ‘ ’ 1

Correlation of Fixed Effects:
            (Intr)
cnsstnt_plV -0.629
\end{Verbatim}
\caption{Mixed-effects logistic regression.} \label{app:tab:nli_stats:PaLM2-M:regression}
\end{subtable}\\
\begin{subtable}{\textwidth}
\begin{Verbatim}[fontsize=\tiny]
Chi-squared test for given probabilities
X-squared = 0.0066225, df = 1, p-value = 0.9351
\end{Verbatim}
\caption{\(\chi^2\) test.} \label{app:tab:nli_stats:PaLM2-M:chisq}
\end{subtable}
\caption{Statistical analyses of PaLM 2-M's performance on the NLI tasks, using (\subref{app:tab:nli_stats:PaLM2-M:regression}) a mixed-effects logistic regression or (\subref{app:tab:nli_stats:PaLM2-M:chisq}) a \(\chi^2\) test. There are no significant content effects.} \label{app:tab:nli_stats:PaLM2-M}
\end{table*}

\begin{table*}[htbp]
\begin{Verbatim}[fontsize=\tiny]
Chi-squared test for given probabilities
X-squared = 0, df = 1, p-value = 1
\end{Verbatim}
\caption{Statistical analysis of PaLM 2-L's performance on the NLI tasks, using a \(\chi^2\) test, as the logistic regression failed to converge. There are no significant content effects.} \label{app:tab:nli_stats:PaLM2-L}
\end{table*}

\begin{table*}[htbp]
\begin{Verbatim}[fontsize=\tiny]
Chi-squared test for given probabilities
X-squared = 0.0064516, df = 1, p-value = 0.936
\end{Verbatim}
\caption{Statistical analysis of Flan-PaLM 2's performance on the NLI tasks, using a \(\chi^2\) test, as the logistic regression failed to converge. There are no significant content effects.} \label{app:tab:nli_stats:Flan-PaLM2}
\end{table*}

\begin{table*}[htbp]
\begin{subtable}{\textwidth}
\begin{Verbatim}[fontsize=\tiny]
Generalized linear mixed model fit by maximum likelihood (Laplace
  Approximation) [glmerMod]
 Family: binomial  ( logit )
Formula: response_correct ~ consistent_plottable + (1 | name)
   Data: 
nli_joint_df %>% filter(subject == "GPT-3.5", consistent_plottable !=  
    "Nonsense")

     AIC      BIC   logLik deviance df.resid 
    31.3     40.5    -12.7     25.3      153 

Scaled residuals: 
      Min        1Q    Median        3Q       Max 
-0.078288  0.000867  0.000867  0.001145  0.001145 

Random effects:
 Groups Name        Variance Std.Dev.
 name   (Intercept) 3151     56.13   
Number of obs: 156, groups:  name, 153

Fixed effects:
                            Estimate Std. Error z value Pr(>|z|)    
(Intercept)                  14.0988     3.3443   4.216 2.49e-05 ***
consistent_plottableViolate  -0.5577     4.1579  -0.134    0.893    
---
Signif. codes:  0 ‘***’ 0.001 ‘**’ 0.01 ‘*’ 0.05 ‘.’ 0.1 ‘ ’ 1
\end{Verbatim}
\caption{Mixed-effects logistic regression.} \label{app:tab:nli_stats:GPT-3.5:regression}
\end{subtable}\\
\begin{subtable}{\textwidth}
\begin{Verbatim}[fontsize=\tiny]
Chi-squared test for given probabilities
X-squared = 0.027027, df = 1, p-value = 0.8694
\end{Verbatim}
\caption{\(\chi^2\) test.} \label{app:tab:nli_stats:GPT-3.5:chisq}
\end{subtable}
\caption{Statistical analyses of GPT-3.5-turbo-instruct's performance on the NLI tasks, using (\subref{app:tab:nli_stats:GPT-3.5:regression}) a mixed-effects logistic regression or (\subref{app:tab:nli_stats:GPT-3.5:chisq}) a \(\chi^2\) test. There are no significant content effects.}
\label{app:tab:nli_stats:GPT-3.5}
\end{table*}

\FloatBarrier
\subsection{Syllogisms} \label{app:statistical_analyses:syllogisms}

We report mixed effects logistic regressions for humans and all models in Tables \ref{app:tab:syl_stats:Human}-\ref{app:tab:syl_stats:GPT-3.5}. We analyze these results using a variable which corresponds to the main content effect \\
(\verb|logic_belief_consistent|), which is 1 when the logical answer matches the believability of the conclusion --- i.e. when the argument is valid and the conclusion is believable, or the argument is invalid and the conclusion is unbelievable --- and 0 when there is a mismatch. This measure corresponds to the difference score reported in Fig. \ref{fig:summary:syllogisms}. We ran three nested models for humans and each language model --- one regression only incorporating the content effect predictor (whether the logic matches the consistency), another adding consistency condition, and a third adding the interaction of the two. 
\begin{Verbatim}[fontsize=\tiny]
response_correct ~ logic_belief_consistent + (1 | syllogism_name)
response_correct ~ logic_belief_consistent + consistent_plottable_f + (1 | syllogism_name)
response_correct ~ logic_belief_consistent * consistent_plottable_f + (1 | syllogism_name)
\end{Verbatim}
For humans and each language model, we report the best-fitting regression, measured by the BIC (and omitting models which failed to converge). However, since the interaction effect is theoretically interesting \citep[e.g.][]{dube2010assessing}, and several of the interaction models fail to converge, we also report two-way \(\chi^2\) tests of the interactions for each model. For PaLM 2-L all regressions failed to converge due to ceiling effects; thus we also report a \(\chi^2\) test of the content effect for this model only. All models show a significant content effect; all except Chinchilla and PaLM 2-M show a significant interaction.

\begin{table*}[htbp]
\begin{subtable}{\textwidth}
\begin{Verbatim}[fontsize=\tiny]
Generalized linear mixed model fit by maximum likelihood (Laplace
  Approximation) [glmerMod]
 Family: binomial  ( logit )
Formula: response_correct ~ logic_belief_consistent * consistent_plottable_f +  
    (1 | syllogism_name)
   Data: syllogism_model_df %>% filter(subject == this_subject)

     AIC      BIC   logLik deviance df.resid 
   692.8    715.1   -341.4    682.8      633 

Scaled residuals: 
    Min      1Q  Median      3Q     Max 
-3.7130 -0.6097  0.3328  0.6018  1.9316 

Random effects:
 Groups         Name        Variance Std.Dev.
 syllogism_name (Intercept) 0.08992  0.2999  
Number of obs: 638, groups:  syllogism_name, 12

Fixed effects:
                                                 Estimate Std. Error z value Pr(>|z|)    
(Intercept)                                        0.7182     0.1359   5.286 1.25e-07 ***
logic_belief_consistent1                           1.9502     0.2081   9.372  < 2e-16 ***
consistent_plottable_f1                            0.1863     0.2114   0.881    0.378    
logic_belief_consistent1:consistent_plottable_f1  -2.4800     0.4172  -5.945 2.76e-09 ***
---
Signif. codes:  0 ‘***’ 0.001 ‘**’ 0.01 ‘*’ 0.05 ‘.’ 0.1 ‘ ’ 1
\end{Verbatim}
\caption{Mixed-effects logistic regression.} \label{app:tab:syl_stats:Human:regression}
\end{subtable}\\
\begin{subtable}{\textwidth}
\begin{Verbatim}[fontsize=\tiny]
Pearson's Chi-squared test with Yates' continuity correction
X-squared = 11.402, df = 1, p-value = 0.0007338
\end{Verbatim}
\caption{\(\chi^2\) test of interaction.} \label{app:tab:syl_stats:Human:chisq}
\end{subtable}
\caption{Statistical analyses of human performance on the Syllogism tasks, using (\subref{app:tab:syl_stats:Human:regression}) a mixed-effects logistic regression or (\subref{app:tab:syl_stats:Human:chisq}) a \(\chi^2\) test of the interaction effect. There is a significant content effect and a significant interaction effect, that is, different sensitivity to logic in the Consistent compared to Violate conditions.} \label{app:tab:syl_stats:Human}
\end{table*}

\begin{table*}[htbp]
\begin{subtable}{\textwidth}
\begin{Verbatim}[fontsize=\tiny]
Generalized linear mixed model fit by maximum likelihood (Laplace
  Approximation) [glmerMod]
 Family: binomial  ( logit )
Formula: response_correct ~ logic_belief_consistent + (1 | syllogism_name)
   Data: syllogism_model_df %>% filter(subject == this_subject)

     AIC      BIC   logLik deviance df.resid 
   125.8    133.5    -59.9    119.8       93 

Scaled residuals: 
    Min      1Q  Median      3Q     Max 
-1.7321 -0.8819  0.5774  0.5774  1.1339 

Random effects:
 Groups         Name        Variance Std.Dev.
 syllogism_name (Intercept) 0        0       
Number of obs: 96, groups:  syllogism_name, 12

Fixed effects:
                         Estimate Std. Error z value Pr(>|z|)   
(Intercept)                0.4236     0.2212   1.915  0.05549 . 
logic_belief_consistent1   1.3499     0.4425   3.051  0.00228 **
---
Signif. codes:  0 ‘***’ 0.001 ‘**’ 0.01 ‘*’ 0.05 ‘.’ 0.1 ‘ ’ 1
\end{Verbatim}
\caption{Mixed-effects logistic regression.} \label{app:tab:syl_stats:Chinchilla:regression}
\end{subtable}\\
\begin{subtable}{\textwidth}
\begin{Verbatim}[fontsize=\tiny]
Pearson's Chi-squared test with Yates' continuity correction
X-squared = 6.0122e-31, df = 1, p-value = 1
\end{Verbatim}
\caption{\(\chi^2\) test of interaction.} \label{app:tab:syl_stats:Chinchilla:chisq}
\end{subtable}
\caption{Statistical analyses of Chinchilla's performance on the Syllogism tasks, using (\subref{app:tab:syl_stats:Chinchilla:regression}) a mixed-effects logistic regression or (\subref{app:tab:syl_stats:Chinchilla:chisq}) a \(\chi^2\) test of the interaction effect. Chinchilla shows significant content effects, but no interaction with consistency.}
\label{app:tab:syl_stats:Chinchilla}
\end{table*}

\begin{table*}[htbp]
\begin{subtable}{\textwidth}
\begin{Verbatim}[fontsize=\tiny]
Generalized linear mixed model fit by maximum likelihood (Laplace
  Approximation) [glmerMod]
 Family: binomial  ( logit )
Formula: response_correct ~ logic_belief_consistent + (1 | syllogism_name)
   Data: syllogism_model_df %>% filter(subject == this_subject)

     AIC      BIC   logLik deviance df.resid 
   102.0    109.6    -48.0     96.0       93 

Scaled residuals: 
    Min      1Q  Median      3Q     Max 
-2.4202 -0.6095  0.4132  0.4132  1.6408 

Random effects:
 Groups         Name        Variance Std.Dev.
 syllogism_name (Intercept) 0        0       
Number of obs: 96, groups:  syllogism_name, 12

Fixed effects:
                         Estimate Std. Error z value Pr(>|z|)    
(Intercept)                0.3886     0.2611   1.488    0.137    
logic_belief_consistent1   2.7581     0.5222   5.281 1.28e-07 ***
---
Signif. codes:  0 ‘***’ 0.001 ‘**’ 0.01 ‘*’ 0.05 ‘.’ 0.1 ‘ ’ 1
\end{Verbatim}
\caption{Mixed-effects logistic regression.} \label{app:tab:syl_stats:PaLM2-M:regression}
\end{subtable}\\
\begin{subtable}{\textwidth}
\begin{Verbatim}[fontsize=\tiny]
Pearson's Chi-squared test with Yates' continuity correction
X-squared = 0, df = 1, p-value = 1
\end{Verbatim}
\caption{\(\chi^2\) test of interaction.} \label{app:tab:syl_stats:PaLM2-M:chisq}
\end{subtable}
\caption{Statistical analyses of PaLM 2-M's performance on the Syllogism tasks, using (\subref{app:tab:syl_stats:PaLM2-M:regression}) a mixed-effects logistic regression or (\subref{app:tab:syl_stats:PaLM2-M:chisq}) a \(\chi^2\) test. PaLM 2-M shows significant content effects, but no interaction with consistency.} \label{app:tab:syl_stats:PaLM2-M}
\end{table*}

\begin{table*}[htbp]
\begin{subtable}{\textwidth}
\begin{Verbatim}[fontsize=\tiny]
Chi-squared test for given probabilities
X-squared = 6.3913, df = 1, p-value = 0.01147
\end{Verbatim}
\caption{\(\chi^2\) test of content effect.} \label{app:tab:syl_stats:PaLM2-L:regression}
\end{subtable}\\
\begin{subtable}{\textwidth}
\begin{Verbatim}[fontsize=\tiny]
Pearson's Chi-squared test with Yates' continuity correction
X-squared = 14.318, df = 1, p-value = 0.0001544
\end{Verbatim}
\caption{\(\chi^2\) test of interaction.} \label{app:tab:syl_stats:PaLM2-L:chisq}
\end{subtable}
\caption{Statistical analyses of PaLM 2-L's performance on the Syllogism tasks, using (\subref{app:tab:syl_stats:PaLM2-L:regression}) a \(\chi^2\) test of the content effect as none of the regressions converged, and (\subref{app:tab:syl_stats:PaLM2-L:chisq}) a \(\chi^2\) test of the interaction. PaLM 2-L shows both significant content effects, and a significant interaction with consistency (as measured by the \(\chi^2\) test, as the regression with an interaction failed to converge).} \label{app:tab:syl_stats:PaLM2-L}
\end{table*}

\begin{table*}[htbp]
\begin{subtable}{\textwidth}
\begin{Verbatim}[fontsize=\tiny]
Generalized linear mixed model fit by maximum likelihood (Laplace
  Approximation) [glmerMod]
 Family: binomial  ( logit )
Formula: response_correct ~ logic_belief_consistent + (1 | syllogism_name)
   Data: syllogism_model_df %>% filter(subject == this_subject)

     AIC      BIC   logLik deviance df.resid 
   100.1    107.8    -47.0     94.1       93 

Scaled residuals: 
    Min      1Q  Median      3Q     Max 
-3.3166 -1.0000  0.3015  0.4761  1.0000 

Random effects:
 Groups         Name        Variance Std.Dev.
 syllogism_name (Intercept) 0        0       
Number of obs: 96, groups:  syllogism_name, 12

Fixed effects:
                         Estimate Std. Error z value Pr(>|z|)    
(Intercept)                1.1989     0.2984   4.019 5.86e-05 ***
logic_belief_consistent1   2.3979     0.5967   4.019 5.86e-05 ***
---
Signif. codes:  0 ‘***’ 0.001 ‘**’ 0.01 ‘*’ 0.05 ‘.’ 0.1 ‘ ’ 1
\end{Verbatim}
\caption{Mixed-effects logistic regression.} \label{app:tab:syl_stats:Flan-PaLM2:regression}
\end{subtable}\\
\begin{subtable}{\textwidth}
\begin{Verbatim}[fontsize=\tiny]
Pearson's Chi-squared test with Yates' continuity correction
X-squared = 17.913, df = 1, p-value = 2.312e-05
\end{Verbatim}
\caption{\(\chi^2\) test of interaction.} \label{app:tab:syl_stats:Flan-PaLM2:chisq}
\end{subtable}
\caption{Statistical analyses of Flan-PaLM 2's performance on the Syllogism tasks, using (\subref{app:tab:syl_stats:Flan-PaLM2:regression}) a mixed-effects logistic regression or (\subref{app:tab:syl_stats:Flan-PaLM2:chisq}) a \(\chi^2\) test. Flan-PaLM 2 shows both significant content effects, and a significant interaction with consistency (as measured by the \(\chi^2\) test, as the regression with an interaction failed to converge).} \label{app:tab:syl_stats:Flan-PaLM2}
\end{table*}

\begin{table*}[htbp]
\begin{subtable}{\textwidth}
\begin{Verbatim}[fontsize=\tiny]
Generalized linear mixed model fit by maximum likelihood (Laplace
  Approximation) [glmerMod]
 Family: binomial  ( logit )
Formula: response_correct ~ logic_belief_consistent + (1 | syllogism_name)
   Data: syllogism_model_df %>% filter(subject == this_subject)

     AIC      BIC   logLik deviance df.resid 
   131.8    139.5    -62.9    125.8       93 

Scaled residuals: 
    Min      1Q  Median      3Q     Max 
-1.4832 -0.9199  0.6742  0.6742  1.0871 

Random effects:
 Groups         Name        Variance Std.Dev.
 syllogism_name (Intercept) 0        0       
Number of obs: 96, groups:  syllogism_name, 12

Fixed effects:
                         Estimate Std. Error z value Pr(>|z|)  
(Intercept)                0.3107     0.2127   1.461   0.1440  
logic_belief_consistent1   0.9555     0.4253   2.247   0.0247 *
---
Signif. codes:  0 ‘***’ 0.001 ‘**’ 0.01 ‘*’ 0.05 ‘.’ 0.1 ‘ ’ 1
\end{Verbatim}
\caption{Mixed-effects logistic regression.} \label{app:tab:syl_stats:GPT-3.5:regression}
\end{subtable}\\
\begin{subtable}{\textwidth}
\begin{Verbatim}[fontsize=\tiny]
Pearson's Chi-squared test with Yates' continuity correction
X-squared = 25.507, df = 1, p-value = 4.408e-07
\end{Verbatim}
\caption{\(\chi^2\) test.} \label{app:tab:syl_stats:GPT-3.5:chisq}
\end{subtable}
\caption{Statistical analyses of GPT-3.5-turbo-instruct's performance on the syl tasks, using (\subref{app:tab:syl_stats:GPT-3.5:regression}) a mixed-effects logistic regression or (\subref{app:tab:syl_stats:GPT-3.5:chisq}) a \(\chi^2\) test. GPT-3.5 shows both significant content effects, and a significant interaction with consistency (as measured by the \(\chi^2\) test, as the regression with an interaction failed to converge).}
\label{app:tab:syl_stats:GPT-3.5}
\end{table*}

\FloatBarrier
\subsection{Wason} \label{app:statistical_analyses:wason}

We report mixed-effects logistic regressions for humans (both all humans, and the fast and slow groups individually) and all models in Tables \ref{app:tab:wason_stats:human_all}-\ref{app:tab:wason_stats:GPT-3.5-turbo-instruct}. We observe a significant effect of content in most cases. However, the fast humans alone do not show a significant content effect. Furthermore, the content effects in PaLM 2-M and Flan-PaLM 2 are only marginally significant, due to high item level variance.

In Appx. \ref{app:statistical_analyses:wason:human} we further analyze the human data whil incorporating response time in the regression.

\begin{table*}[htbp]
\begin{Verbatim}[fontsize=\tiny]
Generalized linear mixed model fit by maximum likelihood (Laplace
  Approximation) [glmerMod]
 Family: binomial  ( logit )
Formula: response_correct ~ wason_condition + (1 | wason_name)
   Data: wason_joint_df %>% filter(subject_no_rt == "Human", wason_condition %in%  
    c("Arbitrary", "Realistic"))

     AIC      BIC   logLik deviance df.resid 
     478      491     -236      472      571 

Scaled residuals: 
    Min      1Q  Median      3Q     Max 
-0.7162 -0.4629 -0.3279 -0.2881  3.4711 

Random effects:
 Groups     Name        Variance Std.Dev.
 wason_name (Intercept) 0.2309   0.4805  
Number of obs: 574, groups:  wason_name, 25

Fixed effects:
                         Estimate Std. Error z value Pr(>|z|)    
(Intercept)               -2.2951     0.2554  -8.988   <2e-16 ***
wason_conditionRealistic   0.8219     0.3235   2.541   0.0111 *  
---
Signif. codes:  0 ‘***’ 0.001 ‘**’ 0.01 ‘*’ 0.05 ‘.’ 0.1 ‘ ’ 1
\end{Verbatim}
\caption{Statistical analysis of human performance (collapsing across fast and slow subjects) on the Wason tasks, using a logistic regression. There is a significant content effect.} \label{app:tab:wason_stats:human_all}
\end{table*}

\begin{table*}[htbp]
\begin{Verbatim}[fontsize=\tiny]
Generalized linear mixed model fit by maximum likelihood (Laplace
  Approximation) [glmerMod]
 Family: binomial  ( logit )
Formula: response_correct ~ wason_condition + (1 | wason_name)
   Data: 
wason_joint_df %>% filter(subject == this_subject, wason_condition %in%  
    c("Arbitrary", "Realistic"))

     AIC      BIC   logLik deviance df.resid 
   305.1    317.4   -149.6    299.1      442 

Scaled residuals: 
    Min      1Q  Median      3Q     Max 
-0.5732 -0.3620 -0.3034 -0.2709  3.7157 

Random effects:
 Groups     Name        Variance Std.Dev.
 wason_name (Intercept) 0.273    0.5225  
Number of obs: 445, groups:  wason_name, 25

Fixed effects:
                         Estimate Std. Error z value Pr(>|z|)    
(Intercept)               -2.4740     0.2948  -8.393   <2e-16 ***
wason_conditionRealistic   0.4570     0.3877   1.179    0.239    
---
Signif. codes:  0 ‘***’ 0.001 ‘**’ 0.01 ‘*’ 0.05 ‘.’ 0.1 ‘ ’ 1
\end{Verbatim}
\caption{Statistical analysis of human (fast subjects only) performance on the Wason tasks, using a logistic regression. We do not observe a significant content effect} \label{app:tab:wason_stats:human_fast}
\end{table*}

\begin{table*}[htbp]
\begin{Verbatim}[fontsize=\tiny]
Generalized linear mixed model fit by maximum likelihood (Laplace
  Approximation) [glmerMod]
 Family: binomial  ( logit )
Formula: response_correct ~ wason_condition + (1 | wason_name)
   Data: 
wason_joint_df %>% filter(subject == this_subject, wason_condition %in%  
    c("Arbitrary", "Realistic"))

     AIC      BIC   logLik deviance df.resid 
   156.8    165.4    -75.4    150.8      126 

Scaled residuals: 
    Min      1Q  Median      3Q     Max 
-0.8934 -0.7468 -0.4277  1.1193  2.3380 

Random effects:
 Groups     Name        Variance Std.Dev.
 wason_name (Intercept) 0.1906   0.4365  
Number of obs: 129, groups:  wason_name, 24

Fixed effects:
                         Estimate Std. Error z value Pr(>|z|)    
(Intercept)               -1.6534     0.4303  -3.842 0.000122 ***
wason_conditionRealistic   1.1518     0.5009   2.299 0.021495 *  
---
Signif. codes:  0 ‘***’ 0.001 ‘**’ 0.01 ‘*’ 0.05 ‘.’ 0.1 ‘ ’ 1
\end{Verbatim}
\caption{Statistical analysis of human (slow) performance on the Wason tasks, using a logistic regression. There is a significant content effect.} \label{app:tab:wason_stats:human_slow}
\end{table*}

\begin{table*}[htbp]
\begin{Verbatim}[fontsize=\tiny]
Generalized linear mixed model fit by maximum likelihood (Laplace
  Approximation) [glmerMod]
 Family: binomial  ( logit )
Formula: response_correct ~ wason_condition + (1 | wason_name)
   Data: 
wason_joint_df %>% filter(subject == this_subject, wason_condition %in%  
    c("Arbitrary", "Realistic"))

     AIC      BIC   logLik deviance df.resid 
   136.9    146.0    -65.5    130.9      147 

Scaled residuals: 
    Min      1Q  Median      3Q     Max 
-1.9739 -0.2756 -0.1256  0.4490  2.6782 

Random effects:
 Groups     Name        Variance Std.Dev.
 wason_name (Intercept) 6.068    2.463   
Number of obs: 150, groups:  wason_name, 25

Fixed effects:
                         Estimate Std. Error z value Pr(>|z|)   
(Intercept)                -3.584      1.154  -3.106  0.00190 **
wason_conditionRealistic    3.576      1.354   2.640  0.00828 **
---
Signif. codes:  0 ‘***’ 0.001 ‘**’ 0.01 ‘*’ 0.05 ‘.’ 0.1 ‘ ’ 1
\end{Verbatim}
\caption{Statistical analysis of Chinchilla's performance on the Wason tasks, using a logistic regression. There is a significant content effect.} \label{app:tab:wason_stats:Chinchilla}
\end{table*}

\begin{table*}[htbp]
\begin{Verbatim}[fontsize=\tiny]
Generalized linear mixed model fit by maximum likelihood (Laplace
  Approximation) [glmerMod]
 Family: binomial  ( logit )
Formula: response_correct ~ wason_condition + (1 | wason_name)
   Data: 
wason_joint_df %>% filter(subject == this_subject, wason_condition %in%  
    c("Arbitrary", "Realistic"))

     AIC      BIC   logLik deviance df.resid 
   115.5    124.6    -54.8    109.5      147 

Scaled residuals: 
     Min       1Q   Median       3Q      Max 
-2.14109 -0.13581 -0.04492  0.15813  1.50757 

Random effects:
 Groups     Name        Variance Std.Dev.
 wason_name (Intercept) 30.12    5.488   
Number of obs: 150, groups:  wason_name, 25

Fixed effects:
                         Estimate Std. Error z value Pr(>|z|)  
(Intercept)                -5.842      2.296  -2.544   0.0110 *
wason_conditionRealistic    5.122      2.777   1.844   0.0651 .
---
Signif. codes:  0 ‘***’ 0.001 ‘**’ 0.01 ‘*’ 0.05 ‘.’ 0.1 ‘ ’ 1
\end{Verbatim}
\caption{Statistical analysis of PaLM 2-M's performance on the Wason tasks, using a logistic regression. There is a marginally-significant content effect, due to high item-level variance.} \label{app:tab:wason_stats:PaLM2-M}
\end{table*}

\begin{table*}[htbp]
\begin{Verbatim}[fontsize=\tiny]
Generalized linear mixed model fit by maximum likelihood (Laplace
  Approximation) [glmerMod]
 Family: binomial  ( logit )
Formula: response_correct ~ wason_condition + (1 | wason_name)
   Data: 
wason_joint_df %>% filter(subject == this_subject, wason_condition %in%  
    c("Arbitrary", "Realistic"))

     AIC      BIC   logLik deviance df.resid 
   142.6    151.6    -68.3    136.6      147 

Scaled residuals: 
    Min      1Q  Median      3Q     Max 
-2.2846 -0.1708  0.1686  0.2894  2.2759 

Random effects:
 Groups     Name        Variance Std.Dev.
 wason_name (Intercept) 9.488    3.08    
Number of obs: 150, groups:  wason_name, 25

Fixed effects:
                         Estimate Std. Error z value Pr(>|z|)  
(Intercept)                -1.921      1.182  -1.625   0.1043  
wason_conditionRealistic    3.908      1.759   2.222   0.0263 *
---
Signif. codes:  0 ‘***’ 0.001 ‘**’ 0.01 ‘*’ 0.05 ‘.’ 0.1 ‘ ’ 1
\end{Verbatim}
\caption{Statistical analysis of PaLM 2-L's performance on the Wason tasks, using a logistic regression. There is a significant content effect.} \label{app:tab:wason_stats:PaLM2-L}
\end{table*}

\begin{table*}[htbp]
\begin{Verbatim}[fontsize=\tiny]
Generalized linear mixed model fit by maximum likelihood (Laplace
  Approximation) [glmerMod]
 Family: binomial  ( logit )
Formula: response_correct ~ wason_condition + (1 | wason_name)
   Data: 
wason_joint_df %>% filter(subject == this_subject, wason_condition %in%  
    c("Arbitrary", "Realistic"))

     AIC      BIC   logLik deviance df.resid 
   129.1    138.1    -61.6    123.1      147 

Scaled residuals: 
    Min      1Q  Median      3Q     Max 
-1.4634 -0.2275 -0.1330  0.1271  2.2738 

Random effects:
 Groups     Name        Variance Std.Dev.
 wason_name (Intercept) 18.14    4.259   
Number of obs: 150, groups:  wason_name, 25

Fixed effects:
                         Estimate Std. Error z value Pr(>|z|)  
(Intercept)                -2.143      1.591  -1.347   0.1778  
wason_conditionRealistic    4.539      2.551   1.779   0.0752 .
---
Signif. codes:  0 ‘***’ 0.001 ‘**’ 0.01 ‘*’ 0.05 ‘.’ 0.1 ‘ ’ 1
\end{Verbatim}
\caption{Statistical analysis of Flan-PaLM 2's performance on the Wason tasks, using a logistic regression. There is a marginally-significant content effect, due to high item-level variance.} \label{app:tab:wason_stats:Flan-PaLM2}
\end{table*}

\begin{table*}[htbp]
\begin{Verbatim}[fontsize=\tiny]
Generalized linear mixed model fit by maximum likelihood (Laplace
  Approximation) [glmerMod]
 Family: binomial  ( logit )
Formula: response_correct ~ wason_condition + (1 | wason_name)
   Data: 
wason_joint_df %>% filter(subject == this_subject, wason_condition %in%  
    c("Arbitrary", "Realistic"))

     AIC      BIC   logLik deviance df.resid 
   154.3    163.3    -74.1    148.3      147 

Scaled residuals: 
    Min      1Q  Median      3Q     Max 
-2.1094 -0.4257 -0.1980  0.3909  2.3489 

Random effects:
 Groups     Name        Variance Std.Dev.
 wason_name (Intercept) 5.008    2.238   
Number of obs: 150, groups:  wason_name, 25

Fixed effects:
                         Estimate Std. Error z value Pr(>|z|)   
(Intercept)                -2.105      0.867  -2.428  0.01517 * 
wason_conditionRealistic    3.092      1.188   2.603  0.00923 **
---
Signif. codes:  0 ‘***’ 0.001 ‘**’ 0.01 ‘*’ 0.05 ‘.’ 0.1 ‘ ’ 1
\end{Verbatim}
\caption{Statistical analysis of GPT-3.5-turbo-instruct's performance on the Wason tasks, using a logistic regression. There is a significant content effect.} \label{app:tab:wason_stats:GPT-3.5-turbo-instruct}
\end{table*}

\FloatBarrier
\subsubsection{Human analyses incorporating response time} \label{app:statistical_analyses:wason:human}

Here we present two regression analyses of the human results that incorporate the response time. In Table \ref{app:tab:wason_stats:human_with_rt} we show a mixed-effects logistic regression controlling for log response time; the content effect remains significant. Thus, the content effects are not solely driven by the differences in response time noted above (Appx. \ref{app:fig:wason_human_rt_distributions}).

However, it is also possible to conceive of the shift in response time as \emph{a part of} the content effect. We can analyze the data this way by \(z\)-scoring response time within each condition; thus, the effect of the mean difference in response time will be included in the condition predictor. We present these results in Table \ref{app:tab:wason_stats:human_with_zscored_rt}. Both content and \(z\)-scored response time remain significant predictors of success.

\begin{table*}[htbp]
\begin{Verbatim}[fontsize=\tiny]
Generalized linear mixed model fit by maximum likelihood (Laplace
  Approximation) [glmerMod]
 Family: binomial  ( logit )
Formula: response_correct ~ wason_condition + scale(log(rt)) + (1 | wason_name)
   Data: wason_human_correct_df

     AIC      BIC   logLik deviance df.resid 
   459.5    476.9   -225.8    451.5      570 

Scaled residuals: 
    Min      1Q  Median      3Q     Max 
-1.0996 -0.4417 -0.3265 -0.2246  4.5293 

Random effects:
 Groups     Name        Variance Std.Dev.
 wason_name (Intercept) 0.2539   0.5039  
Number of obs: 574, groups:  wason_name, 25

Fixed effects:
                         Estimate Std. Error z value Pr(>|z|)    
(Intercept)               -2.3248     0.2655  -8.755  < 2e-16 ***
wason_conditionRealistic   0.6659     0.3350   1.988   0.0468 *  
scale(log(rt))             0.5637     0.1269   4.442 8.89e-06 ***
---
Signif. codes:  0 ‘***’ 0.001 ‘**’ 0.01 ‘*’ 0.05 ‘.’ 0.1 ‘ ’ 1
\end{Verbatim}
\caption{Statistical analysis of human (both fast and slow) performance on the Wason tasks, using a logistic regression and also controlling for (log) response time. The content effect remains significant.} \label{app:tab:wason_stats:human_with_rt}
\end{table*}

\begin{table*}[htbp]
\begin{Verbatim}[fontsize=\tiny]
Generalized linear mixed model fit by maximum likelihood (Laplace
  Approximation) [glmerMod]
 Family: binomial  ( logit )
Formula: response_correct ~ wason_condition + zscored_rt_by_condition +  
    (1 | wason_name)
   Data: wason_human_correct_df

     AIC      BIC   logLik deviance df.resid 
   459.5    476.9   -225.8    451.5      570 

Scaled residuals: 
    Min      1Q  Median      3Q     Max 
-1.1002 -0.4417 -0.3265 -0.2247  4.5269 

Random effects:
 Groups     Name        Variance Std.Dev.
 wason_name (Intercept) 0.254    0.504   
Number of obs: 574, groups:  wason_name, 25

Fixed effects:
                         Estimate Std. Error z value Pr(>|z|)    
(Intercept)               -2.4329     0.2700  -9.011  < 2e-16 ***
wason_conditionRealistic   0.8793     0.3349   2.626  0.00865 ** 
zscored_rt_by_condition    0.5540     0.1247   4.443 8.87e-06 ***
---
Signif. codes:  0 ‘***’ 0.001 ‘**’ 0.01 ‘*’ 0.05 ‘.’ 0.1 ‘ ’ 1
\end{Verbatim}
\caption{Statistical analysis of human (both fast and slow) performance on the Wason tasks, using a logistic regression and also controlling for (log) response time, but \(z\)-scored \emph{within} condition. Again the content effect is significant.} \label{app:tab:wason_stats:human_with_zscored_rt}
\end{table*}

\FloatBarrier
\subsubsection{Multinomial regression of the response patterns on the Wason tasks} \label{app:statistical_analyses:wason:answer_choices_multinomial}

In Table \ref{app:tab:wason_stats:multinomial_choices} we present the results of a multinomial logistic regression predicting which of the six possible subsets of answers the humans and language models chose on the Wason task. This regression quantitatively supports the claim that the behavior is nonrandom, and more generally quantifies the qualitative observations of response patterns made in the main text.

\begin{table*}[htbp]
\begin{subtable}{\textwidth}
\begin{Verbatim}[fontsize=\tiny]
A matrix: 5 × 9 of type dbl
            (Intercept)     Realist      Nonsense    humanslow    chinchi      palm2_m      palm2_l      flanpalm2      gpt3.5
AT,CF (correct) -2.64	 1.43	 0.75	1.20	 2.31	 2.31	 3.17	  3.19	 2.34
AT,AF   	-1.24	-1.35	-2.16	0.32	 0.08	-0.21	-0.06	  0.53	-0.59
AF,CT   	-2.65	 0.06	 0.32	0.02	 2.75	 2.81	 2.72	  2.29	 2.29
AF,CF   	-2.45	 0.64	 0.21	0.22	 2.31	 1.38	 2.05	  2.27	 1.13
CT,CF   	-2.47	 0.30	-1.07	0.97	-1.16	-0.60	-12.95	-12.33   -13.45
\end{Verbatim}
\caption{Coefficients.} \label{app:tab:wason_stats:multinomial_choices:coeffs}
\end{subtable}\\
\begin{subtable}{\textwidth}
\begin{Verbatim}[fontsize=\tiny]
A matrix: 5 × 9 of type dbl
            (Intercept)    Realist    Nonsense    humanslow    chinchi     palm2_m     palm2_l     flanpalm2     gpt3.5
AT,CF (correct) 0.18	0.17	0.17	0.25	0.24	0.23	0.25	  0.25	0.22
AT,AF           0.16	0.29	0.44	0.35	0.44	0.46	0.51	  0.41	0.49
AF,CT           0.22	0.20	0.19	0.50	0.28	0.27	0.30	  0.33	0.28
AF,CF           0.20	0.20	0.21	0.38	0.27	0.30	0.31	  0.29	0.31
CT,CF           0.26	0.33	0.56	0.35	1.03	0.75	0.00	306.59	0.00
\end{Verbatim}
\caption{Standard errors.} \label{app:tab:wason_stats:multinomial_choices:std_errs}
\end{subtable}\\
\caption{Results of a multinomial logistic regression predicting the answer choices (reference level is AT,CT --- the matching bias) from participants and language models based on condition (reference level is the Arbitrary condition), and participant group (reference level is fast humans). The regression was performed with dummy coding, so coefficients represent the difference in log odds relative to the reference level in each case. We present both the (\subref{app:tab:wason_stats:multinomial_choices:coeffs}) coefficients estimated by the regression and (\subref{app:tab:wason_stats:multinomial_choices:coeffs}) their standard errors. There are a variety of noticeable effects, including the overall matching bias in the fast humans (the fact that the intercept coefficients are all negative), the basic content effect that Realitic problems are more likely to yield correct answers, and the finding that language models and slow humans tend to give correct answers more often than fast humans. Additionally, many qualitative patterns reported in Fig. \ref{fig:wason_choice_details} are statistically borne out by this analysis.
Note that due to some models rarely giving some responses, certain coefficient estimates are unstable, particularly in the CT,CF row.} \label{app:tab:wason_stats:multinomial_choices}
\end{table*}

\FloatBarrier
\subsection{Response time and model log-probability differences} \label{app:statistical_analyses:logprob_rt}

In this section we present the mixed-effects linear regressions comparing human response times and model log-probabilities on the NLI and syllogisms tasks, in Tables \ref{app:tab:logprob_rt_stats:nli} and \ref{app:tab:logprob_rt_stats:syl}, respectively. In order to make these comparisons, we breakdown each problem into cases where both humans and models got it correct, and cases where both got it wrong, and only compare log-probabilities and response times within these cases. This breakdown is necessary to control for accuracy in these models, as it is a significantly related to both response times and log-probabilities. Note, however, that this means that problems where a model answered correctly but humans never answered correctly, or vice versa, are omitted.

In both tasks, we see significant effects of the content on the model log-probability differences; even controlling for these we see significant relationships to the human response times, such that on items on which the humans respond more slowly, the models show smaller differences in log-probabilities.

\begin{table*}[htbp]
\begin{Verbatim}[fontsize=\tiny]
Linear mixed model fit by REML ['lmerMod']
Formula: 
zscored_logprob_diff ~ log(Human) + consistent_plottable + response_correct +  
    (1 | model) + (1 | name)
   Data: nli_logprob_rt_corr_df

REML criterion at convergence: 2074.8

Scaled residuals: 
    Min      1Q  Median      3Q     Max 
-2.7422 -0.6124  0.0099  0.6353  3.8258 

Random effects:
 Groups   Name        Variance Std.Dev.
 name     (Intercept) 0.2772   0.5265  
 model    (Intercept) 0.0000   0.0000  
 Residual             0.5468   0.7394  
Number of obs: 831, groups:  name, 171; model, 5

Fixed effects:
                               Estimate Std. Error t value
(Intercept)                      0.7544     0.5353   1.409
log(Human)                      -0.5392     0.1590  -3.392
consistent_plottableConsistent   0.6136     0.1345   4.563
consistent_plottableNonsense    -0.1841     0.1427  -1.291
response_correctTRUE             0.7889     0.2383   3.311
\end{Verbatim}
\caption{Statistical analysis of the relationship between human response times and language model log-probability differences on the NLI tasks, using a mixed-effects regression controlling for the task variables and answer correctness, as well as random effects of the item and LM. Note that the model log-probabilities are significantly affected by the content, even though the model accuracy is not.} \label{app:tab:logprob_rt_stats:nli}
\end{table*}

\begin{table*}[htbp]
\begin{Verbatim}[fontsize=\tiny]
Linear mixed model fit by REML ['lmerMod']
Formula: zscored_logprob_diff ~ log(Human) + logic_belief_consistent +  
    consistent_plottable + response_correct + (1 | model) + (1 |  
    syllogism_name)
   Data: syllogism_logprob_rt_corr_df_2

REML criterion at convergence: 1077

Scaled residuals: 
     Min       1Q   Median       3Q      Max 
-2.33241 -0.73009 -0.02953  0.61543  2.99660 

Random effects:
 Groups         Name        Variance Std.Dev.
 syllogism_name (Intercept) 0.055391 0.23535 
 model          (Intercept) 0.005615 0.07493 
 Residual                   0.829587 0.91082 
Number of obs: 394, groups:  syllogism_name, 36; model, 5

Fixed effects:
                             Estimate Std. Error t value
(Intercept)                   1.18573    0.70847   1.674
log(Human)                   -0.41458    0.20342  -2.038
logic_belief_consistent       0.08477    0.05899   1.437
consistent_plottableviolate  -0.27369    0.07090  -3.860
consistent_plottablenonsense -0.10094    0.09245  -1.092
response_correctTRUE          0.49560    0.10421   4.756
\end{Verbatim}
\caption{Statistical analysis of the relationship between human response times and language model log-probability differences on the Syllogisms tasks, using a mixed-effects regression controlling for the task variables and answer correctness, as well as random effects of the item and LM.} \label{app:tab:logprob_rt_stats:syl}
\end{table*}

\end{document}